\documentclass[10pt,twocolumn,letterpaper]{article}

\usepackage[pagenumbers]{cvpr} %

\usepackage{graphicx}
\usepackage{amsmath}
\usepackage{amssymb}
\usepackage{booktabs}

\usepackage[pagebackref,breaklinks,colorlinks]{hyperref}

\usepackage{nicefrac} 
\usepackage{xcolor}
\usepackage{graphicx}    
\usepackage{xspace}
\usepackage{tabularx}
\usepackage{multirow}
\usepackage{verbatim}
\usepackage{tabularx}
\usepackage{colortbl}
\usepackage{array, makecell} %
\usepackage{xstring}
\makeatletter
\@namedef{ver@everyshi.sty}{}
\makeatother
\usepackage{tikz}
\usetikzlibrary{shapes.geometric,positioning,shapes.multipart,backgrounds,arrows,patterns,calc,fit,3d}
\usepackage{pgfplots}
\pgfplotsset{compat=1.17} %

\usepackage[capitalize]{cleveref}
\crefname{section}{Sec.}{Secs.}
\Crefname{section}{Section}{Sections}
\Crefname{table}{Table}{Tables}
\crefname{table}{Tab.}{Tabs.}

\def\cvprPaperID{1239} %
\def\confName{CVPR}
\def\confYear{2023}

\usepackage{xspace}
\usepackage[export]{adjustbox}
\usepackage{pgf}
\usepackage{xparse}
\usepackage{enumitem}
\usepackage{tikzsymbols}

\usepackage{bold-extra}

\makeatletter

\tikzoption{canvas is xy plane at z}[]{%
  \def\tikz@plane@origin{\pgfpointxyz{0}{0}{#1}}%
  \def\tikz@plane@x{\pgfpointxyz{1}{0}{#1}}%
  \def\tikz@plane@y{\pgfpointxyz{0}{1}{#1}}%
  \tikz@canvas@is@plane
}

\newcommand\notsotiny{\@setfontsize\notsotiny\@vipt\@viipt}

\makeatother

\NewDocumentCommand{\DrawCubes}{O{} m m m m m m }{%
    \def\XGridMin{#2}
    \def\XGridMax{#3}
    \def\YGridMin{#4}
    \def\YGridMax{#5}
    \def\ZGridMin{#6}
    \def\ZGridMax{#7}
    \begin{scope}[canvas is xy plane at z=\ZGridMax]
      \draw [#1] (\XGridMin,\YGridMin) grid (\XGridMax,\YGridMax) rectangle (\XGridMin,\YGridMin);
    \end{scope}
    \begin{scope}[canvas is yz plane at x=\XGridMax]
      \draw [#1] (\YGridMin,\ZGridMin) grid (\YGridMax,\ZGridMax) rectangle (\YGridMin,\ZGridMin);
    \end{scope}
    \begin{scope}[canvas is xz plane at y=\YGridMin]
      \draw [#1] (\XGridMin,\ZGridMin) grid (\XGridMax,\ZGridMax) rectangle (\XGridMin,\ZGridMin);
    \end{scope}
}%

\NewDocumentCommand{\CubeCoord}{O{} m m m m m m }{%
    \def\XGridMin{#2}
    \def\XGridMax{#3}
    \def\YGridMin{#4}
    \def\YGridMax{#5}
    \def\ZGridMin{#6}
    \def\ZGridMax{#7}
    \begin{scope}[canvas is xz plane at y=\YGridMin]
      \coordinate (#1) at ( $ (\XGridMin,\ZGridMin)!0.5!(\XGridMax,\ZGridMax)$ );
    \end{scope}
}%

\NewDocumentCommand{\DrawXYZ}{O{}}{%
    \begin{scope}[canvas is xy plane at z=0]
      \draw [#1, -latex, black] (0,0) -- (0.2,0) node [yshift=-2pt,right, black] {\scriptsize $x$};
    \end{scope}
    \begin{scope}[canvas is yz plane at x=0]
      \draw [#1, -latex, black] (0,0) -- (0.2,0) node [yshift=2pt, right, black] {\scriptsize $t$};
    \end{scope}
    \begin{scope}[canvas is xz plane at y=0]
      \draw [#1, -latex, black] (0,0) -- (0,0.2) node [above, black] {\scriptsize $y$};
    \end{scope}
}%

\newcommand{\ourMethod}{TarViS}

\newcommand{\PAR}[1]{\vskip4pt \noindent {\bf #1~}}

\definecolor{better_blue}{HTML}{4285f4}

\newtoggle{comments}
\togglefalse{comments}

\iftoggle{comments}{%
\newcommand{\ali}[1]{\textcolor{blue}{\textbf{Ali: }{#1}}}
\newcommand{\alex}[1]{\textcolor[rgb]{0,0.5,0}{\textbf{Alex: }{#1}}}
\newcommand{\jono}[1]{\textcolor{brown}{\textbf{Jono: }{#1}}}
\newcommand{\deva}[1]{\textcolor[rgb]{0,0.5,0.5}{\textbf{Deva: }{#1}}}
\newcommand{\bastian}[1]{\textcolor[rgb]{0.5,0.5,0}{\textbf{Bastian: }{#1}}}
\newcommand{\ext}[1]{\textcolor{orange}{\textbf{Ext. Rev: }{#1}}}
\newcommand{\todo}[1]{\textcolor{red}{\small Todo:\,#1}\PackageWarning{TODO:}{#1!}}

}
{%
\newcommand{\ali}[1]{}
\newcommand{\alex}[1]{}
\newcommand{\jono}[1]{}
\newcommand{\deva}[1]{}
\newcommand{\bastian}[1]{}
\newcommand{\ext}[1]{}
\newcommand{\todo}[1]{}

}

\newcolumntype{Y}{>{\centering\arraybackslash}X}
\newcolumntype{P}[1]{>{\centering\arraybackslash}p{#1}}

\usepackage{pifont}%

\newcommand{\cmarkg}{\textcolor{c_green}{\ding{51}}}%
\newcommand{\xmarkr}{\textcolor{c_red}{\ding{55}}}%

\newcommand{\J}{\mathcal{J}}
\newcommand{\F}{\mathcal{F}}
\newcommand{\JnF}{\mathcal{J}\&\mathcal{F}}

\newcommand{\real}[1]{\mathbb{R}^{#1}}

\definecolor{c_red}{HTML}{ea4335}
\definecolor{c_blue}{HTML}{4285f4}
\definecolor{c_orange}{HTML}{ff6d01}
\definecolor{c_green}{HTML}{34a853}
\definecolor{c_purple}{HTML}{9900ff}
\definecolor{c_other}{HTML}{000000}

\newcommand{\genrand}[1]{\pgfmathparse{70*rnd}
\edef\tmpa{\pgfmathresult}
\pgfmathparse{50*rnd+50*rnd+50*rnd-70+50*rnd}
\edef\tmpb{\pgfmathresult}
\pgfmathparse{50*rnd+50*rnd+50*rnd-70+50*rnd}
\edef\tmpc{\pgfmathresult}
\pgfmathparse{50*rnd+50*rnd+50*rnd-70+50*rnd}
\edef\tmpd{\pgfmathresult}
\pgfmathparse{50*rnd+50*rnd+50*rnd-70+50*rnd}
\edef\tmpe{\pgfmathresult}
\edef\maincol{#1}}

\tikzset{querystyle/.style={rectangle split, minimum height=7.5mm, minimum width=1.5mm, draw, inner sep=0pt, outer sep=0pt, ,rectangle split parts=5}, rectangle split part fill={\maincol!\tmpa,\maincol!\tmpb,\maincol!\tmpc,\maincol!\tmpd,\maincol!\tmpe}}

\newif\ifArxivMode

\begin{document}

\ArxivModetrue

\title{\vspace{-10pt}\ourMethod{}: A Unified Approach for Target-based Video Segmentation}

\author{
Ali Athar$^1$ %
\quad
Alexander Hermans$^1$%
\quad
Jonathon Luiten$^{1,2}$
\quad
Deva Ramanan$^2$%
\quad
Bastian Leibe$^1$\\[5pt]
$^1\hspace{1pt}$RWTH Aachen University, Germany \quad $^2\hspace{1pt}$Carnegie Mellon University, USA \\[5pt]
{\tt\small \{athar,hermans,luiten,leibe\}@vision.rwth-aachen.de}
\quad {\tt\small deva@cs.cmu.edu} \\
}

\maketitle

\begin{abstract}

The general domain of video segmentation is currently fragmented into different tasks spanning multiple benchmarks.
Despite rapid progress in the state-of-the-art, current methods are overwhelmingly task-specific and cannot conceptually generalize to other tasks.
Inspired by recent approaches with multi-task capability, we propose \ourMethod{}: a novel, unified network architecture that can be applied to any task that requires segmenting a set of arbitrarily defined `targets' in video.
Our approach is flexible with respect to how tasks define these targets, since it models the latter as abstract `queries' which are then used to predict pixel-precise target masks.
A single \ourMethod{} model can be trained jointly on a collection of datasets spanning different tasks, and can hot-swap between tasks during inference without any task-specific retraining. 
To demonstrate its effectiveness, we apply \ourMethod{} to four different tasks, namely Video Instance Segmentation (VIS), Video Panoptic Segmentation (VPS), Video Object Segmentation (VOS) and Point Exemplar-guided Tracking (PET).
Our unified, jointly trained model achieves state-of-the-art performance on 5/7 benchmarks spanning these four tasks, and competitive performance on the remaining two. 
Code and model weights are available at:  \footnotesize \url{https://github.com/Ali2500/TarViS}
\end{abstract}

\section{Introduction}
\label{sec:introduction}

\begin{figure}
\begin{tikzpicture}[font=\footnotesize]

\node[inner sep=0pt] (rgb_in3) {\includegraphics[width=12mm]{figures/arch_ims/00000_rgb.jpg}};
\node[inner sep=0pt, below left = 2mm and 2mm of rgb_in3.center, anchor=center] (rgb_in2) {\includegraphics[width=12mm]{figures/arch_ims/00000_rgb.jpg}};
\node[inner sep=0pt, below left = 2mm and 2mm of rgb_in2.center, anchor=center] (rgb_in1) {\includegraphics[width=12mm]{figures/arch_ims/00000_rgb.jpg}};
\node[inner sep=2pt,fit=(rgb_in3)(rgb_in1)] (rgb_in) {};

\node[rectangle, draw, c_orange, fill=c_orange!20, thick, below left = 5.5mm and 2mm of rgb_in.south, minimum width=8mm, minimum height=5mm] (vps) {VPS};
\node[rectangle, draw, c_blue, fill=c_blue!20, thick, left = 1mm of vps.west, anchor=east, minimum width=8mm, minimum height=5mm] (vis) {VIS};
\node[rectangle, draw, c_red, fill=c_red!20, thick, right = 1mm of vps.east, anchor=west, minimum width=8mm, minimum height=5mm] (vos) {VOS};
\node[rectangle, draw, c_green, fill=c_green!20, thick, right = 1mm of vos.east, anchor=west, minimum width=8mm, minimum height=5mm] (pet) {PET};

\draw[latex-] (vis.north) -- ++(0,2.5mm);
\draw[latex-] (vps.north) -- ++(0,2.5mm);
\draw[latex-] (vos.north) -- ++(0,2.5mm);
\draw[latex-] (pet.north) -- ++(0,2.5mm);

\draw[-latex] (vis.south) -- ++(0,-2.5mm);
\draw[-latex] (vps.south) -- ++(0,-2.5mm);
\draw[-latex] (vos.south) -- ++(0,-2.5mm);
\draw[-latex] (pet.south) -- ++(0,-2.5mm);

\node[inner sep=0pt, right = 3.5cm of rgb_in3.center, anchor=center, minimum width=1cm] (rgb_in3r) {\includegraphics[width=12mm]{figures/arch_ims/00000_rgb.jpg}};
\node[inner sep=0pt, below left = 2mm and 2mm of rgb_in3r.center, anchor=center] (rgb_in2r) {\includegraphics[width=12mm]{figures/arch_ims/00000_rgb.jpg}};
\node[inner sep=0pt, below left = 2mm and 2mm of rgb_in2r.center, anchor=center] (rgb_in1r) {\includegraphics[width=12mm]{figures/arch_ims/00000_rgb.jpg}};
\node[inner sep=2pt,fit=(rgb_in3r)(rgb_in1r)] (rgb_inr) {};

\genrand{c_blue}
\node (qvis) [querystyle, right= 1.3cm of rgb_in1r.south, anchor=south, draw=c_blue] {};
\genrand{c_orange}
\node (qvps) [querystyle, above right = 2mm and 2mm of qvis.center, anchor=center, draw=c_orange] {};
\genrand{c_red}
\node (qvos) [querystyle, below right = 2mm and 2mm of qvps.center, anchor=center, draw=c_red] {};
\genrand{c_green}
\node (qpet) [querystyle, above right = 2mm and 2mm of qvos.center, anchor=center, draw=c_green] {};
\genrand{c_purple}
\node (qunknown) [querystyle, below right = 2mm and 2mm of qpet.center, anchor=center, draw=c_purple] {};
\node[inner sep=2pt,fit=(qvis)(qunknown)] (targets) {};

\node[rectangle, draw, c_other, fill=c_other!20, thick, right = 1.8cm of pet.south east, anchor=south west, minimum width=16mm, minimum height=7mm] (tarvis) {\normalsize \ourMethod};

\draw[latex-] (tarvis.150) -- ++(0,2.5mm);
\draw[latex-] (tarvis.30) -- ++(0,2.5mm);
\draw[-latex] (tarvis.south) -- ++(0,-2.5mm);

\node[anchor = center, inner sep=0, text width=2.8cm, inner sep=0pt] (before) at ($(vps.center)!0.5!(vos.center)+(0,2.8cm)$) {\centering \normalsize \bf{BEFORE \Sadey \\[2pt] \small Task-specific models}};
\node[anchor = center, inner sep=0, text width=2.8cm] (now) at ($(tarvis.center)+(0,2.7cm)$) {\centering \normalsize \bf{NOW \Smiley \\[2pt] \small Task-specific targets}};

\draw[thick, gray]
  let
    \p1=(before.north),
    \p2=($(pet.south)+(0,-2.5mm)$),
    \p3=($(now.center)!0.5!(before.center)$)
  in
    (\x3,\y1) -- (\x3, \y2);

\node[inner sep=0pt, below = 4.5mm of vis.south west, anchor = north west] (vis_out1) {\includegraphics[width=26.2mm]{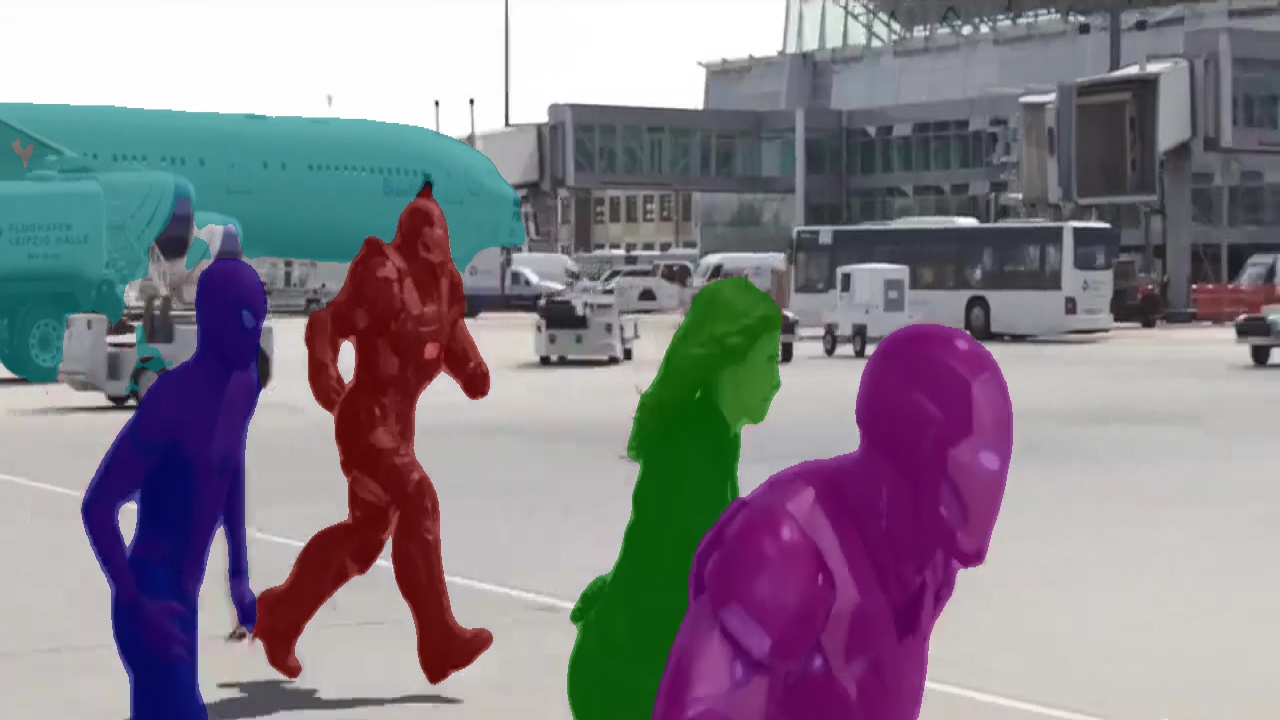}};
\node[inner sep=0pt, right = 0.5mm of vis_out1.north east, anchor = north west] (vis_out2) {\includegraphics[width=26.2mm]{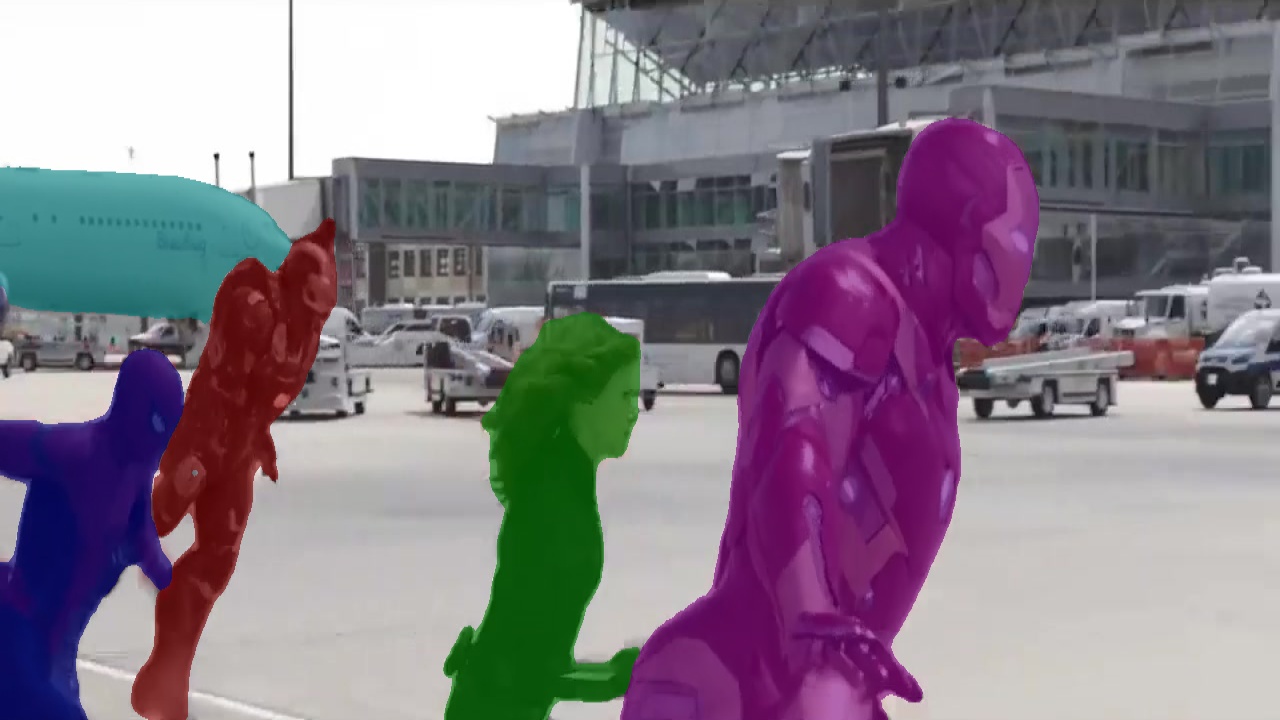}};
\node[inner sep=0pt, right = 0.5mm of vis_out2.north east, anchor = north west] (vis_out3) {\includegraphics[width=26.2mm]{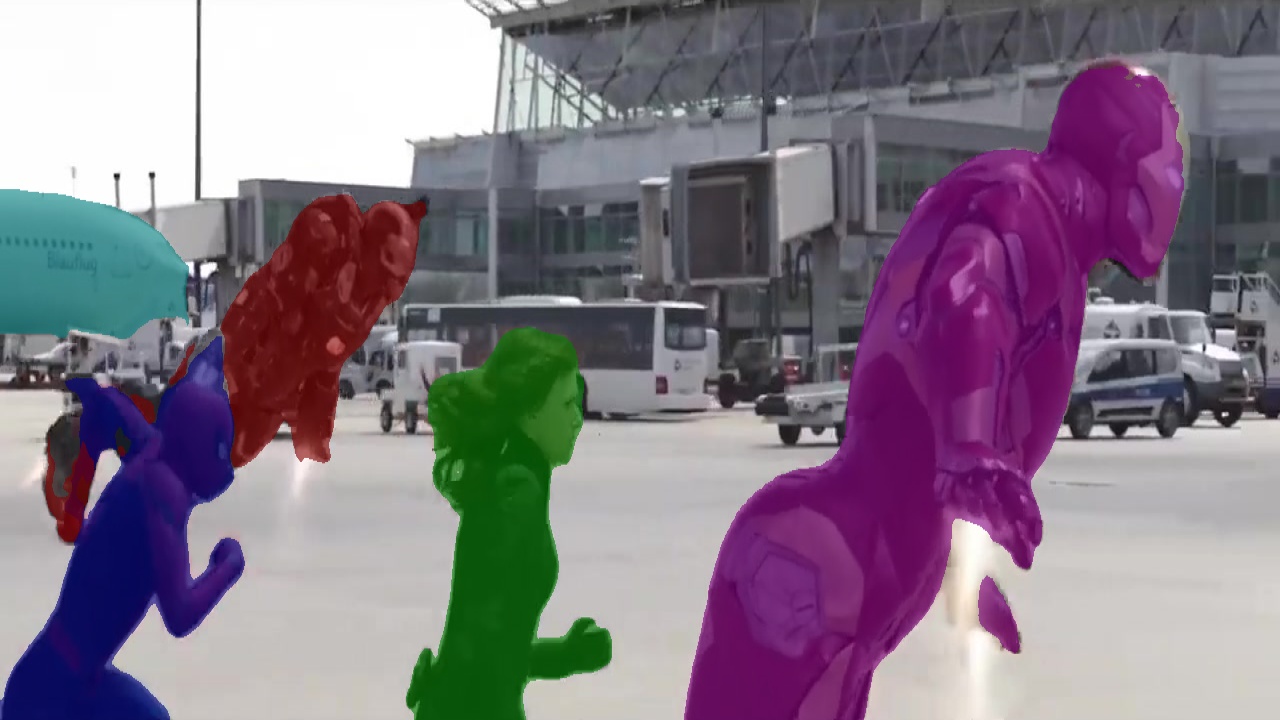}};
\node[c_blue, left = 1mm of vis_out1.west, anchor=south, inner sep=0pt, rotate=90] (vis_seq) {VIS};

\node[inner sep=0pt, below = 1mm of vis_out1.south west, anchor = north west] (vps_out1) {\includegraphics[width=26.2mm,trim={8cm 0cm 7cm 0cm},clip]{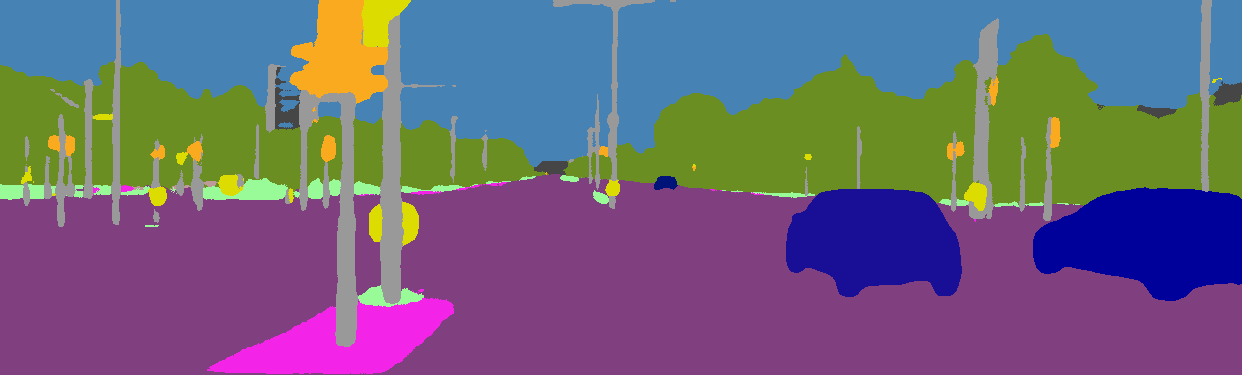}};
\node[inner sep=0pt, right = 0.5mm of vps_out1.north east, anchor = north west] (vps_out2) {\includegraphics[width=26.2mm,trim={8cm 0cm 7cm 0cm},clip]{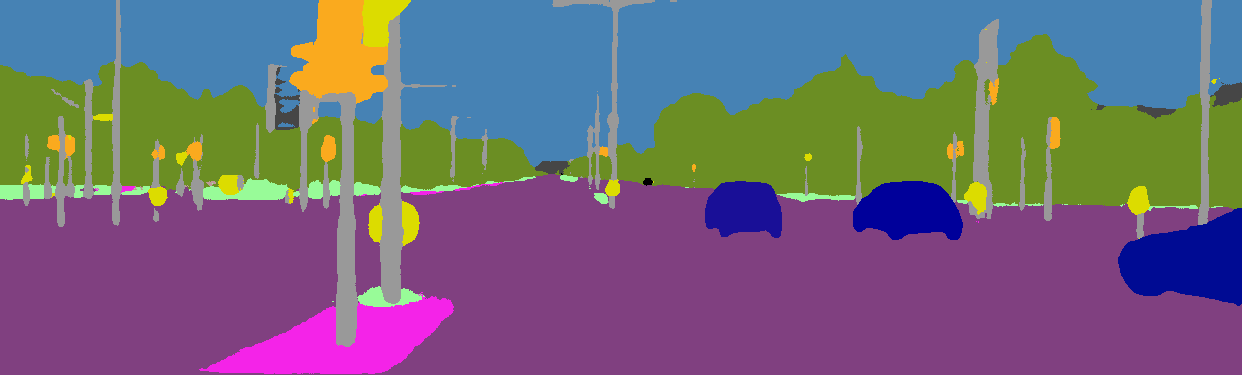}};
\node[inner sep=0pt, right = 0.5mm of vps_out2.north east, anchor = north west] (vps_out3) {\includegraphics[width=26.2mm,trim={8cm 0cm 7cm 0cm},clip]{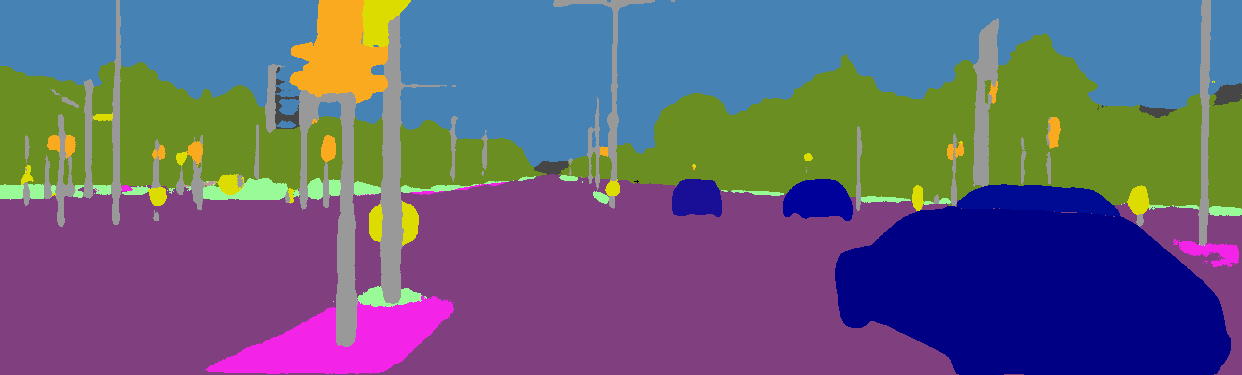}};
\node[c_orange, left = 1mm of vps_out1.west, anchor=south, inner sep=0pt, rotate=90] (vps_seq) {VPS};

\node[inner sep=0pt, below = 1mm of vps_out1.south west, anchor = north west] (vos_out1) {\includegraphics[width=26.2mm]{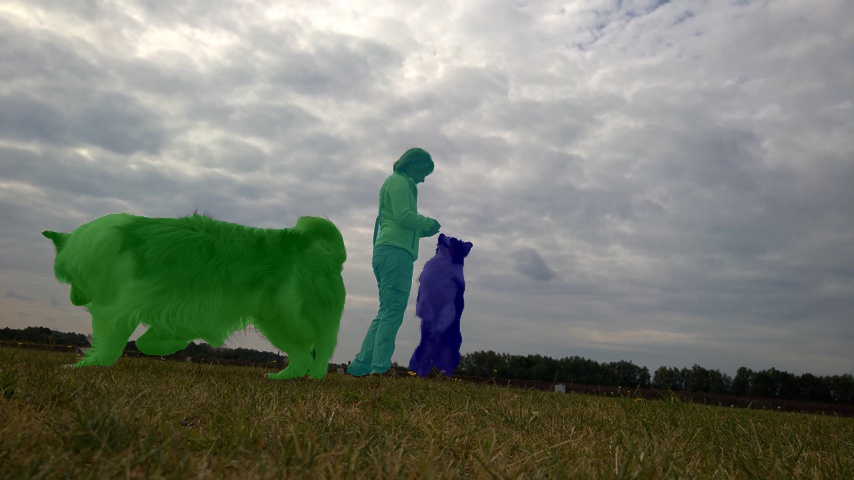}};
\node[inner sep=0pt, right = 0.5mm of vos_out1.north east, anchor = north west] (vos_out2) {\includegraphics[width=26.2mm]{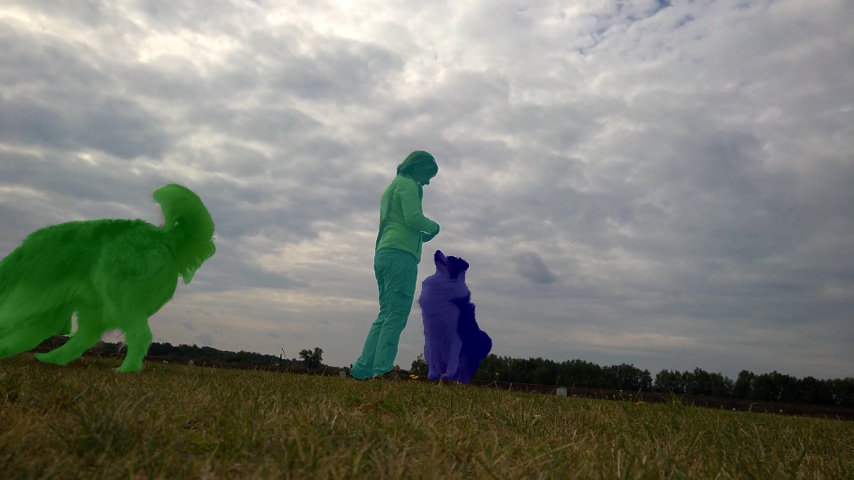}};
\node[inner sep=0pt, right = 0.5mm of vos_out2.north east, anchor = north west] (vos_out3) {\includegraphics[width=26.2mm]{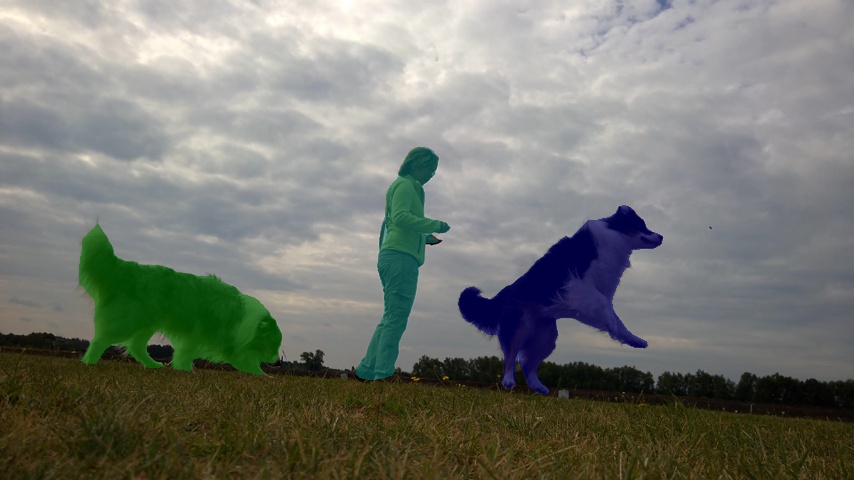}};
\node[left = 1mm of vos_out1.west, anchor=south, inner sep=0pt, rotate=90] (vos_seq) {\textcolor{c_red}{VOS}\textcolor{gray}{/}\textcolor{c_green}{PET}};

\end{tikzpicture}
\caption{Predicted results from a jointly trained \ourMethod{} model for four different video segmentation tasks.}
\label{fig:teaser}
\end{figure}

The ability to understand video scenes has been a long-standing goal of computer vision research because of wide-ranging applications in intelligent vehicles and robots. 
Early approaches tackled simpler tasks involving contour-based~\cite{kass1988snakes,koller1994robust} and box-level tracking~\cite{grabner2006BMVC,leibe2008oldMOT,ess2009oldMOT2,pirsiavash2011oldMOT3}, background subtraction~\cite{elgammal2000nonparam,stauffer1999adaptive}, and motion segmentation~\cite{brox2010fbms,ochs2013fbms2}. The deep learning boom then revolutionized the landscape by enabling methods to perform pixel-precise segmentation on challenging, real-world videos.
In the past few years, a number of benchmarks have emerged, which evaluate how well methods can perform video segmentation according to various task formulations.
Over time, these tasks/benchmarks have ballooned into separate research sub-communities. Although existing methods are rapidly improving the state-of-the-art for these benchmarks, each of them typically tackles only one narrowly-defined task, and generalizing them is non-trivial since the task definition is baked into the core approach. 

We argue that this fragmentation is unnecessary because video target segmentation tasks all require the same high-level capability, namely that of identifying, localizing and tracking rich semantic concepts. 
Meanwhile, recent progress on Transformer networks has enabled the wider AI research community to move towards unified, multi-task architectures~\cite{jaegle2021perceiver,jaegle2021perceiverio,reed2022GATO,alayrac2022flamingo,kolesnikov2022uvim}, because the attention operation~\cite{vaswani2017attention} is well-suited for processing feature sets with arbitrary structure and data modality. These developments give us the opportunity to unify the fractured landscape of target-based video segmentation.
In this paper, we propose \ourMethod{}: a novel architecture which enables a single, unified model to be \emph{jointly trained for multiple video segmentation tasks}. 
During inference, the \emph{same model can perform different tasks at runtime} by specifying the segmentation target.

The core idea is that \ourMethod{} tackles the generic task of segmenting a set of arbitrary \emph{targets} in video (defined as semantic classes or as specific objects). These targets are encoded as \emph{queries} which, together with the video features, are input to a Transformer-based model. The model iteratively refines these queries and produces a pixel-precise mask for each target entity. This formulation conceptually fuses all video segmentation tasks~\cite{yang2019youtubevis,qi2022ovis,weber2021kittistep,athar2023burst} which fall under the umbrella of the above-mentioned generic task, because they differ only in how the \emph{targets} are defined. During both training and inference, \ourMethod{} can hot-swap between tasks at run-time by providing the desired target query set.

To demonstrate our generalization capability, we tackle four different tasks: (1) Video Instance Segmentation (VIS)~\cite{yang2019youtubevis,qi2022ovis}, (2) Video Panoptic Segmentation (VPS)~\cite{kim2020vps}, (3) Video Object Segmentation~\cite{pont2017davis}, and (4) Point Exemplar-guided Tracking~\cite{athar2023burst} (PET). For VIS, the segmentation targets are all objects in the video belonging to a predefined set of classes. The target set for VPS includes that for VIS, and additionally, a set of non-instantiable \emph{stuff} semantic classes. For VOS, the targets are a specific set of objects for which the first-frame ground-truth mask is provided. PET is a more constrained version of VOS which only provides the location of a single point inside the object, rather than the full object mask.

Existing methods for these tasks lack generalization capability because task-specific assumptions are typically baked into the approach (see Sec.~\ref{sec:related_work} and \ref{sec:method} for details).
In contrast, \ourMethod{} can tackle all four tasks with a unified model because we encode the task-specific targets as a set of queries, thus decoupling the network architecture from the task definition. 
Moreover, our approach can theoretically generalize further, \eg, one could potentially define the target set as all objects described by a given text prompt, though this is beyond the scope of this paper.

To summarize, our contributions are as follows: we propose \ourMethod{}, a novel architecture that can perform any task requiring segmentation of a set of \emph{targets} from video. For the first time, we are able to jointly train and infer a single model on a collection of datasets spanning the four aforementioned tasks (VIS, VPS, VOS, PET). Our experimental results show that \ourMethod{} performs competitively for VOS, and achieves state-of-the-art results for VIS, VPS and PET.

\section{Related Work}
\label{sec:related_work}

\PAR{Multi-task Models.}
Multi-task learning has a long history~\cite{Caruana93ICML} with several architectures and training strategies~\cite{Kokkinos17CVPR,Ruder17Arxiv,Zamir18CVPR,Kirillov19panoptic, Pfeiffer19GCPR, Ghiasi21CVPR}.
Earlier approaches mostly consist of a shared backbone with fixed task-specific heads, whereas we design a more general architecture for video segmentation with task-specific targets to specify what to segment.
Our approach is inspired by recent attention-based models, \eg, PerceiverIO~\cite{jaegle2021perceiver,jaegle2021perceiverio}, which can be trained on diverse data modalities and task-specific heads are replaced with output queries.
UViM~\cite{kolesnikov2022uvim} follows a similar direction by creating a unified architecture for diverse dense prediction tasks. However, both of these models are trained separately for different tasks.
Recent, powerful multi-task vision language models such as Flamingo~\cite{alayrac2022flamingo} and GATO~\cite{reed2022GATO} tackle a multitude of tasks by requiring a sequence of task-specific input-output examples to prime the model.
This is conceptually similar to our task-specific targets, however, our model does not require per-task priming. Moreover, our targets are not modeled as sequence prompts, and we aim for a video segmentation model which is several orders of magnitude smaller.
In the realm of video tracking and segmentation, the recently proposed UNICORN~\cite{yan2022unicorn} model tackles multiple object tracking-related tasks with a unified architecture.
Unlike \ourMethod{}, however, UNICORN follows the task-specific output head approach and is generally oriented towards box-level tracking tasks~\cite{milan2016mot16,fan2019lasot,muller2018trackingnet,yu2020bdd100k}, thus requiring non-trivial modifications to tackle VPS or PET.

\begin{figure*}
\centering

\colorlet{vos_text}{c_red}
\colorlet{vos_bg}{vos_text!20}
\colorlet{visvps_text}{c_blue}
\colorlet{visvps_bg}{visvps_text!20}
\colorlet{obj_query}{vos_text}
\colorlet{bg_query}{black}
\colorlet{sem_query}{visvps_text}
\colorlet{inst_query}{c_orange}
\colorlet{network_col}{c_green}
\newcommand{\dotprodsymbol}{$\times$}

\begin{tikzpicture}[font=\footnotesize]

\genrand{sem_query}
\node (qsem1) [querystyle, sem_query] {};
\genrand{sem_query}
\node (qsem2) [querystyle, below left= 1mm and 1mm of qsem1.center, anchor=center, sem_query] {};
\genrand{sem_query}
\node (qsem3) [querystyle, below left= 4mm and 4mm of qsem2.center, anchor=center, sem_query] {};
\genrand{sem_query}
\node (qsem4) [querystyle, below left= 1mm and 1mm of qsem3.center, anchor=center, sem_query] {};
\genrand{sem_query}
\node (qsem5) [querystyle, below left= 1mm and 1mm of qsem4.center, anchor=center, sem_query] {};
\draw [dotted, line width=0.4mm, gray] (qsem2.center)+(-1.2mm,-1.2mm) -- +(-3mm,-3mm);
\node[fit=(qsem1)(qsem5)](qsemgroup){};
\node (qsem) [below=-1mm of qsemgroup] {$Q_\text{sem}$};

\genrand{inst_query}
\node (q1) [querystyle, right = 12mm of qsem1.center, anchor=center, inst_query] {};
\genrand{inst_query}
\node (q2) [querystyle, below left= 1mm and 1mm of q1.center, anchor=center, inst_query] {};
\genrand{inst_query}
\node (q3) [querystyle, below left= 4mm and 4mm of q2.center, anchor=center, inst_query] {};
\genrand{inst_query}
\node (q4) [querystyle, below left= 1mm and 1mm of q3.center, anchor=center, inst_query] {};
\genrand{inst_query}
\node (q5) [querystyle, below left= 1mm and 1mm of q4.center, anchor=center, inst_query] {};
\draw [dotted, line width=0.4mm, gray] (q2.center)+(-1.2mm,-1.2mm) -- +(-3mm,-3mm);
\node[fit=(q1)(q5)](qgroup){};
\node (qinst) [below=-1mm of qgroup] {$Q_\text{inst}$};

\node (qbg_dummy) [querystyle, right = 12mm of q1.center, anchor=center, draw=none, rectangle split part fill=none] {};
\genrand{bg_query}
\node (qbg1) [querystyle, right = 12mm of qgroup.center, anchor=center] {};
\node (qbg) [right = 12mm of qinst.center, anchor=center] {$Q_\text{bg}$};

\node (visvpslabel) [inner xsep=0pt, above=14mm of qsem5.west, anchor=west, visvps_text] {VIS or VPS};

\node[inner sep=2pt,fit=(qsem)(qbg_dummy)(visvpslabel)] (visvps) {};
\scoped[on background layer]{
      \draw [fill=visvps_bg, draw=visvps_bg] (visvps.south west) rectangle (visvps.north east);}

\node[inner sep=0pt, below = 22mm of qsem5.west, anchor=west] (vos_gt) {\includegraphics[width=15mm]{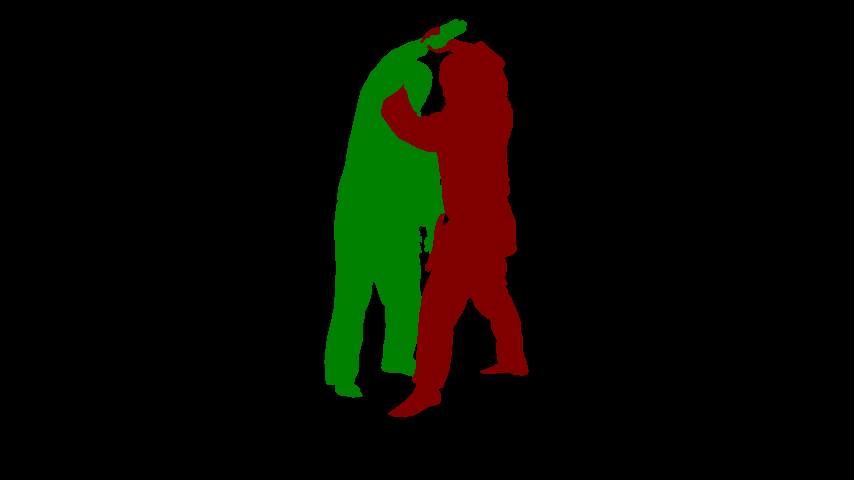}};

\node[inner sep=0pt, below = 10mm of vos_gt.west, anchor=west] (pvos_gt) {\includegraphics[width=15mm]{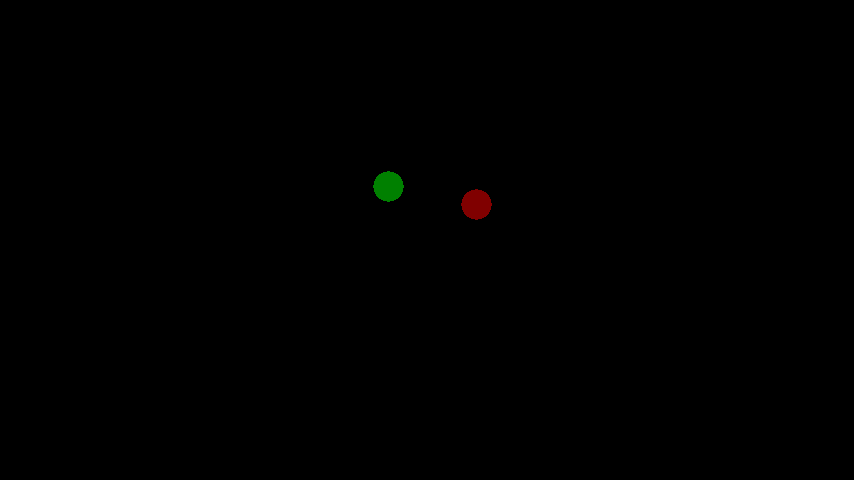}};

\coordinate (betweenims) at ( $ (vos_gt.center)!0.5!(pvos_gt.center) $ );

\node (object_encoder) [right = 12mm of betweenims, anchor=south, trapezium, trapezium angle=86, minimum width=6mm, minimum height=15mm, rotate=-90, inner xsep=1.5pt, fill=vos_text!30, draw=vos_text!60, thick] {};
\node (object_encoder_text) [anchor=center,align=center]  at (object_encoder.center)  {$\mathtt{EncObj}$};

\draw [-latex] ($(object_encoder.west)+(0,2.5mm)$) -- (object_encoder.west) node[pos=0, left] {$\mathcal{F}$};
\draw [dotted] ($(object_encoder.west)+(0,2mm)$) -- ($(object_encoder.west)+(0,5mm)$);

\path (vos_gt.east) edge[-latex] [out=0, in=180] ($(object_encoder.south) + (0,2mm)$);
\path (pvos_gt.east) edge[-latex] [out=0, in=180] ($(object_encoder.south) + (0,-2mm)$);
\path (object_encoder.north) edge[-latex] ++(4mm,0);

\genrand{obj_query}
\node (qvsem1) [querystyle, above right= 3mm and 28mm of object_encoder.south, anchor=center, obj_query] {};
\genrand{obj_query}
\node (qvsem2) [querystyle, below left= 1mm and 1mm of qvsem1.center, anchor=center, obj_query] {};
\genrand{obj_query}
\node (qvsem3) [querystyle, below left= 4mm and 4mm of qvsem2.center, anchor=center, obj_query] {};
\genrand{obj_query}
\node (qvsem4) [querystyle, below left= 1mm and 1mm of qvsem3.center, anchor=center, obj_query] {};
\genrand{obj_query}
\node (qvsem5) [querystyle, below left= 1mm and 1mm of qvsem4.center, anchor=center, obj_query] {};
\draw [dotted, line width=0.4mm, gray] (qvsem2.center)+(-1.2mm,-1.2mm) -- +(-3mm,-3mm);
\node[fit=(qvsem1)(qvsem5)](qvsemgroup){};
\node (qvsem) [below=-1mm of qvsemgroup] {$Q_\text{obj}$};

\genrand{bg_query}
\node (qvbg1) [querystyle, right = 12mm of qvsem1.center, anchor=center] {};
\genrand{bg_query}
\node (qvbg2) [querystyle, below left= 1mm and 1mm of qvbg1.center, anchor=center] {};
\genrand{bg_query}
\node (qvbg3) [querystyle, below left= 4mm and 4mm of qvbg2.center, anchor=center] {};
\genrand{bg_query}
\node (qvbg4) [querystyle, below left= 1mm and 1mm of qvbg3.center, anchor=center] {};
\genrand{bg_query}
\node (qvbg5) [querystyle, below left= 1mm and 1mm of qvbg4.center, anchor=center] {};
\draw [dotted, line width=0.4mm, gray] (qvbg2.center)+(-1.2mm,-1.2mm) -- +(-3mm,-3mm);
\node[fit=(qvbg1)(qvbg5)](qvbggroup){};
\node (qvbg) [below=-1mm of qvbggroup] {$Q_\text{bg}$};

\node (voslabel) [inner xsep=0pt, below=29mm of visvpslabel.west, anchor=west, vos_text] {VOS or PET};

\node[inner sep=2pt,fit=(qvbg1)(qvsem)(voslabel)] (vos) {};
\scoped[on background layer]{
   \draw [fill=vos_bg,draw=vos_bg] (vos.south west) rectangle (vos.north east);}

\node[inner sep=0pt, above right = 6mm and 8mm of visvps.east, anchor=west] (rgb_in3) {\includegraphics[width=15mm]{figures/arch_ims/00000_rgb.jpg}};
\node[inner sep=0pt, below left = 2mm and 2mm of rgb_in3.center, anchor=center] (rgb_in2) {\includegraphics[width=15mm]{figures/arch_ims/00000_rgb.jpg}};
\node[inner sep=0pt, below left = 2mm and 2mm of rgb_in2.center, anchor=center] (rgb_in1) {\includegraphics[width=15mm]{figures/arch_ims/00000_rgb.jpg}};
\node[inner sep=2pt,fit=(rgb_in3)(rgb_in1)] (rgb_in) {};

\node (backbone) [right = 3mm of rgb_in, anchor=south, trapezium, trapezium angle=82, minimum width=16.7mm, minimum height=15mm, rotate=-90, inner xsep=2.5pt, fill=network_col!20, draw=network_col!60, thick] {};
\node (backbone_text) [anchor=center,align=center]  at (backbone.center)  {Backbone};

\draw[-latex] (rgb_in.east) -- (backbone.south);
\draw[-latex] (backbone.north) -- ++(3mm,0mm);

\node (neck) [right = 26mm of backbone.north, anchor=north, trapezium, trapezium angle=82, minimum width=16.7mm, minimum height=15mm, rotate=90, inner xsep=2.5pt, fill=network_col!20, draw=network_col!60, thick] {};
\node (neck_text) [anchor=center,align=center]  at (neck.center)  {Temporal \\Neck};

\draw[latex-] (neck.north) -- ++(-3mm,0mm);

\coordinate (bottleneck) at ($(backbone.north)!0.5!(neck.north)$);
\foreach \x in {-0.7,-0.6,...,0.4} {
  \foreach \y in {-0.4,-0.3,...,.1} {
      \pgfmathsetmacro\col{50*rnd+50*rnd+50*rnd+50*rnd-70}
      \node[fill=network_col!\col,inner sep=0.5mm,outer sep=0pt,anchor=center, draw=c_green!60!black, very thin] at ($(\x,\y) + (bottleneck) + (3.9mm,3.7mm)$) {}; 
  }
}
\foreach \x in {-0.7,-0.6,...,0.4} {
  \foreach \y in {-0.4,-0.3,...,.1} {
      \pgfmathsetmacro\col{50*rnd+50*rnd+50*rnd+50*rnd-70}
      \node[fill=network_col!\col,inner sep=0.5mm,outer sep=0pt,anchor=center, draw=c_green!60!black, very thin] at ($(\x,\y) + (bottleneck) + (1.5mm,1.4mm)$) {}; 
  }
}
\foreach \x in {-0.7,-0.6,...,0.4} {
  \foreach \y in {-0.4,-0.3,...,.1} {
      \pgfmathsetmacro\col{50*rnd+50*rnd+50*rnd+50*rnd-70}
      \node[fill=network_col!\col,inner sep=0.5mm,outer sep=0pt,anchor=center, draw=c_green!60!black, very thin] at ($(\x,\y) + (bottleneck) - (0.8mm,0.9mm)$) {}; 
  }
}

\node (tf_dec) [below  = 32mm of neck.south, anchor=east, rectangle, minimum width=49mm, minimum height=33mm, fill=network_col!20, draw=network_col!60, thick] {};
\node () [anchor = south] at (tf_dec.south) {Transformer Decoder};

\draw[-latex,rounded corners] (visvps.-25) - ++(31mm,0) |- (tf_dec.west);

\node (tf_dec_layer1) [rounded corners, below right= 1mm and 1mm of tf_dec.north west, anchor=north west, rectangle, minimum width=15.3mm, minimum height=27mm, fill=network_col!40, draw=network_col!80, thick] {};
\node () [anchor = south east] at (tf_dec_layer1.south east) {\scriptsize Layer 1};

\node (mca) [rounded corners, below right = 1mm and 1mm of tf_dec_layer1.north west, anchor=north east, rectangle, minimum width=21mm, rotate=90, fill=network_col!60, draw=network_col, thick] {\tiny Masked Cross-Attention};
\node (sa) [rounded corners, right = 1mm of mca.south, anchor=north, rectangle, minimum width=21mm, rotate=90, fill=network_col!60, draw=network_col, thick] {\tiny \tiny Self-Attention};
\node (ffn) [rounded corners, right = 1mm of sa.south, anchor=north, rectangle, minimum width=21mm, rotate=90, fill=network_col!60, draw=network_col, thick] {\tiny \tiny FFN};

\node (tf_dec_layer2) [rounded corners, right= 2mm of tf_dec_layer1.north east, anchor=north west, rectangle, minimum width=11mm, minimum height=27mm, fill=network_col!40, draw=network_col!80, thick] {};
\node () [anchor = south east] at (tf_dec_layer2.south east) {\scriptsize Layer 2};

\node (tf_dec_layerL) [rounded corners, below left= 1mm and 1mm of tf_dec.north east, anchor=north east, rectangle, minimum width=11mm, minimum height=27mm, fill=network_col!40, draw=network_col!80, thick] {};
\node () [anchor = south east] at (tf_dec_layerL.south east) {\scriptsize Layer $L$};

\draw[-latex]
  let
    \p1=(vos.east),
    \p2=($(tf_dec.west)$)
  in
    (\x1,\y2) -- (\x2, \y2) node [pos=0.7, above] {$Q_\text{in}$};

\draw[-latex]
  let
    \p1=(tf_dec_layer1.east),
    \p2=(tf_dec.west),
    \p3=(tf_dec_layer2.west)
  in
    (\x1,\y2) -- (\x3, \y2);
    
\draw[-latex]
  let
    \p1=(tf_dec_layer2.east),
    \p2=(tf_dec.west),
  in
    (\x1,\y2) -- ($(\x1, \y2) + (2mm,0)$);
    
\draw[-latex]
  let
    \p1=(tf_dec_layerL.west),
    \p2=(tf_dec.west),
  in
    ($(\x1, \y2) - (2mm,0)$) -- (\x1,\y2);
    
\draw[dotted]
  let
    \p1=(tf_dec_layer2.east),
    \p2=(tf_dec.west),
    \p3=(tf_dec_layerL.west)
  in
    ($(\x1, \y2) + (2mm,0)$) -- ($(\x3, \y2) - (2mm,0)$);

\draw[-latex,rounded corners]
  let
    \p1=(neck.190),
    \p2=(sa.east)
  in
    (\x1,\y1) -- node [pos=0.5, left] {$\mathcal{F}$} ($(\x1, \y1) - (0,4mm)$) -| (\x2, \y2);

\draw[-latex,rounded corners]
  let
    \p1=(neck.190),
    \p2=(tf_dec_layer2.north)
  in
    ($(\x1, \y1) - (2mm,4mm)$) -| (\x2, \y2);

\draw[-latex]
  let
    \p1=(neck.190),
    \p2=(tf_dec_layerL.north)
  in
    (\x1,\y1) -- (\x1, \y2);

\draw[] (neck.south) edge ($(neck.south) + (2mm,0)$) edge [dotted] ($(neck.south) + (5mm,0)$) node [above] {\hspace{14pt}$F_4$};

\draw[] (tf_dec.east) edge ($(tf_dec.east) + (2mm,0)$) edge [dotted] ($(tf_dec.east) + (5mm,0)$) node [above] {\hspace{18pt}$Q_\text{out}$};

\node[inner sep=0pt, above right = 4.5mm and 38mm of neck.south, anchor=west] (vis_out3) {\includegraphics[width=15mm]{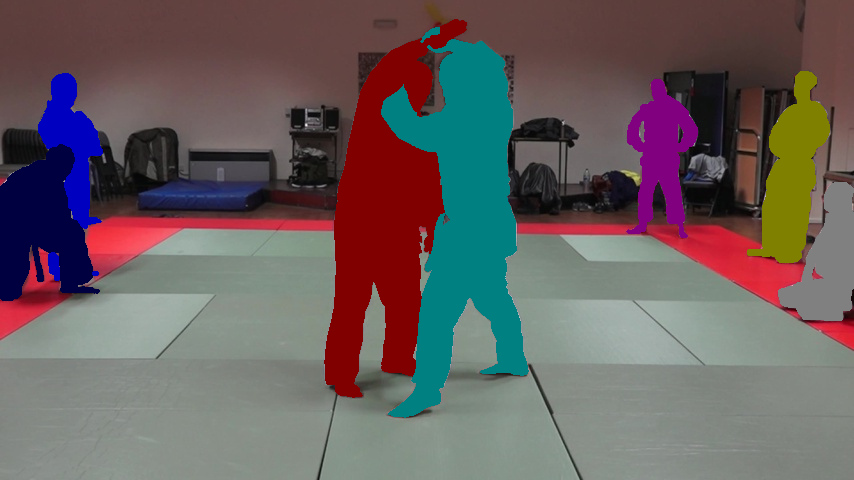}};
\node[inner sep=0pt, below left = 2mm and 2mm of vis_out3.center, anchor=center] (vis_out2) {\includegraphics[width=15mm]{figures/arch_ims/instance.png}};
\node[inner sep=0pt, below left = 2mm and 2mm of vis_out2.center, anchor=center] (vis_out1) {\includegraphics[width=15mm]{figures/arch_ims/instance.png}};
\node[inner sep=2pt,fit=(vis_out3)(vis_out1)] (vis_out) {};

\node[inner sep=0pt, left = 5mm of vis_out1.west, anchor=center, circle, draw=black, fill=white] (dot_vis) {\dotprodsymbol};
\draw[-latex] (dot_vis.east) -- (vis_out1.west);
\draw[-latex] ($(dot_vis.west) - (15mm,0)$) -- (dot_vis.west) node [midway,above, inner sep=1pt] {$Q'_\text{inst}$};
\draw[dotted] ($(dot_vis.west) - (15mm,0)$) -- ($(dot_vis.west) - (18mm,0)$);
\draw[-latex,rounded corners] ($(dot_vis.west) - (15mm,-5mm)$) -| (dot_vis.north) node [pos=0.25,above, inner sep=1pt] {$F_4$};
\draw[dotted] ($(dot_vis.west) - (15mm,-5mm)$) -- ($(dot_vis.west) - (18mm,-5mm)$);

\node[inner sep=0pt, below = 16mm of vis_out3.west, anchor=west] (vps_out3) {\includegraphics[width=15mm]{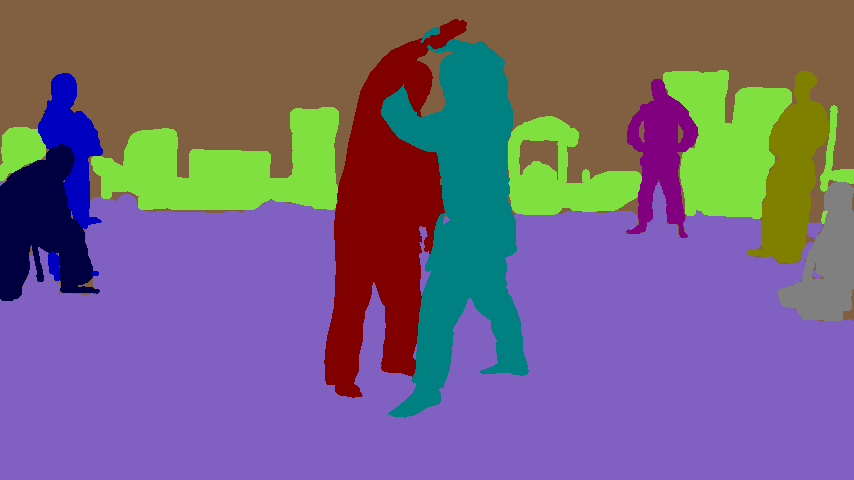}};
\node[inner sep=0pt, below left = 2mm and 2mm of vps_out3.center, anchor=center] (vps_out2) {\includegraphics[width=15mm]{figures/arch_ims/panoptic.png}};
\node[inner sep=0pt, below left = 2mm and 2mm of vps_out2.center, anchor=center] (vps_out1) {\includegraphics[width=15mm]{figures/arch_ims/panoptic.png}};
\node[inner sep=2pt,fit=(vps_out3)(vps_out1)] (vps_out) {};

\node[inner sep=0pt, left = 5mm of vps_out1.west, anchor=center, circle, draw=black, fill=white] (dot_vps) {\dotprodsymbol};
\draw[-latex] (dot_vps.east) -- (vps_out1.west);
\draw[-latex] ($(dot_vps.west) - (15mm,0)$) -- (dot_vps.west) node [midway,above, inner sep=1pt] {$Q'_\text{inst},Q'_\text{sem}$};
\draw[dotted] ($(dot_vps.west) - (15mm,0)$) -- ($(dot_vps.west) - (18mm,0)$);
\draw[-latex,rounded corners] ($(dot_vps.west) - (15mm,-5mm)$) -| (dot_vps.north) node [pos=0.25,above, inner sep=1pt] {$F_4$};
\draw[dotted] ($(dot_vps.west) - (15mm,-5mm)$) -- ($(dot_vps.west) - (18mm,-5mm)$);

\node[inner sep=0pt, below = 13mm of dot_vps.center, anchor=center, circle, draw=black, fill=white] (dot_cls) {\dotprodsymbol};
\path
    let
        \p1=(dot_cls.east),
        \p2=(vps_out1.west)
    in
        node at (\x2,\y1) [inner sep=3pt, anchor=west, draw=visvps_text, thick, rounded corners, text=visvps_text] (cls) {Classification};
\draw[-latex] (dot_cls.east) -- (cls.west);
\draw[-latex] ($(dot_cls.west) - (15mm,0)$) -- (dot_cls.west) node [midway,above, inner sep=1pt] {$Q'_\text{sem},Q'_\text{bg}$};
\draw[dotted] ($(dot_cls.west) - (15mm,0)$) -- ($(dot_cls.west) - (18mm,0)$);
\draw[-latex,rounded corners] ($(dot_cls.west) - (15mm,-5mm)$) -| (dot_cls.north) node [pos=0.25,above, inner sep=1pt] {$Q'_\text{inst}$};
\draw[dotted] ($(dot_cls.west) - (15mm,-5mm)$) -- ($(dot_cls.west) - (18mm,-5mm)$);

\coordinate (left_visvps_out_block) at ($(dot_vis.west) - (18mm,0)$);
\node[inner sep=3pt,fit=(cls)(vis_out3)(left_visvps_out_block)] (visvps_out) {};
\scoped[on background layer]{
      \draw [fill=visvps_bg, draw=visvps_bg] (visvps_out.south west) rectangle (visvps_out.north east);}

\node[inner sep=0pt, below = 28mm of vps_out3.west, anchor=west] (vos_out3) {\includegraphics[width=15mm]{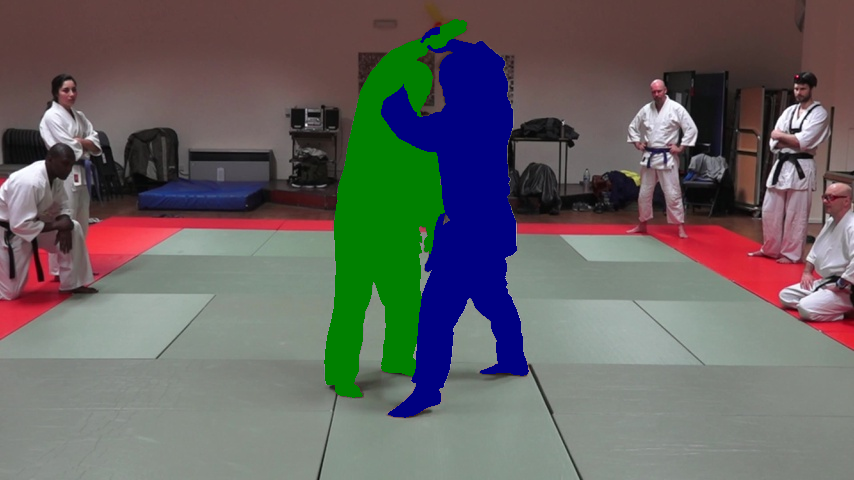}};
\node[inner sep=0pt, below left = 2mm and 2mm of vos_out3.center, anchor=center] (vos_out2) {\includegraphics[width=15mm]{figures/arch_ims/vos.png}};
\node[inner sep=0pt, below left = 2mm and 2mm of vos_out2.center, anchor=center] (vos_out1) {\includegraphics[width=15mm]{figures/arch_ims/vos.png}};
\node[inner sep=2pt,fit=(vos_out3)(vos_out1)] (vos_out) {};

\node[inner sep=0pt, left = 5mm of vos_out1.west, anchor=center, circle, draw=black, fill=white] (dot_vos) {\dotprodsymbol};
\draw[-latex] (dot_vos.east) -- (vos_out1.west);
\draw[-latex] ($(dot_vos.west) - (15mm,0)$) -- (dot_vos.west) node [midway,above, inner sep=1pt] {$Q'_\text{obj}$};
\draw[dotted] ($(dot_vos.west) - (15mm,0)$) -- ($(dot_vos.west) - (18mm,0)$);
\draw[-latex,rounded corners] ($(dot_vos.west) - (15mm,-5mm)$) -| (dot_vos.north) node [pos=0.25,above, inner sep=1pt] {$F_4$};
\draw[dotted] ($(dot_vos.west) - (15mm,-5mm)$) -- ($(dot_vos.west) - (18mm,-5mm)$);

\coordinate (left_vos_out_block) at ($(dot_vos.west) - (18mm,0)$);
\node[inner sep=3pt,fit=(vos_out1)(vos_out3)(left_vos_out_block)] (vos_out) {};
\scoped[on background layer]{
      \draw [fill=vos_bg, draw=vos_bg] (vos_out.south west) rectangle (vos_out.north east);}

\end{tikzpicture}
\caption{\textbf{\ourMethod{} Architecture.} Segmentation targets for different tasks are represented by a set of abstract target queries $Q_\text{in}$. The core network (in green) is agnostic to the task definitions. The inner product between the output queries $Q_\text{out}$ and video feature $F_4$ yields segmentation masks as required by the task.} 
\label{fig:architecture}
\end{figure*}

\PAR{Query-based Transformer Architectures.}
Several works~\cite{carion2020detr,cheng2021mask2former,misra2021_3detr,zhu2020deformdetr,wang2021vistr,athar2022hodor,jaegle2021perceiver,jaegle2021perceiverio} use query-based Transformer architectures for various tasks. The workhorse for task learning here is the iterative application of self- and cross-attention, where a set of query vectors (\eg, representing objects) are refined by interacting with each other, and with the input data (\eg, an image). 
Unlike existing methods which use queries in a task-specific context,
\ourMethod{} adopts a query-based Transformer architecture in which the queries serve as a mechanism for decoupling the task definition from the architecture, \ie, our model can learn to tackle different tasks while being agnostic to their definition because the latter is %
abstracted behind a set of queries.

\PAR{Task-specific Video Segmentation.}
Current Video Instance Segmentation (VIS) methods broadly work by predicting object tracks in the video, followed by classification into a pre-defined set of categories. Several approaches \cite{voigtlaender2019mots,yang2019youtubevis,qi2022ovis,bertasius2020maskprop,cao2020sipmask,fu2021compfeat,ke2021crossvis,wu2022idol,huang2022minvis,lin2021proposereduce} are based on the tracking-by-detection paradigm, some model video as a joint spatio-temporal volume~\cite{athar2020stemseg,athar2021stemseg_ovis}, whereas recent works~\cite{wang2021vistr,wu2021seqformer,hwang2021IFC,cheng2021mask2formervideo,heo2022vita} adopt Transformer-based architectures. %

For Video Panoptic Segmentation (VPS), methods \cite{kim2020vps,qiao2021vipdeeplab,weber2021kittistep} generally extend image-level panoptic approaches~\cite{cheng2020panopticdeeplab} by employing multi-head architectures for semantic segmentation and instance mask regression, classification, and temporal association. In the Video Object Segmentation (VOS) community, state-of-the-art methods are broadly based on the seminal work of Oh~\etal~\cite{Oh19STMVOS}, which learns \emph{space-time correspondences} between pixels in different video frames, and then uses these to propagate the first-frame masks across the video. Subsequent methods~\cite{voigtlaender2019feelvos,Yang20CFBI,xie2021RMVOS,Yang21AOTL,seong2021HMMVOS,Cheng21STCN,cheng2021mivos,cheng2022xmem,Yang20CFBI} have significantly improved the performance and efficiency of this approach. Point Exemplar-guided Tracking (PET)~\cite{athar2023burst,homayounfar2021videoclick} is a fairly new task for which the current best approach~\cite{athar2023burst} regresses a pseudo-ground-truth mask from the given point coordinates, and then applies a VOS method~\cite{Cheng21STCN} to this mask.

The above methods thus incorporate task-specific assumptions into their core approach. This can benefit per-task performance, but makes it difficult for them to generalize across tasks. By contrast, \ourMethod{} can tackle all four aforementioned tasks, and generally any target-based video segmentation task, with a single, %
jointly trained model.

\section{Method}
\label{sec:method}

\ourMethod{} can segment arbitrary targets in video since the architecture is flexible with respect to how these targets are defined, thus enabling us to conceptually unify and jointly tackle the four aforementioned tasks (VIS, VPS, VOS, PET).
The architecture is illustrated in Fig.~\ref{fig:architecture}. 

For all tasks, the common input to the network is an RGB video clip of length $T$ denoted by $V \in \real{H\times W\times T\times 3}$. This is input to a 2D backbone network which produces image-level feature maps, followed by a \emph{Temporal Neck}, which enables feature interaction across time and outputs a set of temporally consistent, multi-scale, $D$-dimensional feature maps $\mathcal{F} = \{F_{32}, F_{16}, F_8, F_4 \}$ where $F_s \in \real{\frac{H}{s}\times \frac{W}{s}\times T\times D}$. 
The feature maps $\mathcal{F}$ are then fed to our Transformer decoder, together with a set of queries $Q_\text{in}$ which represent the segmentation targets.
The decoder applies successive layers of self- and masked cross-attention wherein the queries are iteratively refined by attending to each other, and to the feature maps, respectively. The refined queries output by the decoder are denoted with $Q_\text{out}$. The following subsections explain how \ourMethod{} tackles each task in detail.

\subsection{Video Instance Segmentation}
\label{subsec:method_video_instance_segmentation}

VIS defines the segmentation target set as all objects belonging to a set of predefined classes. Accordingly, the input query set $Q_\text{in}$ for VIS contains three types of queries: (1) \emph{semantic queries} denoted by $Q_\text{sem} \in \real{C\times D}$ where $C$ is the number of classes defined by the dataset, \ie, each $D$-dimensional vector in $Q_\text{sem}$ represents a particular semantic class. (2) \emph{instance queries} denoted by $Q_\text{inst} \in \real{I\times D}$ where $I$ is assumed to be an upper bound on the number of instances in the video clip, and (3) a \emph{background query} denoted by $Q_\text{bg} \in \real{1\times D}$ to capture inactive instance queries.

The three query sets are concatenated, \ie, $Q_\text{in} = \mathtt{concat}(Q_\text{sem}, Q_\text{inst}, Q_\text{bg})$, and input to the Transformer decoder, which refines their feature representation through successive attention layers and outputs a set of queries $Q_\text{out} = \mathtt{concat}(Q'_\text{sem}, Q'_\text{inst}, Q'_\text{bg})$.
These are then used to produce temporally consistent instance mask logits by computing the inner product $\langle F_4 , Q'_\text{inst} \rangle \in \real{H\times W\times T\times I}$. To obtain classification logits, we compute the inner product $\langle Q'_\text{inst}, \mathtt{concat} (Q'_\text{sem}, Q'_\text{bg}) \rangle \in \real{I\times (C+1)}$. 

The three types of queries are initialized randomly at the start of training and optimized thereafter. The instance queries $Q_\text{inst}$ enable us to segment a varying number of objects from the input clip. During training, we apply Hungarian matching between the predicted and ground-truth instance masks to assign instance queries to video instances, and then supervise their predicted masks and classification logits accordingly. 
When training on multiple datasets with heterogeneous classes, the semantic query sets are separately initialized per dataset, but $Q_\text{inst}$ and $Q_\text{bg}$ are shared.

\PAR{Comparison to Instance Segmentation Methods.} Several Transformer-based methods~\cite{carion2020detr,zhu2020deformdetr,wang2021vistr,cheng2021mask2former} for image/video instance segmentation also use queries to segment a variable number of input instances. The key difference to our approach is the handling of object classes: existing works employ only instance queries which are input to a fully-connected layer with a fan-out of $C+1$ to obtain classification (and background) scores. The notion of class-guided instance segmentation is thus baked into the approach.
By contrast, \ourMethod{} is agnostic to the task-specific notion of object classes because it models them as arbitrary queries which are dynamic inputs to the network. The semantic representation for these queries is thus \mbox{decoupled} from the core architecture and is only learned via loss supervision.
An important enabler for this approach is the background query $Q_\text{bg}$, which serves as a `catch-all' class to represent everything that is not in $Q_\text{sem}$. It is used to classify non-active instance queries, and its mask logits are supervised to segment all non-object input pixels.

\subsection{Video Panoptic Segmentation}
\label{subsec:method_video_panoptic_segmentation}

VPS defines the segmentation targets as all objects belonging to a set of \emph{thing} classes (\eg, `person', `car'), and additionally, a set of non-instantiable \emph{stuff} classes (\eg, `sky', `grass') which cover all non-object pixels. \ourMethod{} can tackle VPS with virtually no modification to the workflow in Sec.~\ref{subsec:method_video_instance_segmentation}. We can compute semantic segmentation masks for the input clip by simply taking the inner product between $Q_\text{sem}$ and the video features: $\langle F_4, Q'_\text{sem}\rangle \in \real{H\times W\times T\times C}$. Note that here, $Q_\text{sem}$ contains queries representing both \emph{thing} and \emph{stuff} classes.

\PAR{Comparison to VPS Methods.} 
Current VPS datasets~\cite{weber2021kittistep,kim2020vps} involve driving scene videos captured from moving vehicles. Methods tackling this task~\cite{kim2020vps,qiao2021vipdeeplab} are based on earlier image panoptic segmentation approaches~\cite{cheng2020panopticdeeplab} which involve multi-head networks for semantic and instance segmentation prediction. In terms of image-level panoptic segmentation, Mask2Former~\cite{cheng2021mask2former} uses a Transformer-based architecture, but it models \emph{stuff} classes as instances which are Hungarian-matched to the ground-truth target during training, whereas \ourMethod{} models semantic classes and instances using separate, designated queries.

\subsection{Video Object Segmentation and Point Exemplar-guided Tracking}
\label{subsec:method_video_object_segmentation}

VOS and PET are instantiations of a general task where the segmentation targets are a set of $O$ objects for which some ground-truth cue $\mathcal{G}$ is given.
For VOS, $\mathcal{G}$ is provided as the first-frame object masks $M_\text{obj} \in \real{O\times H\times W}$, whereas for PET, $\mathcal{G}$ is provided as the $(x,y)$ coordinates $P_\text{obj} \in \real{O\times 2}$ of a point inside each of the objects. 
\ourMethod{} jointly tackles both tasks by adopting a generalized approach in which the $O$ target objects are encoded into a set of \emph{object queries} $Q_\text{obj}$.
Thus, both VOS and PET boil down to designing a function $\mathtt{\footnotesize EncodeObjects}(\cdot)$ which regresses $Q_\text{obj}$ from the ground-truth cues $\mathcal{G}$ and feature maps $\mathcal{F}$:
\vspace{-3pt}
\begin{equation}
    Q_\text{obj} \xleftarrow{} \mathtt{EncodeObjects(\mathcal{G}, \mathcal{F})}.
\end{equation}
\vspace{-3pt}

Note that $Q_\text{obj}$ is conceptually analogous to $Q_\text{sem}$ and $Q_\text{inst}$ used for VIS in that all three are abstract representations for their respective task-specific segmentation targets.

\PAR{Video Object Segmentation.} We seek inspiration from HODOR~\cite{athar2022hodor} to implement \texttt{\footnotesize EncodeObjects} for VOS, a recent method for weakly-supervised VOS, which encodes objects into concise \emph{descriptors} as follows: the descriptors are initialized by average pooling the image features inside the object masks, followed by iterative refinement where the descriptors attend to each other (self-attention) and to their respective soft-masked image features (cross-attention).

For \ourMethod{}, we employ a lightweight \emph{Object Encoder} with a similar workflow to encode the objects as a set of queries $Q_\text{obj}$, but with two differences to HODOR~\cite{athar2022hodor}: instead of cross-attending to the entire image feature map ($H\cdot W$ points) with soft-masked attention, we apply hard-masked cross-attention to at most $p_\text{max}$ feature points per object, where $p_\text{max} \ll H\cdot W$. Object masks containing more than $p_\text{max}$ points are sub-sampled accordingly. This significantly improves the memory/run-time overhead of our Object Encoder. Secondly, we note that the process of distilling object features into a single descriptor involves a loss of object appearance information, which degrades performance. We therefore model each object with $q_o$ queries (instead of one) by spatially dividing each object mask into $q_o$ segments, \ie, $Q_\text{obj} \in \real{O\times q_o\times D}$ (we use $q_o = 4$).

In addition to $Q_\text{obj}$, we initialize a set of background queries $Q_\text{bg} \in \real{B\times D}$ to model the non-target pixels in the reference frame. Following HODOR~\cite{athar2022hodor}, we employ multiple background queries, which are initialized dynamically by dividing the video frame containing the ground-truth masks $M_\text{obj}$ into a $4\times 4$ grid and average pooling the non-object pixels in each grid cell. 
The Object Encoder jointly refines the background and object queries to yield $Q_\text{in} = \mathtt{concat}(Q_\text{obj}, Q_\text{bg})$.
During training, the mask logits for the multiple background queries are aggregated per-pixel by applying $\texttt{max}(\cdot)$ and supervised to segment all pixels not part of the target object set.

The remaining workflow follows that for VIS and VPS: $Q_\text{in}$ is input to the Transformer decoder together with the video features $\mathcal{F}$. The refined output query set $Q_\text{out} = \mathtt{concat}(Q_\text{obj}', Q_\text{bg}')$ is then used to compute the inner product $\langle F_4, Q_\text{obj}' \rangle \in \real{H\times W\times T\times O\times q_o}$. Subsequently, $\mathtt{max(\cdot)}$ is applied on the $q_o$-sized dimension to obtain the final mask logits for the $O$ target objects.

\PAR{Point Exemplar-guided Tracking.} 
For PET we implement $\mathtt{EncodeObjects}$ in the exactly same way as VOS: the given point coordinates $P_\text{obj}$ are converted into a mask with just one non-zero pixel, followed by iterative refinement by the Object Encoder (with shared weights for VOS and PET). The only difference is that here we represent each of the $O$ objects with just one query, \ie, $Q_\text{obj} \in \real{O\times D}$ ($q_o=1$). The subsequent workflow is also identical to that for VOS: the queries are refined by the Transformer decoder followed by an inner product with $F_4$ to obtain object mask logits.

\PAR{Comparison to VOS and PET Methods.}
Current state-of-the-art VOS methods are largely based on STM~\cite{Oh19STMVOS}.
It involves learning pixel-to-pixel correspondences across video frames, which are then used to propagate the given object mask across the video.
This approach is effective since every pixel in the given mask can be individually mapped to future frames, thus preserving fine-grained object details.
The core approach is, however, task-specific since it assumes the availability of first-frame object masks, and does not generalize to the PET (see Sec.~\ref{subsec:benchmark_results}). PET can be viewed as a more constrained version of VOS, where only a single object point is provided instead of the full mask.
Consequently, PET~\cite{athar2023burst} is currently tackled by casting it as a VOS problem by using an image instance segmentation network~\cite{cheng2021mask2former} to regress pseudo-ground-truth object masks from the given point coordinates $P_\text{obj}$.

On the other hand, our approach of encoding objects as concise queries causes loss of fine-grained object appearance information, but it has the advantage of being agnostic to how $\mathcal{G}$ is defined. As evident from the unified workflow for VOS and PET, any variation of these tasks with arbitrary ground-truth cues $\mathcal{G}$ can be seamlessly fused into our architecture as long as we can implement an effective $\mathtt{EncodeObjects}$ function to regress $Q_\text{obj}$ from the given $\mathcal{G}$.

\subsection{Network Architecture}
\label{subsec:method_network_architecture}

\begin{table*}[ht]
\centering{}
\setlength{\tabcolsep}{4.0pt}
\newcommand\RotText[1]{\rotatebox{90}{\parbox{2cm}{\centering#1}}}
\small

\newcommand\bd[1]{\textbf{#1}}
\newcommand\fade[1]{\textcolor{gray!90}{#1}}

\caption{
Results for Video Instance Segmentation (VIS) on the YouTube-VIS 2021~\cite{yang2019youtubevis} and OVIS~\cite{qi2022ovis} validation sets.
}
\renewcommand{\arraystretch}{0.95}
\begin{tabularx}{\textwidth}{llcp{0.05cm}YYYYYp{0.05cm}YYYYY}
\toprule 
\multirow{2}{*}{Method} & \multirow{2}{*}{Backbone} & \multirow{2}{1cm}{\hspace{2pt}Shared\\ \hspace{2pt}Model} & & \multicolumn{5}{c}{YouTube-VIS 2021}  & & \multicolumn{5}{c}{OVIS} \\
\cmidrule{5-9} \cmidrule{11-15} 
                                                 &           &        & & AP   & AP50 & AP75 & AR1  & AR10   & &  AP  & AP50 & AP75 & AR1  & AR10 \\
\midrule
Mask2Former-VIS~\cite{cheng2021mask2formervideo} & R-50        & \xmarkr &  & 40.6 & 60.9 & 41.8 &  -   &  -    & &  -   &  -   &  -   &  -   &  -   \\
IDOL~\cite{wu2022idol}                           & R-50        & \xmarkr &  & 43.9 & 68.0 & 49.6 & 38.0 & 50.9  & & 30.2 & 51.3 & 30.0 & 15.0 & 37.5 \\
MinVIS~\cite{huang2022minvis}                    & R-50        & \xmarkr &  & 44.2 & 66.0 & 48.1 & 39.2 & 51.7  & & 25.0 & 45.5 & 24.0 & 13.9 & 29.7 \\
VITA~\cite{heo2022vita}                          & R-50        & \xmarkr &  & 45.7 & 67.4 & 49.5 & \bd{40.9} & 53.6  & & 19.6 & 41.2 & 17.4 & 11.7 & 26.0 \\
\textbf{\ourMethod{}}                     & R-50        & \cmarkg &  & \bd{48.3} & \bd{69.6} & \bd{53.2} & 40.5 & \bd{55.9}  & & \bd{31.1} & \bd{52.5} & \bd{30.4} & \bd{15.9} & \bd{39.9} \\
\midrule
Mask2Former-VIS~\cite{cheng2021mask2formervideo} & Swin-T      & \xmarkr &  & 45.9 & 68.7 & 50.7 &  -   &  -    & &  -   &  -   &  -   &  -   &  -    \\
\textbf{\ourMethod{}}                     & Swin-T      & \cmarkg &  & \bd{50.9} & \bd{71.6} & \bd{56.6} & 42.2 & 57.2  & & 34.0 & 55.0 & 34.4 & 16.1 & 40.9 \\
\midrule
IDOL~\cite{wu2022idol}                           & Swin-L      & \xmarkr &  & 56.1 & 80.8 & 63.5 & 45.0 & 60.1  & & 42.6 & 65.7 & \bd{45.2} & 17.9 & 49.6  \\
VITA~\cite{heo2022vita}                          & Swin-L      & \xmarkr &  & 57.5 & 80.6 & 61.0 & 47.7 & 62.6  & & 27.7 & 51.9 & 24.9 & 14.9 & 33.0 \\
\bd{\ourMethod{}}                         & Swin-L      & \cmarkg &  & \bd{60.2} & \bd{81.4} & \bd{67.6} & \bd{47.6} & \bd{64.8}  & & \bd{43.2} & \bd{67.8} & 44.6 & \bd{18.0} & \bd{50.4} \\

\bottomrule 
\end{tabularx}

\label{tab:benchmark_results_vis}
\end{table*}

\begin{table*}
\centering{}
\setlength{\tabcolsep}{4.0pt}
\newcommand\RotText[1]{\rotatebox{90}{\parbox{2cm}{\centering#1}}}
\small

\newcommand\bd[1]{\textbf{#1}}
\renewcommand{\arraystretch}{0.95}

\caption{
Video Panoptic Segmentation (VPS) results for validation sets of KITTI-STEP~\cite{weber2021kittistep}, CityscapesVPS~\cite{kim2020vps} and VIPSeg~\cite{miao2022vipseg}.
}

\renewcommand{\arraystretch}{0.95}

\begin{tabularx}{\textwidth}{lcYYYp{0.05cm}YYYp{0.05cm}YYYY}
\toprule 
\multirow{2}{*}{Method}      & \multirow{2}{1cm}{\hspace{2pt}Shared\\ \hspace{2pt}Model} & \multicolumn{3}{c}{KITTI-STEP} & & \multicolumn{3}{c}{CityscapesVPS} & & \multicolumn{4}{c}{VIPSeg} \\
\cmidrule{3-5} \cmidrule{7-9} \cmidrule{11-14}
                             & & STQ  & AQ   & SQ   & & VPQ  & $\text{VPQ}^\text{Th}$ & $\text{VPQ}^\text{St}$ & & VPQ  & $\text{VPQ}^\text{Th}$ & $\text{VPQ}^\text{St}$ & STQ \\ 
\midrule
Mask Propagation~\cite{weber2021kittistep} & \xmarkr & 0.67 & 0.63 & 0.71 & &  -   &   -  &   -    & &  -   &  -    &   -   &    \\
Track~\cite{kim2020vps}      & \xmarkr &  -   &  -   &  -        & & 55.9 & 43.7 & 64.8   & &  -   &   -   &   -   &    \\
VPSNet~\cite{kim2020vps}     & \xmarkr & 0.56 & 0.52 & 0.61      & & 57.0 & 44.7 & 66.0   & & 14.0  & 14.0  & 14.2 & 20.8 \\
VPSNet-SiamTrack~\cite{woo2021vpsnet_siamtrack} & \xmarkr & - & - & - & & 57.3 & 44.7 & 66.4 & & 17.2 & 17.3 & 17.3 & 21.1 \\
VIP-Deeplab~\cite{qiao2021vipdeeplab}      & \xmarkr &  -   &   -  &  -   & & \bd{63.1} & \bd{49.5} & \bd{73.0} & & 16.0 & 12.3 & 18.2 & 22.0 \\
Clip-PanoFCN~\cite{miao2022vipseg}         & \xmarkr &   -   &   -  &  -  & &  -   &  -   &  -   & & 22.9 & 25.0 & 20.8 & 31.5 \\
\bd{\ourMethod{} (R-50)} & \cmarkg & 0.70 & 0.70 & 0.69     & & 53.3 & 35.9 & 66.0 & & 33.5 & 39.2 & 28.5 & 43.1 \\
\bd{\ourMethod{} (Swin-T)} & \cmarkg & 0.71 & 0.71 & 0.70     & & 58.0 & 42.9 & 69.0 & & 35.8 & 42.7 & 29.7 & 45.3 \\
\bd{\ourMethod{} (Swin-L)} & \cmarkg & \bd{0.72} & \bd{0.72} & \bd{0.73}     & & 58.9 & 43.7 & 69.9 & & \bd{48.0} & \bd{58.2} & \bd{39.0} & \bd{52.9} \\

\bottomrule 
\end{tabularx}

\label{tab:benchmark_results_vps}
\end{table*}

\begin{figure}
\centering

\begin{tikzpicture}[scale=3.5,
   y={(0.5cm,0.25cm)},x={(0.5cm,-0.25cm)},z={(0cm,{veclen(0.5,0.25)*1cm})}
    ]

    \begin{scope}[xshift=-0.1cm, yshift=-0.35cm]
    \DrawXYZ[]
    \end{scope}

    \begin{scope}[xshift=-0.17cm, yshift=0.05cm]
    \node[anchor=center, inner sep=0, rotate=90, text width=0.8cm] {\small Deformable \\ Attention};
    \end{scope}

    \DrawCubes[step=1mm,thin, draw=gray, fill=white]{0.0}{0.4}{0.4}{0.5}{0.0}{0.4}
    \DrawCubes[step=1mm,thin, draw=gray, fill=white]{0.0}{0.4}{0.2}{0.3}{0.0}{0.4}
    \DrawCubes[step=1mm,thin, draw=c_blue, fill=c_blue!30]{0.0}{0.4}{0}{0.1}{0.0}{0.4}
    \begin{scope}[xshift=0.6cm]
    \DrawCubes[step=0.75mm,thin, draw=gray, fill=white]{0}{0.6}{0.3}{0.375}{0}{0.6}
    \DrawCubes[step=0.75mm,thin, draw=gray, fill=white]{0}{0.6}{0.15}{0.225}{0}{0.6}
    \DrawCubes[step=0.75mm,thin, draw=c_blue, fill=c_blue!30]{0}{0.6}{0}{0.075}{0}{0.6}
    \DrawCubes[step=0.75mm,thick, draw=c_blue!70!black, fill=c_blue]{0.525}{0.6}{0}{0.075}{0.525}{0.6}
    \CubeCoord[anchor]{0.525}{0.6}{0}{0.075}{0.525}{0.6}
    \end{scope}

    \begin{scope}[xshift=1.25cm]
    \DrawCubes[step=0.5mm,thin, draw=gray]{0}{0.8}{0.3}{0.35}{0}{0.8}
    \DrawCubes[step=0.5mm,thin, draw=gray, fill=white]{0}{0.8}{0.15}{0.2}{0}{0.8}
    \DrawCubes[step=0.5mm,thin, draw=c_blue, fill=c_blue!30]{0}{0.8}{0}{0.05}{0}{0.8}
    \end{scope}
    
    \draw[-latex, c_blue!70!black] (anchor) -- (0.025,0.085);
    \draw[-latex, c_blue!70!black] (anchor) -- (0.29,0.01);
    \draw[-latex, c_blue!70!black] (anchor) -- (0.018,0.22);
    
    \draw[-latex, c_blue!70!black] (anchor) -- (0.6,0.9);
    \draw[-latex, c_blue!70!black] (anchor) -- (0.55,0.7);
    \draw[-latex, c_blue!70!black] (anchor) -- (0.3,1.1);
    
    \draw[-latex, c_blue!70!black] (anchor) -- (1.2,1.9);
    \draw[-latex, c_blue!70!black] (anchor) -- (1.0,2.0);
    \draw[-latex, c_blue!70!black] (anchor) -- (1.2,1.6);

    \begin{scope}[xshift=0.38cm, yshift=0.38cm]
    \node[anchor=center, inner sep=0] {\small $F_{32}$};
    \end{scope}
    
    \begin{scope}[xshift=1.02cm, yshift=0.41cm]
    \node[anchor=center, inner sep=0] {\small $F_{16}$};
    \end{scope}
    
    \begin{scope}[xshift=1.77cm, yshift=0.44cm]
    \node[anchor=center, inner sep=0] {\small $F_{8}$};
    \end{scope}
    
    \begin{scope}[xshift=-0.17cm, yshift=-0.75cm]
    \node[anchor=center, inner sep=0, rotate=90, text width=0.8cm] {\small Temporal \\ Attention};
    \end{scope}
    
    \begin{scope}[xshift=0.0cm, yshift=-0.8cm]
    \DrawCubes[step=1mm,thin, draw=gray, fill=white]{0.0}{0.4}{0.4}{0.5}{0.0}{0.4}
    \DrawCubes[step=1mm,thin, draw=c_red, fill=c_red!30]{0.2}{0.4}{0.4}{0.5}{0.2}{0.4}
    \DrawCubes[step=1mm,thin, draw=gray, fill=white]{0.0}{0.4}{0.2}{0.3}{0.0}{0.4}
    \DrawCubes[step=1mm,thin, draw=c_red, fill=c_red!30]{0.2}{0.4}{0.2}{0.3}{0.2}{0.4}
    \DrawCubes[step=1mm,thin, draw=gray, fill=white]{0.0}{0.4}{0}{0.1}{0.0}{0.4}
    \DrawCubes[step=1mm,thin, draw=c_red, fill=c_red!30]{0.2}{0.4}{0}{0.1}{0.2}{0.4}
    \end{scope}

    \begin{scope}[xshift=0.6cm, yshift=-0.8cm]
    \DrawCubes[step=0.75mm,thin, draw=gray, fill=white]{0}{0.6}{0.3}{0.375}{0}{0.6}
    \DrawCubes[step=0.75mm,thin, draw=c_red, fill=c_red!30]{0.3}{0.6}{0.3}{0.375}{0.3}{0.6}
    \DrawCubes[step=0.75mm,thin, draw=gray, fill=white]{0}{0.6}{0.15}{.225}{0}{0.6}
    \DrawCubes[step=0.75mm,thin, draw=c_red, fill=c_red!30]{0.3}{0.6}{0.15}{0.225}{0.3}{0.6}
    \DrawCubes[step=0.75mm,thin, draw=gray, fill=white]{0}{0.6}{0}{0.075}{0}{0.6}
    \DrawCubes[step=0.75mm,thin, draw=c_red, fill=c_red!30]{0.3}{0.6}{0}{0.075}{0.3}{0.6}
    \DrawCubes[step=0.75mm,thick, draw=c_red!70!black, fill=c_red]{0.525}{0.6}{0}{0.075}{0.525}{0.6}

    \end{scope}
    \begin{scope}[xshift=1.25cm, yshift=-0.8cm]
    \DrawCubes[step=0.5mm,thin, draw=gray]{0}{0.8}{0.3}{0.35}{0}{0.8}
    \DrawCubes[step=0.5mm,thin, draw=gray, fill=white]{0}{0.8}{0.15}{0.2}{0}{0.8}
    \DrawCubes[step=0.5mm,thin, draw=gray, fill=white]{0}{0.8}{0}{0.05}{0}{0.8}
    \end{scope}
    
\end{tikzpicture}
\caption{\textbf{Temporal Neck Layer.} Colored regions denote the attention field w.r.t the selected pixel (darkened). Deformable Attention is spatially unrestricted but temporally limited to a single frame, whereas Temporal Attention is spatially localized, but temporally unrestricted. $F_8$ is inactive for the temporal attention.}
\label{fig:neck}
\end{figure}
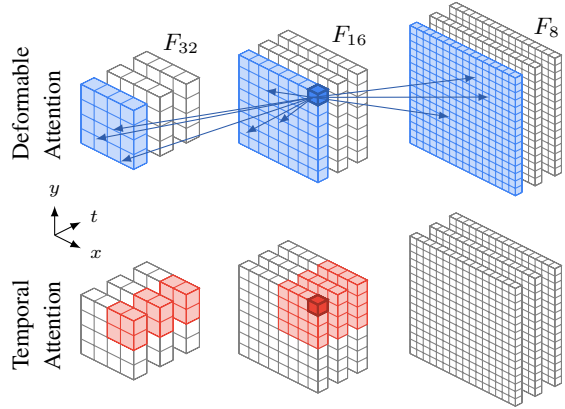

\PAR{Temporal Neck.}
\ourMethod{} produces target masks by computing the inner product between $Q_\text{out}$ and the video feature map $F_4$. For this to work, the per-pixel features $\mathcal{F}$ must be aligned for the same, and dissimilar for different targets.
Some image instance segmentation methods~\cite{zhu2020deformdetr,cheng2021mask2former} apply \emph{Deformable Attention}~\cite{zhu2020deformdetr} to the backbone feature maps to efficiently learn multi-scale image features. For \ourMethod{}, however, the features must also be temporally consistent across the entire input video clip.
To achieve this, we propose a novel \emph{Temporal Neck} architecture inspired from the work of Bertasius~\etal~\cite{gberta2021timesformer} for video action classification. 
We enable efficient spatio-temporal feature interaction by applying two types of self-attention in an alternating fashion: the first is spatially global and temporally localized, whereas the second is spatially localized and temporally global. The first operation is implemented with Deformable Attention, following existing work~\cite{zhu2020deformdetr,cheng2021mask2formervideo}.
The second operation, Temporal Attention, involves dividing the input space-time volume into a grid along the spatial axes, and then applying self-attention to the space-time feature volume inside each grid cell. Both operations allow feature interaction across multiple scales. Both attention operations are illustrated in Fig.~\ref{fig:neck}. We exclude $F_8$ from temporal attention since we found this to be more memory-efficient without negatively impacting prediction quality.

\PAR{Transformer Decoder.}
The decoder architecture follows that of Mask2Former~\cite{cheng2021mask2former}: the input queries are iteratively refined over multiple layers. In each layer, the queries first cross-attend to their respective masked video features, then self-attend to each other, followed by feed-forward layers. 

\subsection{Inference}
\label{subsec:method_inference}

To infer on videos with arbitrary length, we split videos into clips of length $T_\text{clip}$ with an overlap of $T_\text{ov}$ between successive clips. Object tracks are associated across clips based on their mask IoU in the overlapping frames. For the VOS tasks, the object queries for an intermediate clip are initialized by using the predicted masks in the overlapping frames from the previous clip as a pseudo-ground-truth. For VPS, we average the semantic segmentation logits in the overlapping frames. Our approach is thus \emph{near-online} because the time delay in obtaining the output for a given frame is at most $T_\text{clip}-T_\text{ov}-1$ (except for the first clip in the video).

\section{Experiments}
\label{sec:experiments}
\subsection{Implementation Details}
\label{subsec:experiments_implementation_details}
Our Temporal Neck contains 6 layers of Deformable and Temporal Attention.
We pretrain for 500k iterations on pseudo-video clips generated by applying on-the-fly augmentations to images from COCO~\cite{lin2014COCO}, ADE20k~\cite{zhou2017ade20k}, Mapillary~\cite{neuhold2017mapillary} and Cityscapes~\cite{cordts2016cityscapes}. The samples are either trained for VPS, VIS, VOS or PET. This is followed by fine-tuning for 90k iterations jointly on samples from YouTube-VIS~\cite{yang2019youtubevis}, OVIS~\cite{qi2022ovis}, KITTI-STEP~\cite{weber2021kittistep}, CityscapesVPS~\cite{kim2020vps}, VIPSeg~\cite{miao2022vipseg}, DAVIS~\cite{pont2017davis} and BURST~\cite{athar2023burst}. We train on 32 Nvidia A100 GPUs with batch size 1 per GPU. For each of the query types ($Q_\text{sem}$, $Q_\text{inst}$, $Q_\text{obj}$, $Q_\text{bg}$) discussed in Sec.~\ref{sec:method}, we employ a learned query embedding, which is used when computing the $\texttt{Key}^T\texttt{Query}$ affinity matrix for multi-head attention inside the decoder. We refer to the supplementary for more details.

\subsection{Benchmark Results}
\label{subsec:benchmark_results}

All results are computed with a single, jointly trained model which performs different tasks by simply providing the corresponding query set at run-time. 

\PAR{Video Instance Segmentation (VIS).}
We evaluate on %
(1) YouTube-VIS 2021~\cite{yang2019youtubevis} which covers 40 object classes and contains 2985/421 videos for training/validation, and (2) OVIS~\cite{qi2022ovis} which covers 25 object classes. It contains 607/140 videos for training/validation which are comparatively longer and more occluded. The AP scores for both are reported in Tab.~\ref{tab:benchmark_results_vis}.
For all three backbones, \ourMethod{} achieves state-of-the-art results for both benchmarks even though other methods are trained separately per benchmark whereas we use a single model. 
On YouTube-VIS, \ourMethod{} achieves 48.3 AP with a ResNet-50 backbone compared to the 45.7 achieved by VITA~\cite{heo2022vita}. With Swin-L, we achieve 60.2 AP which is also higher than the 57.5 by VITA. On OVIS with ResNet-50, our 31.1 AP is higher than the 30.2 for IDOL~\cite{wu2022idol}, and with Swin-L, \ourMethod{} (43.2 AP) outperforms the current state-of-the-art IDOL (42.6 AP). 

\PAR{Video Panoptic Segmentation (VPS).}
We evaluate VPS on three datasets: (1) KITTI-STEP~\cite{weber2021kittistep}, which contains 12/9 lengthy driving scene videos for training/validation with 19 semantic classes (2 \emph{thing} and 17 \emph{stuff} classes), (2) CityscapesVPS~\cite{kim2020vps}, which contains 50 short driving scene clips, each with 6 annotated frames, and (3) VIPSeg~\cite{miao2022vipseg}, which is a larger dataset with 2806/343 in-the-wild videos for training/validation and 124 semantic classes. The results are reported in Tab.~\ref{tab:benchmark_results_vps}. For KITTI-STEP, \ourMethod{} achieves 70\% STQ with a ResNet-50 backbone which is better than all existing approaches. The performance further improves to 72\% with Swin-L. For CityscapesVPS, \ourMethod{} achieves 58.9 VPQ which is higher than all other methods except VIP-Deeplab~\cite{qiao2021vipdeeplab} (63.1). However, VIP-Deeplab performs monocular depth estimation for additional guidance, and therefore requires ground-truth depth-maps for training.

For VIPSeg, \ourMethod{} outperforms existing approaches by a significant margin. With a ResNet-50 backbone, our 33.5 VPQ is 10.6\% higher than the 22.9 by Clip-PanoFCN~\cite{miao2022vipseg}. With a Swin-Large backbone, \ourMethod{} achieves 48.0 VPQ which is more than double that of Clip-PanoFCN (22.9). Note that VIP-Deeplab performs significantly worse for VIPSeg (16.0 VPQ), showing that \ourMethod{} generalizes better across benchmarks. Finally, we note that larger backbones results in significant performance gains for datasets with in-the-wild internet videos as in VIPSeg, but for specialized driving scene datasets (\eg KITTI-STEP and Cityscapes-VPS), the improvements are much smaller.

\PAR{Video Object Segmentation (VOS).}
We evaluate VOS on the DAVIS 2017~\cite{pont2017davis} dataset, which contains 60/30 YouTube videos for training/validation. The results in Tab.~\ref{tab:benchmark_results_vos} show that \ourMethod{} achieves 85.3 $\JnF$ which is higher than all existing methods except STCN~\cite{Cheng21STCN} (85.4) and XMem~\cite{cheng2022xmem} (86.2). 
As mentioned in Sec.~\ref{subsec:method_video_object_segmentation}, encoding objects as queries incurs a loss of fine-grained information, which is detrimental to performance. On the other hand, space-time correspondence (STC) based approaches learn pixel-to-pixel affinities between frames, which enables them to propagate fine-grained object appearance information. We note, however, that \ourMethod{} is the first method not based on the STC paradigm which achieves this level is performance (85.3 $\JnF$), outperforming several STC-based methods as well as all non-STC based methods \eg HODOR~\cite{athar2022hodor} (81.5) and UNICORN~\cite{yan2022unicorn} (70.6). 

\begin{table}[t]
\centering{}
\setlength{\tabcolsep}{3pt}
\newcommand\RotText[1]{\rotatebox{90}{\parbox{2cm}{\centering#1}}}
\small

\newcommand\bd[1]{\textbf{#1}}

\caption{
Results for VOS on DAVIS~\cite{pont2017davis} and PET on BURST~\cite{athar2023burst}. Detailed PET metrics are provided in supplementary.
}
\begin{tabularx}{\linewidth}{lp{0.0cm}YYYp{0.0cm}YY}
\toprule 
\multirow{2}{*}{Method} & & \multicolumn{3}{c}{\footnotesize DAVIS (VOS)}  & & \multicolumn{2}{c}{\footnotesize BURST (PET)} \\
\cmidrule{3-5} \cmidrule{7-8}
& & \footnotesize $\JnF$ & \footnotesize $\J$ & \footnotesize $\F$ & & \footnotesize $\mathrm{H}_\text{all}^\text{val}$ & \footnotesize $\mathrm{H}_\text{all}^\text{test}$  \\
\cmidrule{1-1} \cmidrule{3-5} \cmidrule{7-8} 
$\text{UNICORN}*$~\cite{yan2022unicorn}                  & & 70.6 & 66.1 & 75.0 & &  -   &  -   \\
HODOR~\cite{athar2022hodor}                              & & 81.3 & 78.4 & 83.9 & &  -   &  -   \\
STM~\cite{Oh19STMVOS}                                    & & 81.8 & 79.2 & 84.3 & &  -   &  -   \\
CFBI~\cite{Yang20CFBI}                                   & & 81.9 & 79.1 & 84.6 & & -    &  -    \\
HMMN~\cite{seong2021HMMVOS}                              & & 84.7 & 81.9 & 87.5 & & -    &  -   \\
AOT~\cite{Yang21AOTL}                                    & & 84.9 & 82.3 & 87.5 & & -    &  -   \\
STCN~\cite{Cheng21STCN}                                  & & 85.4 & 82.2 & 88.6 & & -    &  -   \\
XMem~\cite{cheng2022xmem}                                & & \bd{86.2} & \bd{82.9} & \bd{89.5} & &  -   &  -    \\
Box Tracker~\cite{luiten2020trackeval}                   & &  -   &   -  &  -   & & 12.7 & 10.1 \\
STCN+M2F~\cite{Cheng21STCN,cheng2021mask2former} & &  -   &   -  &  -   & & 24.4 & 24.9 \\
\textbf{\ourMethod{} (R-50)}                               & & 82.6 & 79.3 & 85.9 & & 30.9 & 32.1  \\
\textbf{\ourMethod{} (Swin-T)}                              & & 82.8 & 79.6 & 86.0 & & 36.0 & \bd{36.4} \\
\textbf{\ourMethod{} (Swin-L)}                             & & 85.3 & 81.7 & 88.5 & & \bd{37.5} & 36.1 \\
\bottomrule 
\end{tabularx}

\label{tab:benchmark_results_vos}
\end{table}

\PAR{Point Exemplar-guided Tracking (PET).}
PET is evaluated on the recently introduced BURST benchmark~\cite{athar2023burst} which contains 500/1000/1500 diverse videos for training/validation/testing. It is a constrained version of VOS which only provides the point coordinates of the object mask centroid instead of the full mask. Tab.~\ref{tab:benchmark_results_vos} shows that existing methods can only tackle either VOS or PET. To verify this, we tried adapting STCN~\cite{Cheng21STCN} for PET by training it with point masks, but the training did not converge. By contrast, \ourMethod{} encodes objects into queries, which enables it to tackle both tasks with a single model since the object guidance (point or mask) is abstracted behind the $\mathtt{EncodeObjects}(\cdot)$ function.

\begin{table*}[t]
\centering{}
\setlength{\tabcolsep}{3.0pt}
\newcommand\RotText[1]{\rotatebox{90}{\parbox{2cm}{\centering#1}}}
\small

\caption{
Ablation experiment results with ResNet-50 backbone. C-VPS: CityscapesVPS, YTVIS: YouTube-VIS, KITTI: KITTI-STEP.
}

\newcommand{\dat}[1]{\multirow{2}{*}{\rotatebox{90}{\rlap{\footnotesize #1}\hspace{21pt}}}}

\begin{tabularx}{\textwidth}{cp{3cm}cccccccp{0.2cm}YYp{0.0cm}YYYp{0.0cm}Yp{0.0cm}Y}
\toprule
                            && \multicolumn{7}{c}{Video Training Data} & &  \multicolumn{2}{c}{VIS} & & \multicolumn{3}{c}{VPS} & & VOS & & PET \\ 
\cmidrule{3-9}\cmidrule{11-12}\cmidrule{14-16}\cmidrule{18-18}\cmidrule{20-20}
\\[-6pt]%
&Setting                     & \dat{YTVIS} & \dat{OVIS} & \dat{KITTI} & \dat{C-VPS} & \dat{VIPSeg} & \dat{DAVIS} & \dat{BURST} & &  YTVIS & OVIS & & KITTI & C-VPS & VIPSeg & & DAVIS & & BURST        \\
                            && & & & & & & & &      (mAP)     & (mAP)  & &   (STQ) & (VPQ) & (VPQ)      & & ($\JnF$) & & ($\mathrm{HOTA}_\text{all}^\text{val}$)   \\
\midrule
1. & VIS                         & \cmarkg & \cmarkg & & & & & & & 46.3 & 31.5 & & - & - & -   & &  -   & &  -       \\
2. & VPS                         & & & \cmarkg & \cmarkg & \cmarkg & & & & - & - & & 0.70 & 49.7 & 32.4   & &  -   & &  -       \\
3.&VOS + PET                   & & & & & & \cmarkg & \cmarkg & &  -   & -    & & - & - & -   & & 81.1 & & 34.7  \\
4.&No Semantic Queries           & \cmarkg & \cmarkg & & & & & & & 44.7 & 29.8 & & - & - & -   & &  -   & &  -       \\ 
5.&No Temporal Neck            & \cmarkg & \cmarkg & \cmarkg & \cmarkg & \cmarkg & \cmarkg & \cmarkg & & 42.8 & 22.3 & & 0.69 & 51.2 & 28.9 & & 78.7  & &  30.3   \\
\midrule
&Final                       & \cmarkg & \cmarkg & \cmarkg & \cmarkg & \cmarkg & \cmarkg & \cmarkg & & 48.3 & 31.1 & & 0.70 & 53.3 & 33.5 & & 82.0 & & 30.9     \\
\bottomrule
\end{tabularx}

\label{tab:ablations}
\end{table*}
\ourMethod{} achieves a $\mathrm{HOTA}_\text{all}$ score of 37.5 and 36.4 on the validation and test sets, respectively, which is significantly better than the 24.4 and 24.9 achieved by the best performing baseline method which casts PET as a VOS problem by regressing a pseudo-ground-truth mask from the given point, followed by applying a VOS approach (STCN~\cite{Cheng21STCN}).

\begin{figure}[t]
    \centering
    \def\cellWidth{0.32\linewidth}
    \rotatebox{90}{\footnotesize \hspace{14pt}\textcolor{c_blue}{VIS}}\;%
    \includegraphics[width=\cellWidth]{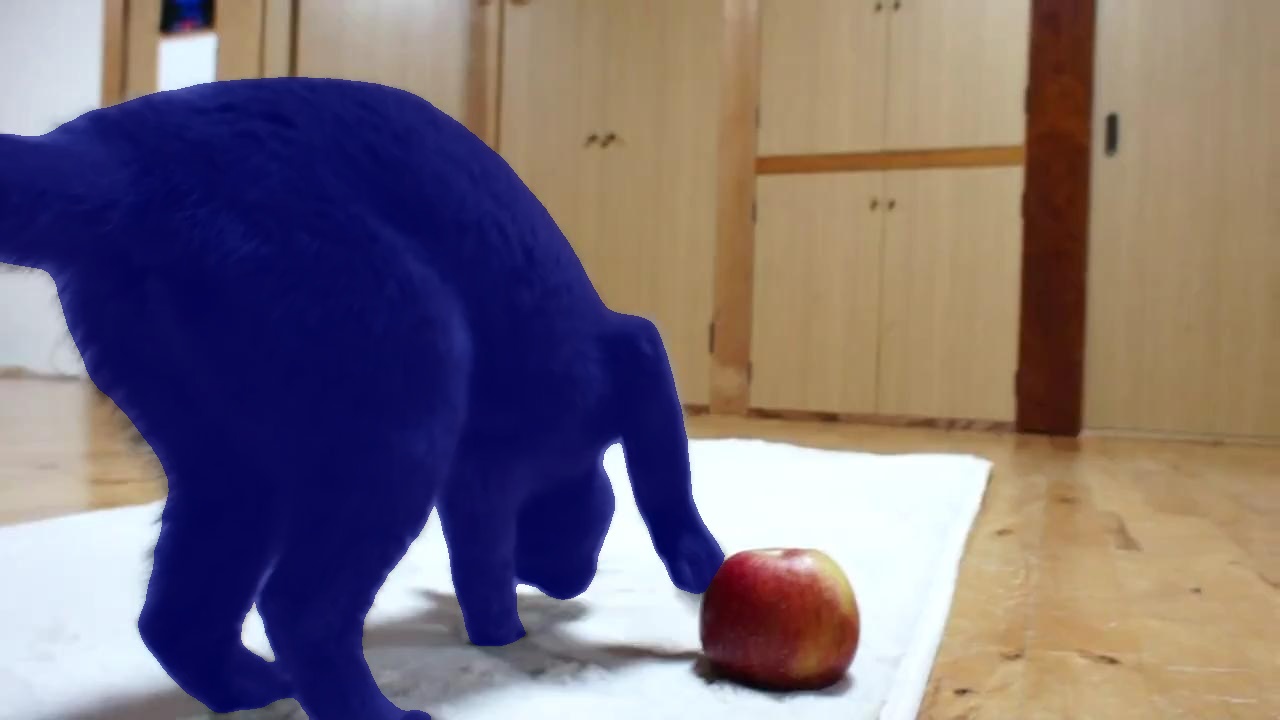}\hfill%
    \includegraphics[width=\cellWidth]{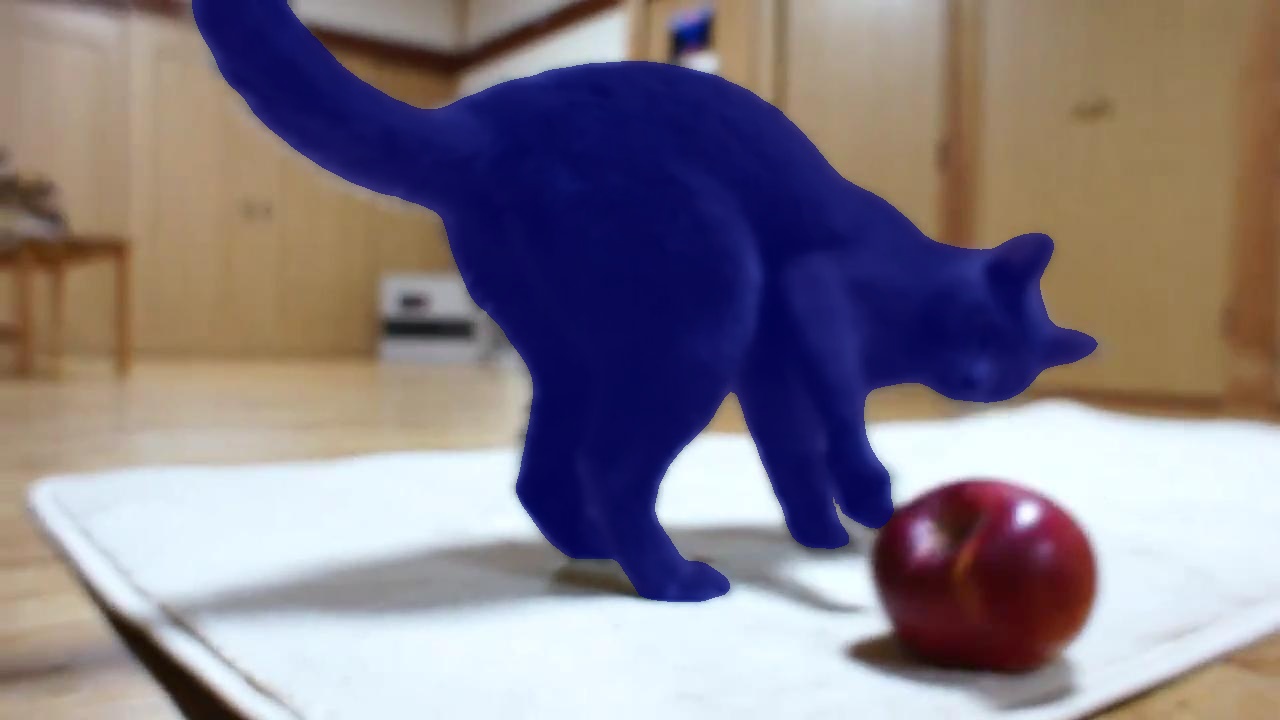}\hfill%
    \includegraphics[width=\cellWidth]{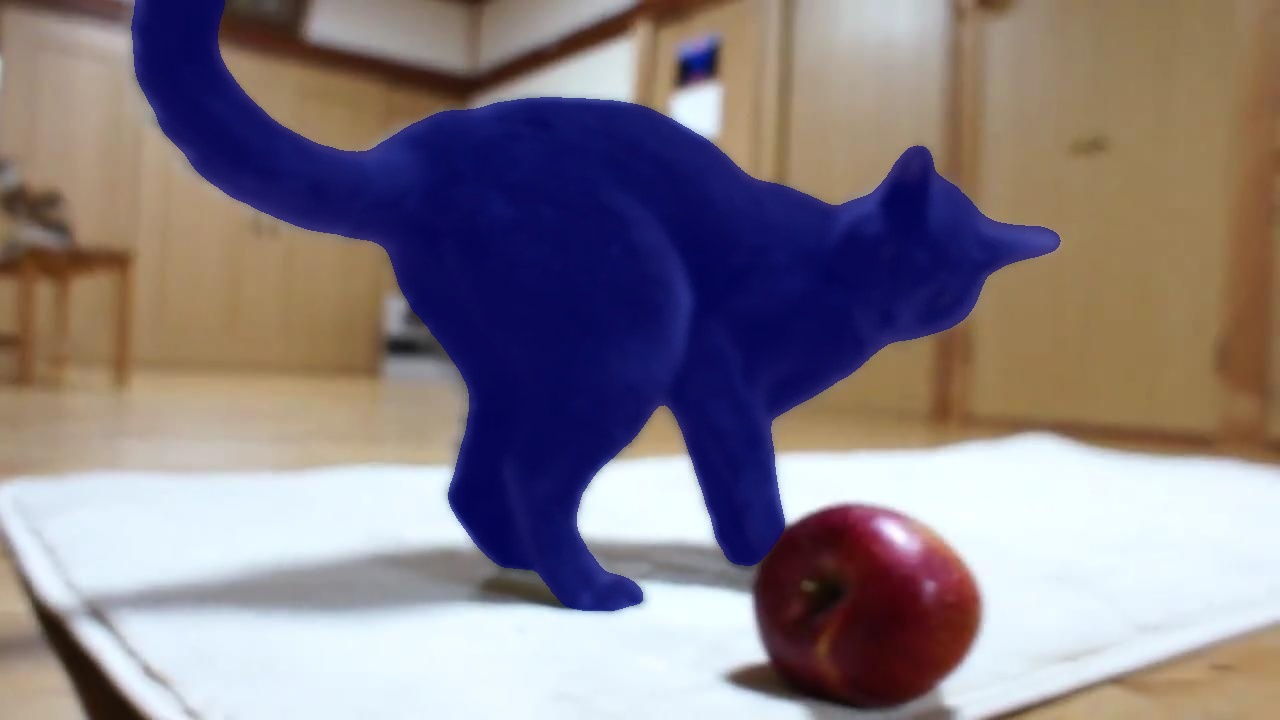} \\
    \rotatebox{90}{\footnotesize \hspace{9pt}\textcolor{c_orange}{VPS}}\;%
    \includegraphics[width=\cellWidth,trim={0cm 0cm 15cm 0cm},clip]{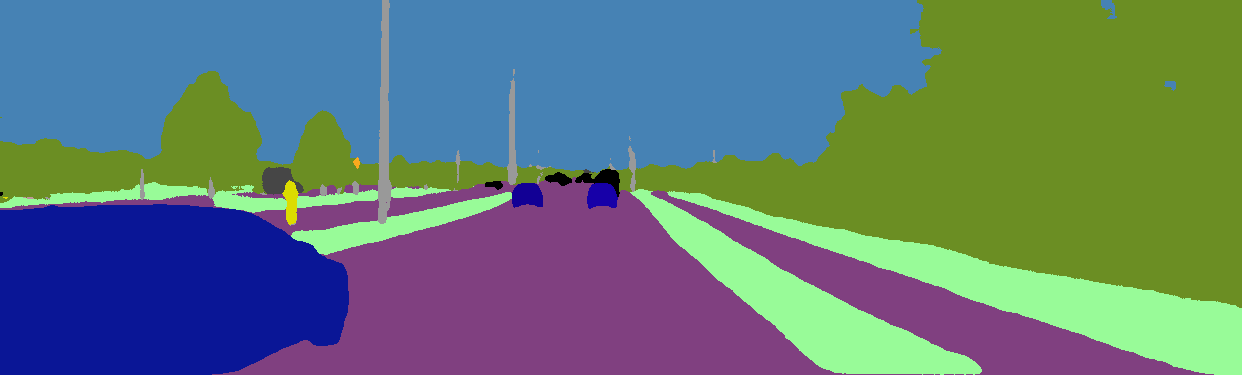}\hfill%
    \includegraphics[width=\cellWidth,trim={0cm 0cm 15cm 0cm},clip]{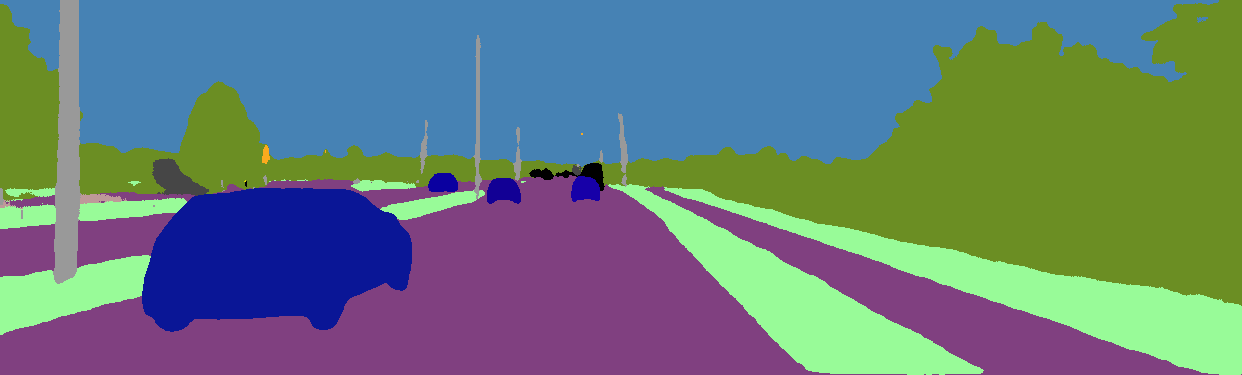}\hfill%
    \includegraphics[width=\cellWidth,trim={0cm 0cm 15cm 0cm},clip]{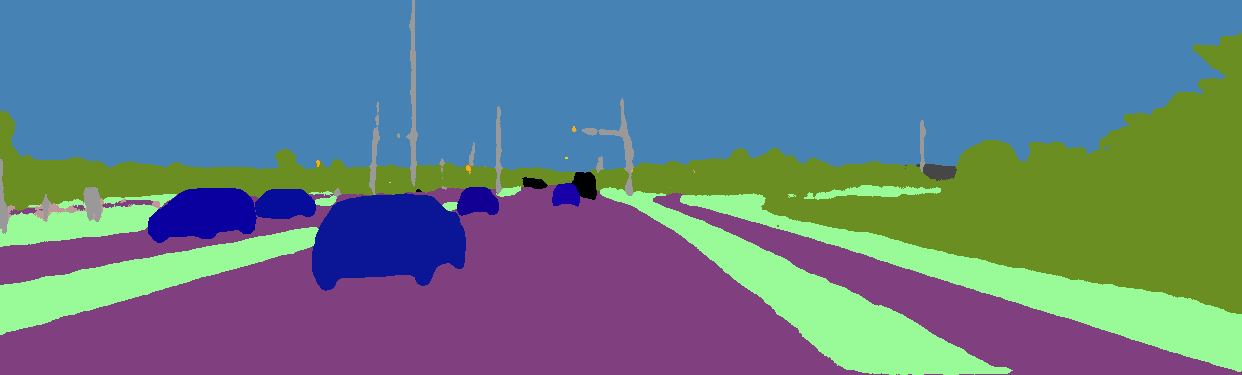}\hfill \\
    \rotatebox{90}{\footnotesize \hspace{14pt}\textcolor{c_red}{VOS}}\;%
    \includegraphics[width=\cellWidth]{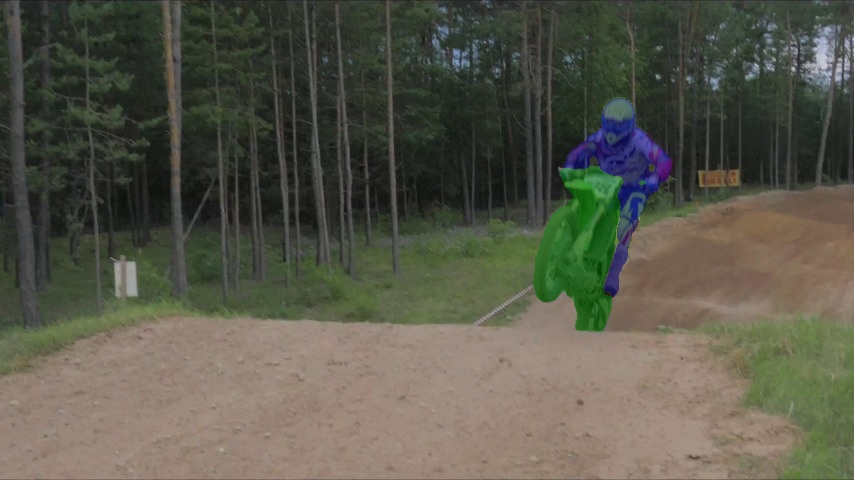}\hfill%
    \includegraphics[width=\cellWidth]{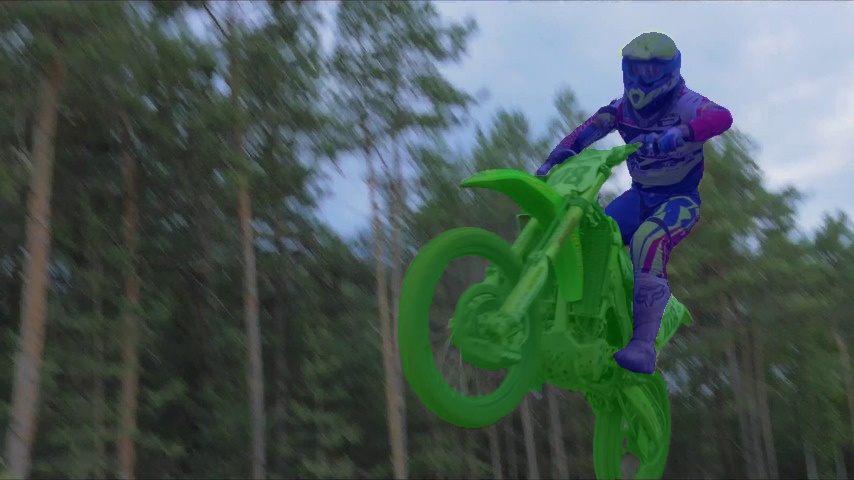}\hfill%
    \includegraphics[width=\cellWidth]{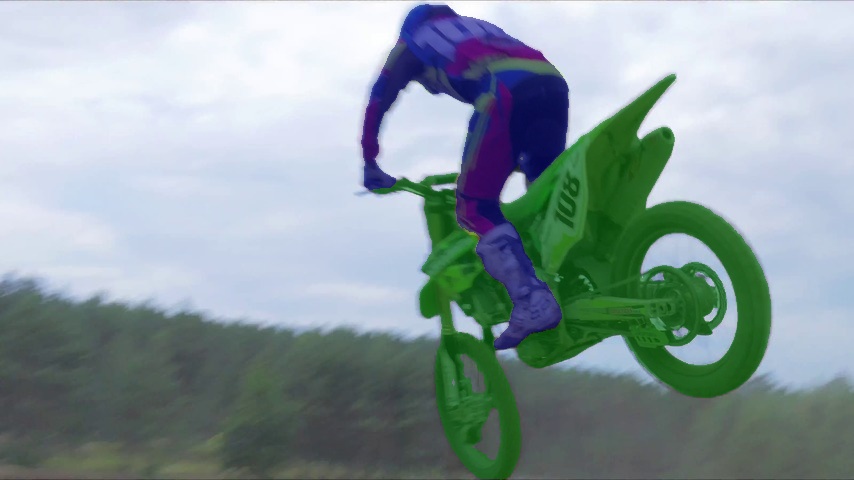} \\
    \rotatebox{90}{\footnotesize \hspace{17pt}\textcolor{c_green}{PET}}\;%
    \includegraphics[width=\cellWidth]{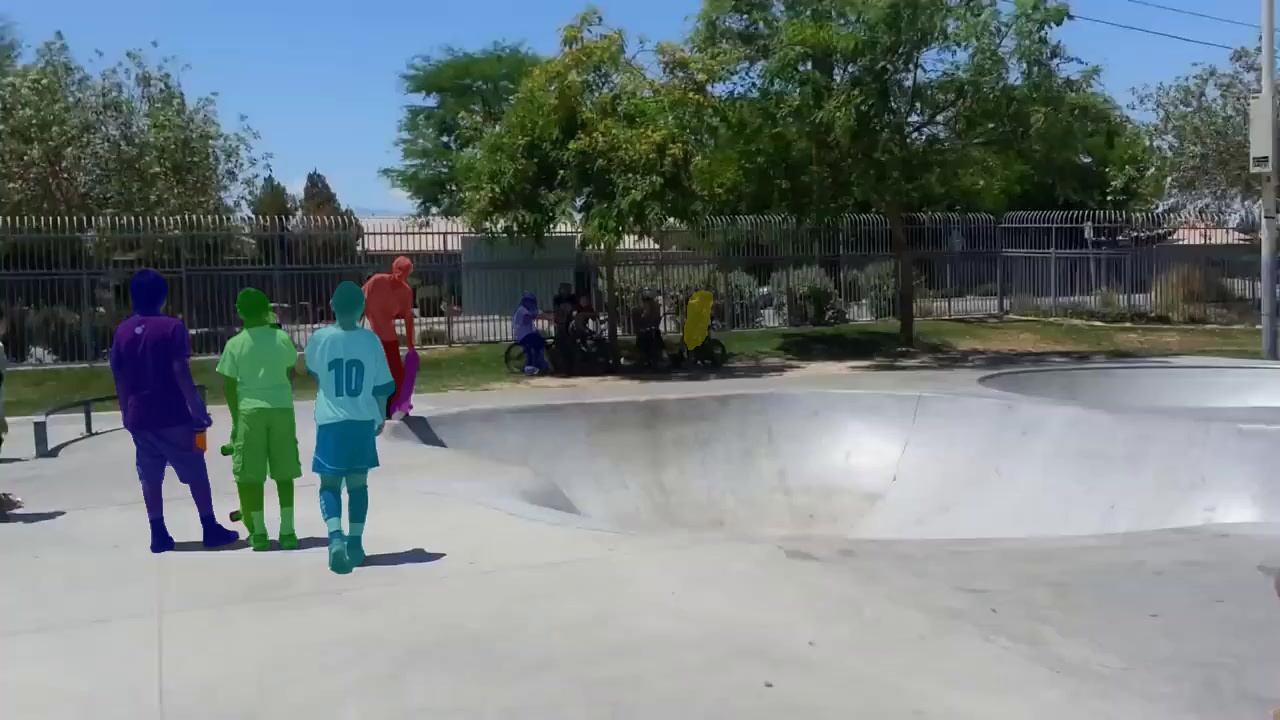}\hfill%
    \includegraphics[width=\cellWidth]{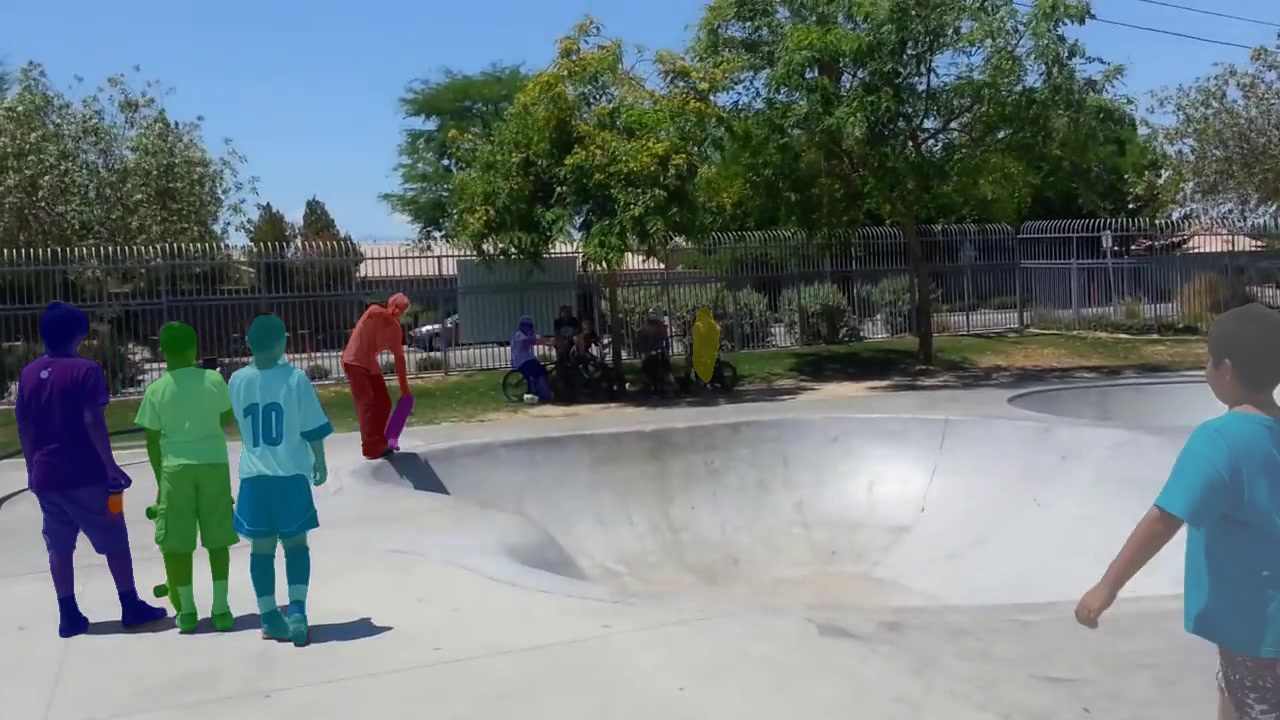}\hfill%
    \includegraphics[width=\cellWidth]{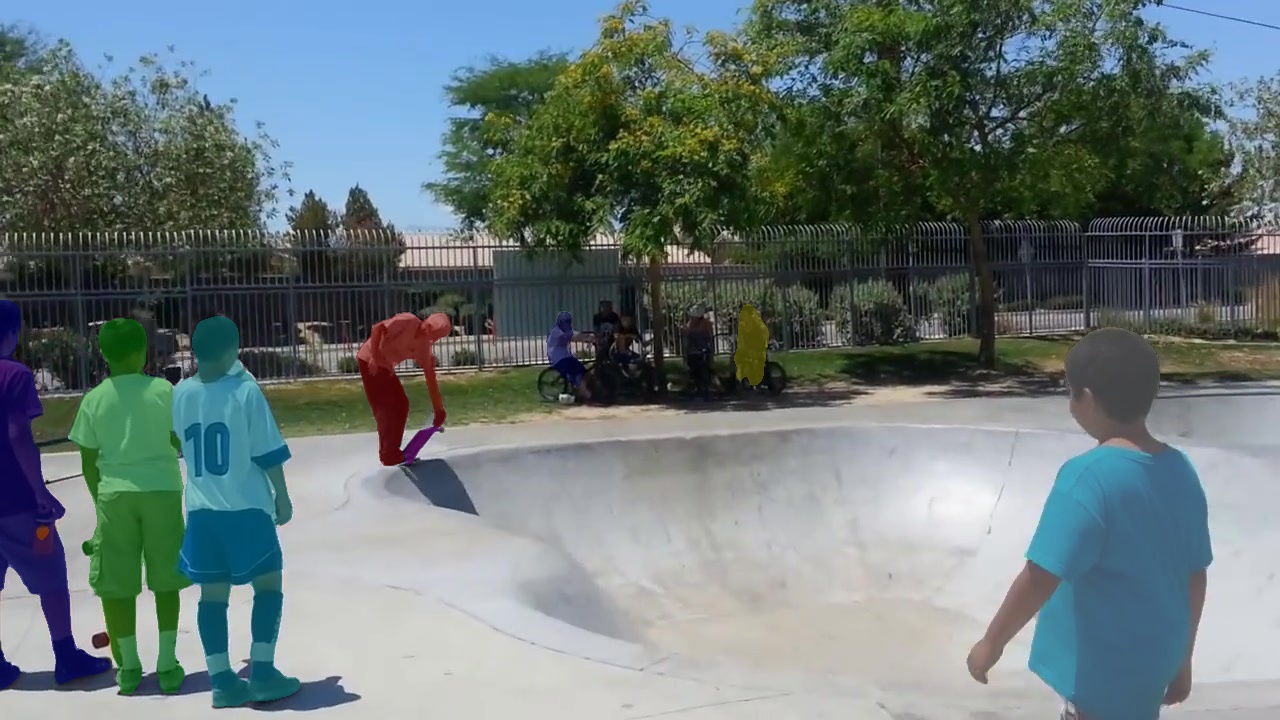} \\
    \caption{Qualitative results from a single \ourMethod{} model for all four tasks. Further results are shown in the supplementary.}
    \label{fig:qualitative_results}
\end{figure}

\subsection{Ablations}
\label{subsec:ablations}

Table~\ref{tab:ablations} shows several architecture/training ablations.

\PAR{Task-specific Training (row 1-3).} The first three rows show results for task-specific models. 
We train a single model for VOS and PET since both tasks are closely related. We note that the VIS-only model performs worse than the multi-task model on YouTube-VIS (46.3 vs. 48.3) but slightly better on OVIS (31.5 vs. 31.1). For VPS, the performance on KITTI-STEP is unchanged, but Cityscapes-VPS and VIPSeg both show improvements with the multi-task model. Lastly, for VOS the task-specific model performs slightly worse on DAVIS (81.1 vs. 82.0) but significantly better on BURST for PET (34.7 vs. 30.9). 
To summarize, the final, multi-task model performs better on 4/7 benchmarks, worse on 2/7, and matches performance on 1/7 when compared to task-specific models. We thus conclude that the combination of multi-task supervision and more data is generally beneficial for performance.

\PAR{Semantic Queries for VIS (row 4).} \ourMethod{} represents object classes as dynamic query inputs to the network ($Q_\text{sem}$, Sec.~\ref{subsec:method_video_instance_segmentation}). We ablate this by modifying our network to work for only VIS by discarding the semantic/background queries and adopting a technique similar to existing methods~\cite{wang2021vistr,cheng2021mask2former}, \ie using instance queries $Q_\text{inst}$ in conjunction with a linear layer for classification (separate for each dataset). Comparing the results with the VIS-only setting which is trained on similar data, we see that this architecture performs worse than the VIS-only setting on both YouTube-VIS (44.7 vs. 46.3) and OVIS (29.8 vs. 31.5). Thus, our semantic query based classification makes the network architecture task-agnostic and also yields better performance.

\PAR{Temporal Neck (row 4).} We validate our novel Temporal Neck (Sec.~\ref{subsec:method_network_architecture}) by training a model with a simpler neck that contains only Deformable Attention layers~\cite{zhu2020deformdetr}, similar to Mask2Former~\cite{cheng2021mask2former}, \ie there is no feature interaction across frames. 
Doing this degrades performance across all datasets, with particularly large drops for YouTube-VIS (42.8 vs. 48.3) and OVIS (22.3 vs. 31.1).
This shows that inter-frame feature interactions enabled by our Temporal Neck are highly beneficial for down-stream tasks.

\section{Discussion}
\label{sec:discussion}
\vspace{-5pt}
\PAR{Limitations.}
Training on multiple datasets/tasks does not necessarily improve performance on all benchmarks. For VOS, the model exhibits class bias and sometimes fails to track unusual objects which were not seen during training. 

\begin{figure}[t]
    \centering
    \def\cellWidth{0.497\linewidth}
    \includegraphics[width=\cellWidth]{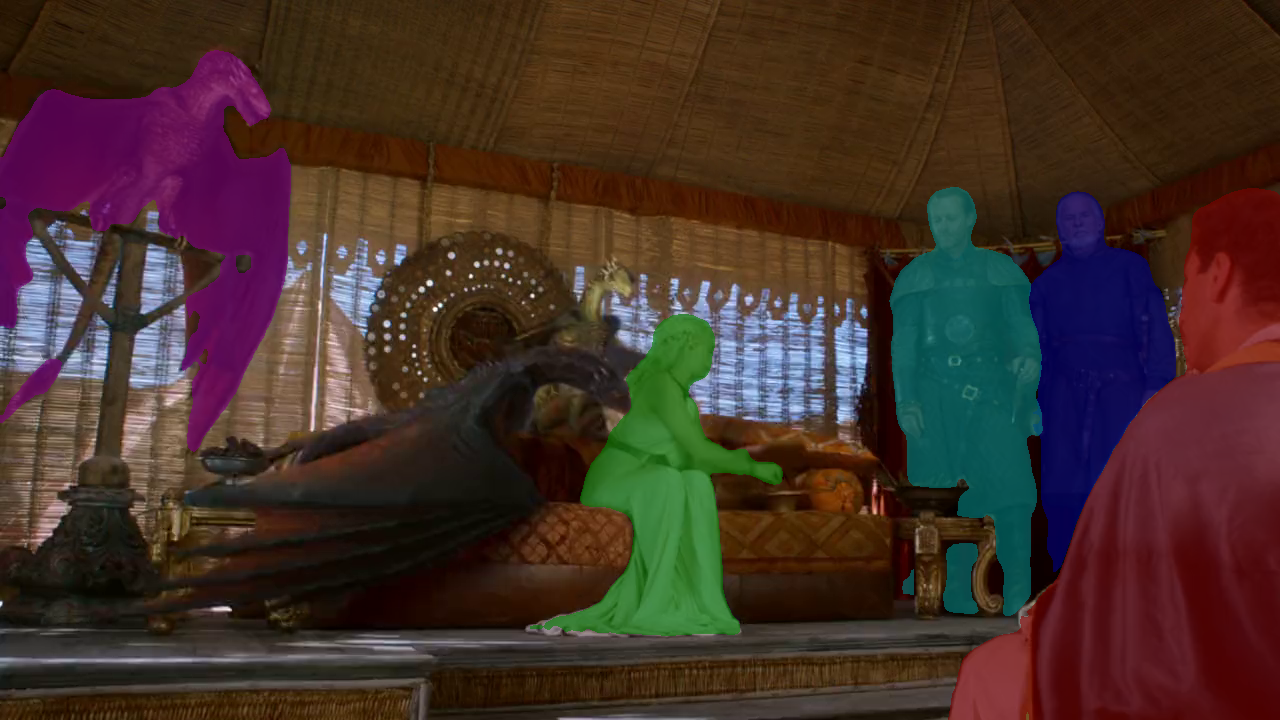}\hfill%
    \includegraphics[width=\cellWidth]{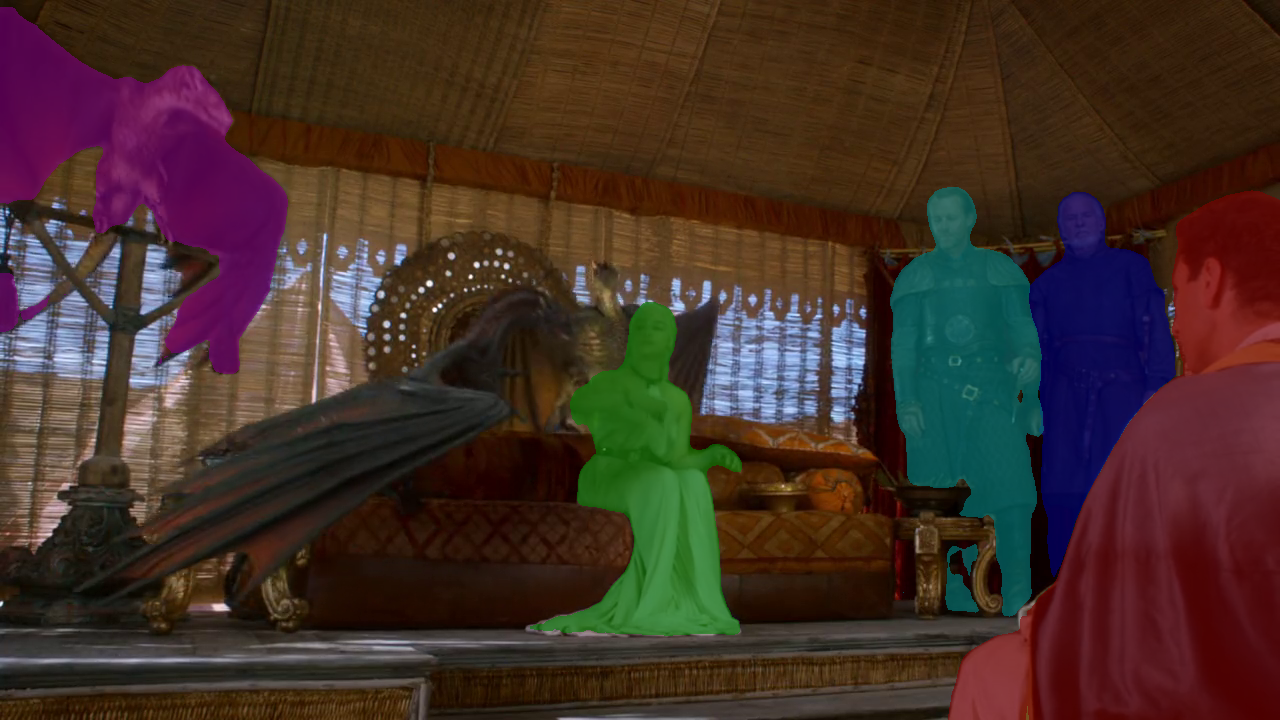}\\
    \includegraphics[width=\cellWidth]{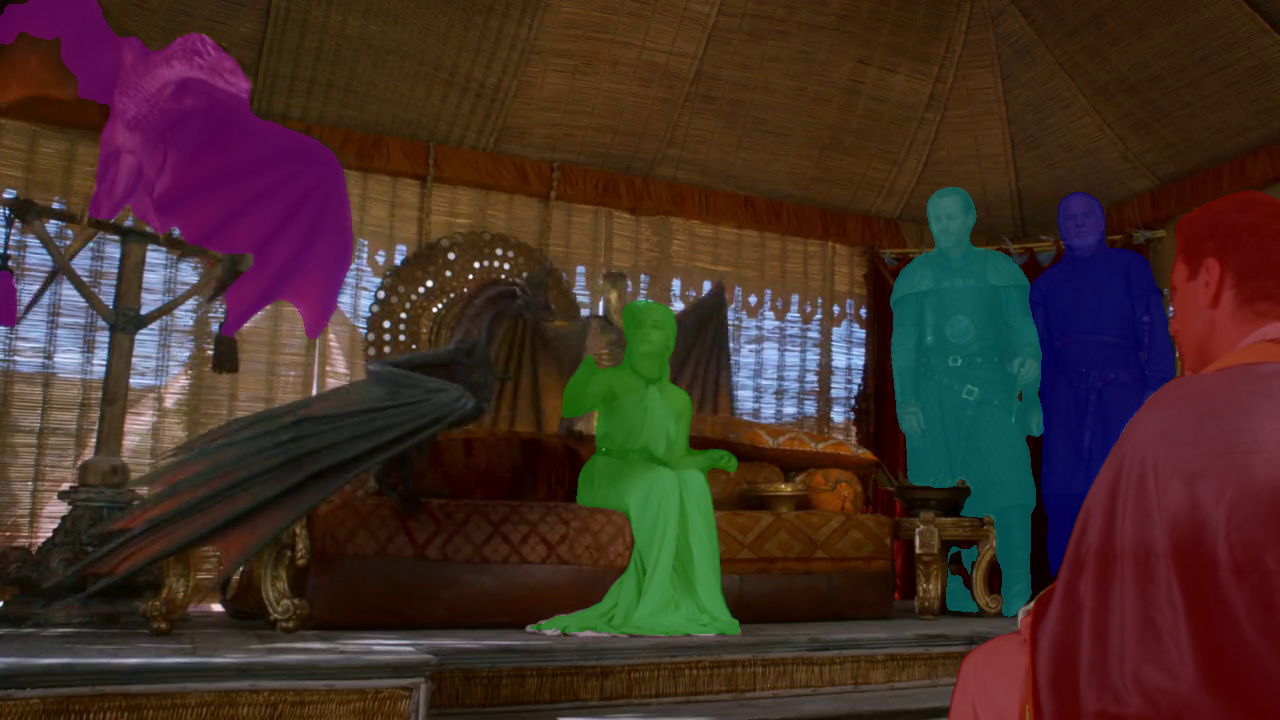}\hfill%
    \includegraphics[width=\cellWidth]{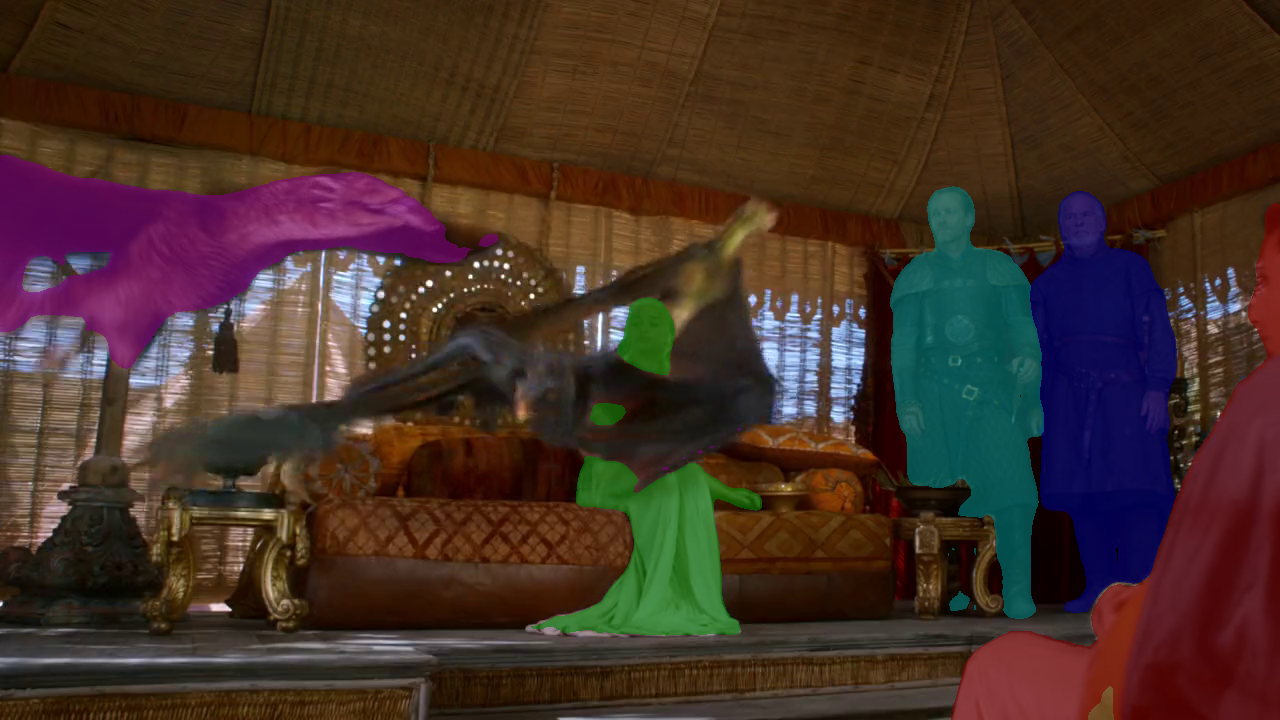}%
    \caption{\ourMethod{} performing VIS and VOS in a single forward pass. We provide the mask for the dragon on the left, and the semantic query for the `person' class.}
    \label{fig:dragon_sequence}
\end{figure}

\PAR{Future Outlook.}
We jointly trained \ourMethod{} for four different tasks to validate its generalization capability. The architecture can, however, tackle any video segmentation task for which the targets can be encoded as queries. The recent emergence of joint language-vision models~\cite{radford2021lclip,li2022GLIP,ramesh2021DallE} thus makes it possible to perform multi-object segmentation based on a text prompt if the latter can be encoded as a target query using a language encoder~\cite{devlin2019bert}. Another interesting possibility is that \ourMethod{} could be applied to multiple tasks \emph{in the same forward pass} by simply concatenating the task-specific queries. Fig.~\ref{fig:dragon_sequence} offers a promising outlook for this; it shows our model's output for a video clip from a popular TV series where we perform VIS and VOS simultaneously by providing the semantic query for the `person' class (from YouTube-VIS~\cite{yang2019youtubevis}), and the VOS-based object queries for the dragon by annotating its first frame mask, \ie $Q_\text{in} = \mathtt{concat}(Q_\text{sem},Q_\text{inst},Q_\text{obj},Q_\text{bg})$. \ourMethod{} successfully segments all four persons in the scene (VIS) and the dragon (VOS), even though our model was never trained to simultaneously tackle both tasks in a single forward pass.

\section{Conclusion}
\label{sec:conclusion}
We presented \ourMethod{}: a novel, unified approach for tackling any task requiring pixel-precise segmentation of a set of \emph{targets} in video. We adopt a generalized paradigm where the task-specific targets are encoded into a set of \emph{queries} which are then input to our network together with the video features. The network is trained to produce segmentation masks for each target entity, but is inherently agnostic to the task-specific definition of these targets. To demonstrate the effectiveness of our approach, we applied it to four different video segmentation tasks (VIS, VPS, VOS, PET). We showed that a single \ourMethod{} model can be jointly trained for all tasks, and during inference can hot-swap between tasks without any task-specific fine-tuning. Our model achieved state-of-the-art performance on five benchmarks %
and has multiple, promising directions for future work.

\PAR{Acknowledgments.}\hspace{-5pt}
This project was partially funded by ERC CoG DeeVise (ERC-2017-COG-773161) and BMBF project NeuroSys-D (03ZU1106DA).
Compute resources were granted by RWTH Aachen under project ID $\texttt{supp0003}$, and by the Gauss Centre for Supercomputing e.V. through the John von Neumann Institute for Computing on the GCS Supercomputer JUWELS at Jülich Supercomputing Centre.

\clearpage

{\small
\bibliographystyle{ieee_fullname}
\bibliography{references}
}

\clearpage
\twocolumn[{%
 \centering
 \LARGE \textbf{Supplementary Material\\[1cm]}
}]

\setcounter{equation}{0}
\setcounter{figure}{0}
\setcounter{table}{0}
\setcounter{page}{1}
\setcounter{section}{0}
\makeatletter
\renewcommand{\theequation}{S\arabic{equation}}
\renewcommand{\thefigure}{S\arabic{figure}}
\renewcommand{\thetable}{S\arabic{table}}
\renewcommand{\thesection}{S\arabic{section}}

\section{Extended VOS/PET Ablations}
\label{sec:supp_extended_ablations}

Extended ablation results are given in Table~\ref{tab:extended_ablations} and discussed below. For these experiments we use a shorter/lighter training schedule compared to the results presented in the main text: the network is pre-trained on augmented image sequences generated from the COCO dataset for 360k iterations on 8 GPUs, followed by fine-tuning on actual video data from the DAVIS~\cite{pont2017davis} and BURST~\cite{athar2023burst} datasets.

\begin{table}[h]
\centering{}
\setlength{\tabcolsep}{4.0pt}
\newcommand\RotText[1]{\rotatebox{90}{\parbox{2cm}{\centering#1}}}
\small

\caption{
Extended ablation results for VOS and PET tasks on DAVIS~\cite{pont2017davis} and BURST~\cite{athar2023burst} benchmarks, respectively.
}

\begin{tabular}{lcc}
\toprule 
Setting      & VOS ($\JnF$) & PET ($\mathrm{HOTA}_\text{all}$) \\
\midrule
Without $Q_\text{bg}$     & 78.0 & 25.9  \\
$|Q_\text{obj}|=q_0=1$    & 80.6 & 28.2    \\
Final                     & 81.5 & 29.2 \\
\bottomrule 
\end{tabular}

\label{tab:extended_ablations}
\end{table}

\PAR{Background Queries (row 1).}
We stated in the main text that we model the non-object pixels in the input video using background queries for the VOS and PET task. We ablate this design decision by training \ourMethod{} without this sort of background modeling, \ie for both VOS and PET tasks, the input set of queries contains only the object queries $Q_\text{obj}$. This reduces the $\JnF$ score for VOS from 81.5 to 78.0, and the $\mathrm{HOTA}_\text{all}$ score for PET from 29.2 to 25.9. Thus, we conclude that background modeling has a noticeable, positive impact on prediction quality.

\PAR{Number of Object Queries (row 2).}
We mentioned in the main text that we modify the approach adopted by HODOR~\cite{athar2022hodor} for VOS by using multiple ($q_0$) object queries to represent a single target object. We ablate this by training our model using $q_0=1$ (in the final setting we use $q_0=4$). We see that this causes the performance on DAVIS to reduce from 81.5 to 80.6, and that on BURST from 29.2 to 28.2. Note that $q_0=1$ for PET even for the final setting, but because PET inference over lengthy videos involves VOS-style mask-guidance, the choice of $q_0$ for VOS affects performance for PET as well.

\begin{table}[t]
\centering{}
\setlength{\tabcolsep}{3pt}
\newcommand\RotText[1]{\rotatebox{90}{\parbox{2cm}{\centering#1}}}
\small

\newcommand\bd[1]{\textbf{#1}}

\caption{
Extended results for PET on the BURST~\cite{athar2023burst} validation and test sets. (`$\mathrm{H}$' denotes `$\mathrm{HOTA}$'~\cite{luiten2020hota}).
}

\begin{tabularx}{\linewidth}{lp{0.0cm}YYYp{0.0cm}YYY}
\toprule 
\multirow{2}{*}{Method} & & \multicolumn{3}{c}{\footnotesize BURST (val)} & & \multicolumn{3}{c}{\footnotesize BURST (test)} \\
\cmidrule{3-5}\cmidrule{7-9}
& & \footnotesize $\mathrm{H}_\text{all}$ & \footnotesize $\mathrm{H}_\text{com}$ & \footnotesize $\mathrm{H}_\text{unc}$ & & \footnotesize $\mathrm{H}_\text{all}$ & \footnotesize $\mathrm{H}_\text{com}$ & \footnotesize $\mathrm{H}_\text{unc}$  \\
\cmidrule{1-1} \cmidrule{3-5}\cmidrule{7-9}
Box Tracker~\cite{luiten2020trackeval}                   &  & 12.7 & 31.7 & 7.9 & &  10.1 & 24.4 & 7.3  \\
STCN+M2F~\cite{Cheng21STCN,cheng2021mask2former} & & 24.4 & 44.0 & 19.5 & &  24.9   &  39.5  &  22.0  \\
\textbf{\ourMethod{} (R-50)}                               & & 30.9 & 43.2 & 27.8 & & 32.1  &  41.5  &  30.2  \\
\textbf{\ourMethod{} (Swin-T)}                              & & 36.0 & 47.7 & 33.0 & &  36.4   &  45.0  &  34.7  \\
\textbf{\ourMethod{} (Swin-L)}                             &  & \bd{37.5} & \bd{51.7} & \bd{34.0} & & \bd{36.1}  &  \bd{47.1}  & \bd{33.8}  \\
\bottomrule 
\end{tabularx}

\label{tab:benchmark_results_pet_extended}
\end{table}

\section{Detailed BURST Metrics}
\label{sec:supp_detailed_burst_results}

Due to space constraints, we only presented the final $\mathrm{HOTA}_\text{all}$ score for the BURST benchmark in the main paper. Table~\ref{tab:benchmark_results_pet_extended} gives a more detailed breakdown for those results.

\section{Implementation Details}
\label{sec:supp_implementation_details}
Several details related to the training and inference setup which were omitted from the main paper are given below.

\PAR{Hardware Setup and Training Schedule.}
We train our models on 32 Nvidia A100 GPUs with a batch size of 32 with clips of 3 frames. The pretraining takes 2-3 days depending on the backbone whereas finetuning takes 10-16 hours. An AdamW optimizer is used with a learning rate of $10^{-4}$ at the start, followed by two step decays with a factor of 0.1 each.

\PAR{Inference.} 
Inference is performed on a single RTX 3090 and runs at 6-10 fps using a Swin-T backbone. The variation mainly arises because different datasets have different image resolutions. For most datasets, we use clips containing 12 frames with a 6 frame overlap between successive clips.

\PAR{Loss Supervision.}
Table~\ref{tab:loss_types} shows the type of loss function applied for mask regression for different tasks. Generally, the supervision signal is a combination of DICE and cross-entropy losses. For instances/objects we apply per-pixel binary cross-entropy whereas for semantic segmentation (where multiple classes compete for each pixel), we apply a multi-class cross-entropy loss. We apply a sparse loss similar to Cheng~\etal~\cite{cheng2021mask2former}, \ie, the loss is not applied to every pixel in the mask, but rather only to a subset of pixels which contain a certain fraction of hard negatives and other randomly sampled points. This type of supervision strategy was first proposed by Kirillov~\etal~\cite{kirillov2020pointrend}.

\begin{table}[t]
\centering{}
\setlength{\tabcolsep}{4.0pt}
\newcommand\RotText[1]{\rotatebox{90}{\parbox{2cm}{\centering#1}}}
\small

\caption{
Loss functions used for mask prediction for different targets. BCE: Binary cross-entropy, MCE: Multi-class cross-entropy, DICE: soft IoU loss
}

\begin{tabular}{lp{0.1cm}cp{0.1cm}cc}
\toprule 

Target Type      && Task  & & Loss  \\
\midrule
Instance         && \multirow{2}{*}{VIS}   & & DICE + BCE   \\
Semantic Class   &&    & & MCE       \\
\midrule
Instance         && \multirow{2}{*}{VPS}   & & DICE + BCE  \\
Semantic Class   &&       & & MCE         \\
\midrule
Object           &&  VOS/PET  & & DICE + BCE \\

\bottomrule 
\end{tabular}

\label{tab:loss_types}
\end{table}
\PAR{Pretraining.}
We pretrain on synthetic video samples generated by applying random, on-the-fly augmentations from the following image-level datasets: COCO~\cite{lin2014COCO}, ADE20k~\cite{zhou2017ade20k}, Mapillary~\cite{neuhold2017mapillary}, Cityscapes~\cite{cordts2016cityscapes}. Since these datasets provide panoptic annotations, we can train the data samples as any of the four target tasks (VPS, VIS, VOS, PET) \eg to train for VOS/PET, we assume that the first-frame mask/point is available for a random subset of ground-truth objects and ignore the class labels. The task weights for pretraining are given in Table~\ref{tab:pretrain_weights}.

\begin{table}[h]
\centering{}
\setlength{\tabcolsep}{4.0pt}
\newcommand\RotText[1]{\rotatebox{90}{\parbox{2cm}{\centering#1}}}
\small

\caption{
Task weights during pretraining stage.
}

\begin{tabular}{lcccc}
\toprule 
Task      & VPS & VIS & VOS & PET \\
\midrule
Weight    & 0.3 & 0.3 & 0.28 & 0.12 \\
\bottomrule 
\end{tabular}

\label{tab:pretrain_weights}
\end{table}

\PAR{Video Finetuning.}
The finetuning is done on actual video datasets for all four tasks. The sampling weights for each dataset/task are given in Table~\ref{tab:finetune_weights}. Note that data samples from DAVIS~\cite{pont2017davis} and BURST~\cite{athar2023burst} can be trained for both VOS and PET.

\begin{table}[h]
\centering{}
\setlength{\tabcolsep}{4.0pt}
\newcommand\RotText[1]{\rotatebox{90}{\parbox{2cm}{\centering#1}}}
\small

\caption{
Dataset weightage during video finetuning.
}

\begin{tabular}{lcc}
\toprule 
Dataset      & Task & Weight \\
\midrule
KITTI-STEP~\cite{weber2021kittistep}  & VPS & 0.075 \\
CityscapesVPS~\cite{kim2020vps}       & VPS & 0.075 \\
VIPSeg~\cite{miao2022vipseg}          & VPS & 0.15 \\
YouTube-VIS~\cite{yang2019youtubevis} & VIS & 0.225 \\
OVIS~\cite{qi2022ovis}                & VIS & 0.225 \\
DAVIS~\cite{pont2017davis}            & VOS/PET & 0.05 \\
BURST~\cite{athar2023burst}           & VOS/PET & 0.2 \\
\bottomrule 
\end{tabular}

\label{tab:finetune_weights}
\end{table}

\PAR{Point Exemplar-guided Tracking Inference.}
As mentioned in Sec.~3 of the main text, the PET task is tackled using the same workflow as for VOS \ie the target objects are encoded as object queries using the Object Encoder. An additional detail about inference on arbitrary length video sequences which is not mentioned in the main text is as follows: the point $\xrightarrow{}$ object query regression is only used for the first clip in which the object appears. For subsequent clips, we have access to the dense mask predictions for that object from our model. Hence, for subsequent clips, we regress the object query from the previous mask predictions (as we do for VOS).

\section{Query Visualization}
\label{sec:supp_query_viz}

\begin{figure*}
\begin{minipage}[c]{0.58\textwidth}%
    \centering
    \def\cellWidth{0.32\linewidth}
    \includegraphics[width=\cellWidth]{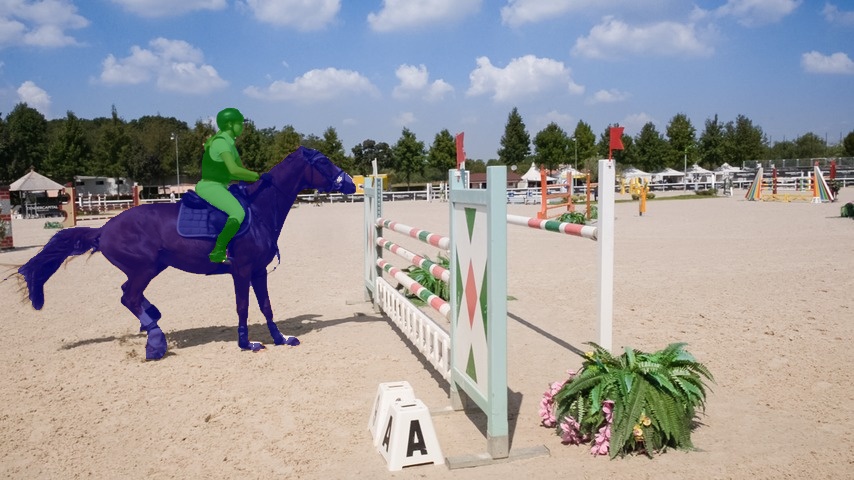}\hfill%
    \includegraphics[width=\cellWidth]{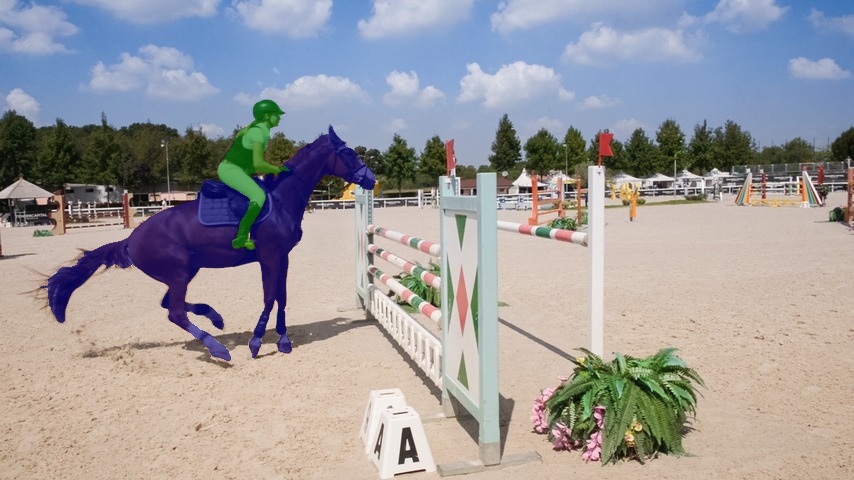}\hfill%
    \includegraphics[width=\cellWidth]{figures/supplementary/query_viz/horsejump/00002.jpg} \\
    \includegraphics[width=\cellWidth]{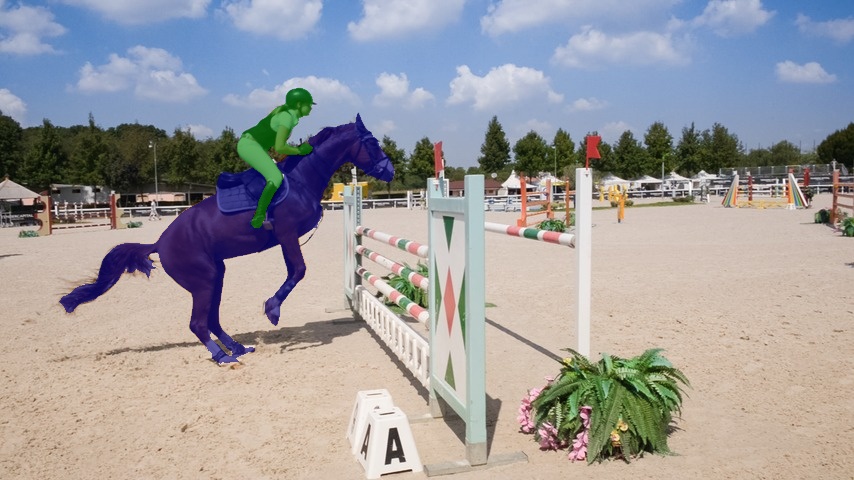}\hfill%
    \includegraphics[width=\cellWidth]{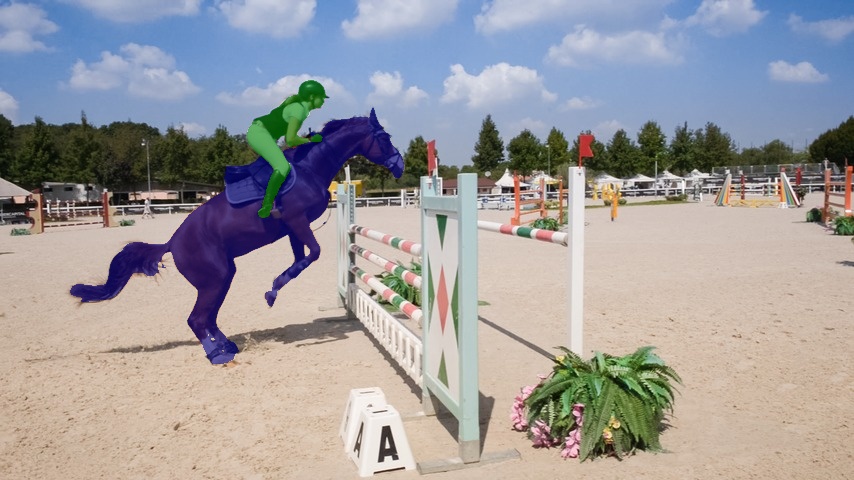}\hfill%
    \includegraphics[width=\cellWidth]{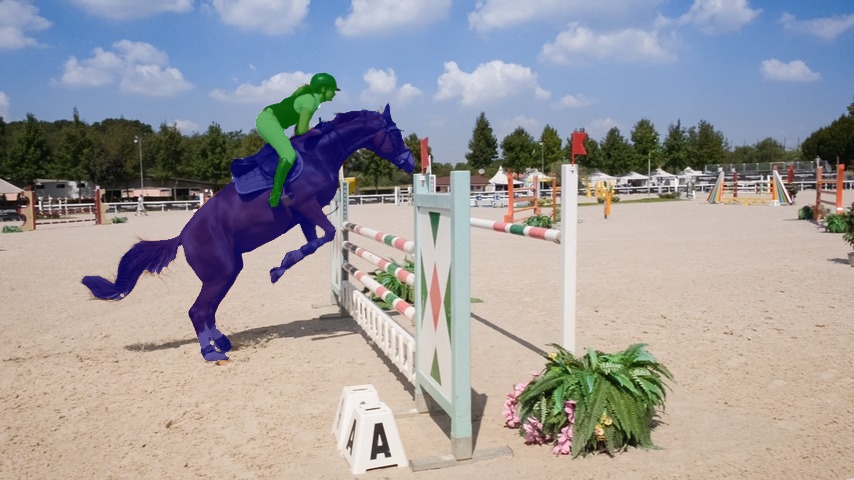} \\
    \includegraphics[width=\cellWidth]{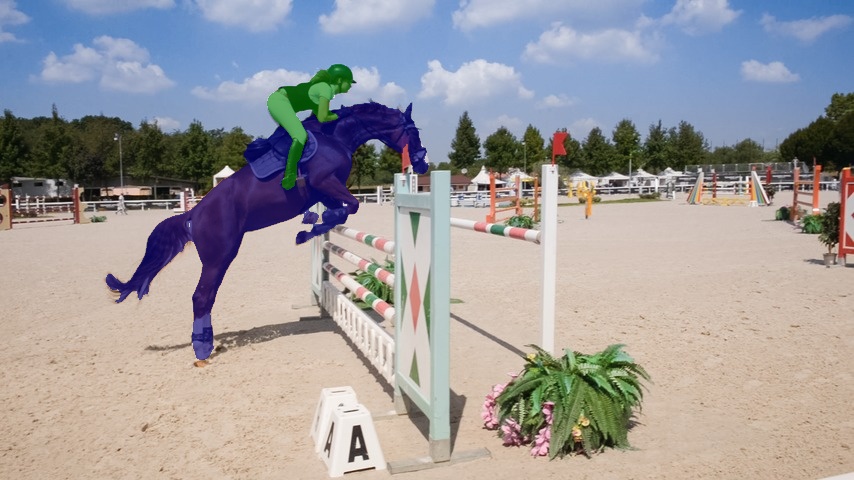}\hfill%
    \includegraphics[width=\cellWidth]{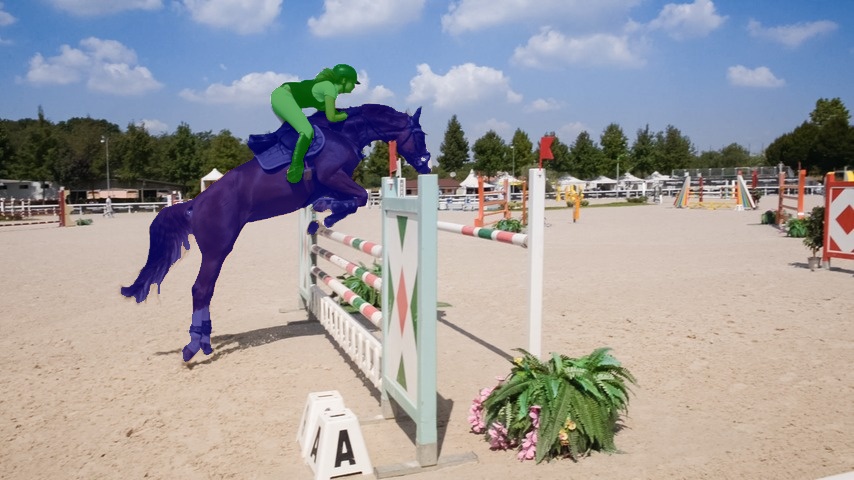}\hfill%
    \includegraphics[width=\cellWidth]{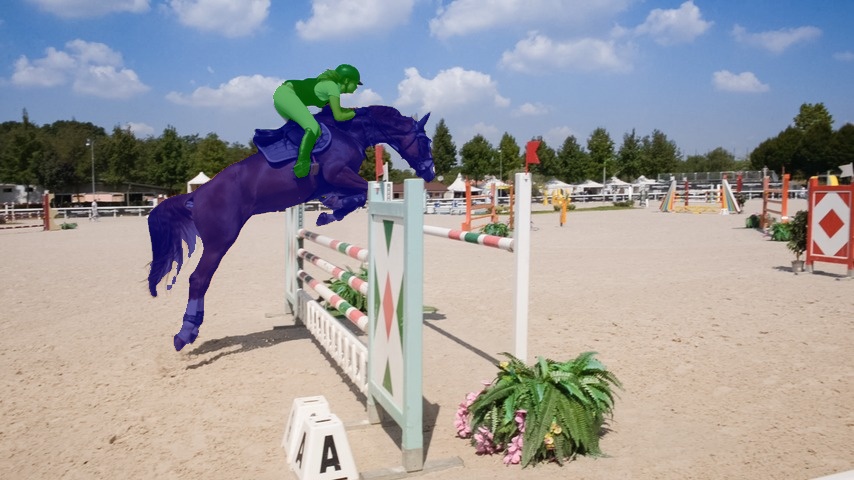} \\
\end{minipage}%
\begin{minipage}[c]{0.42\textwidth}%
\includegraphics[width=\linewidth,trim={1.22cm 0.7cm 1.6cm 1.3cm},clip]{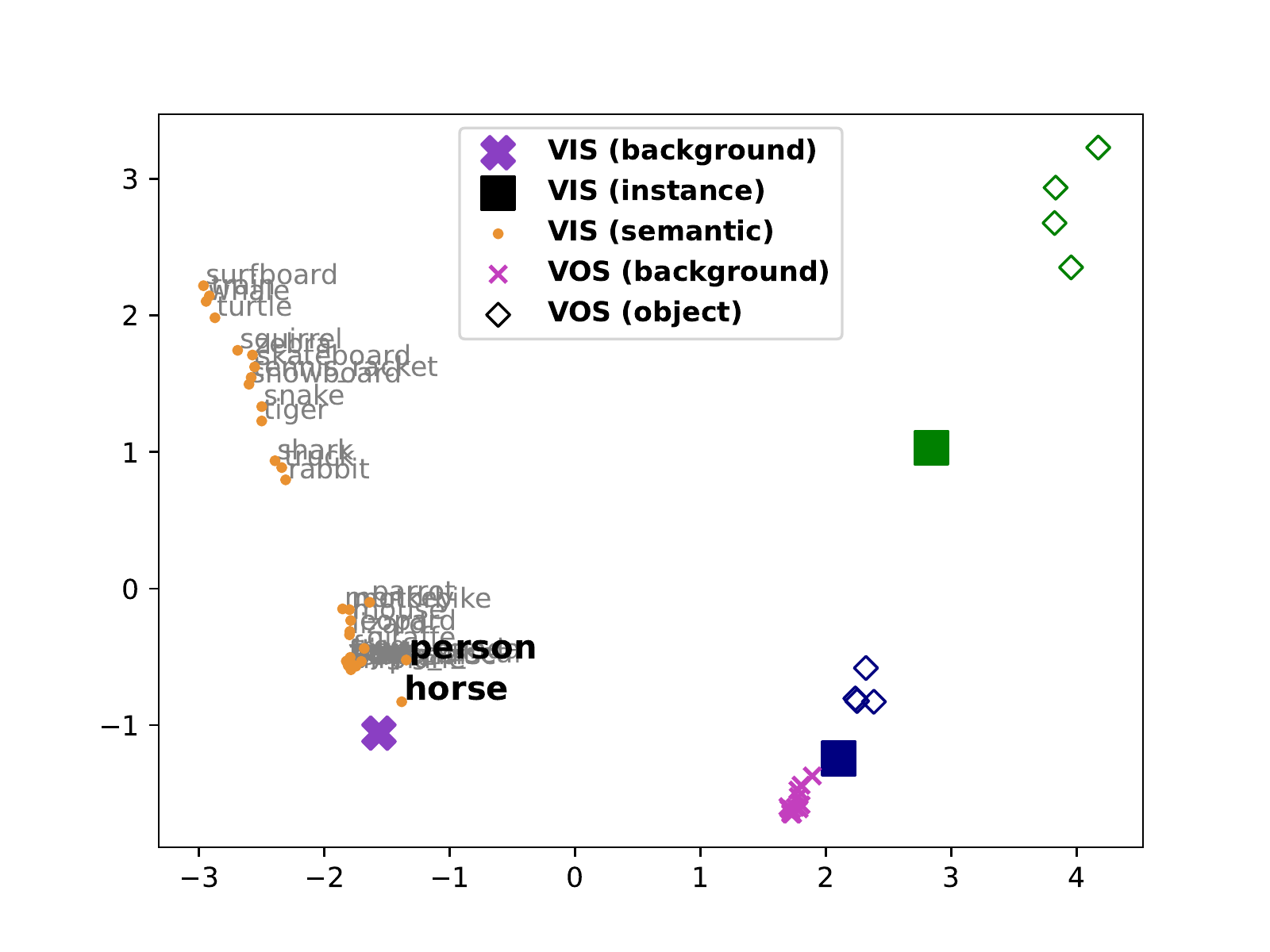}%
\end{minipage}%
\caption{Target query visualization for the `horsejump-high' sequence in DAVIS.}
\label{fig:supp_query_viz_horsejump}
\end{figure*}

\begin{figure*}
\begin{minipage}[c]{0.58\textwidth}%
    \centering
    \def\cellWidth{0.32\linewidth}
    \includegraphics[width=\cellWidth]{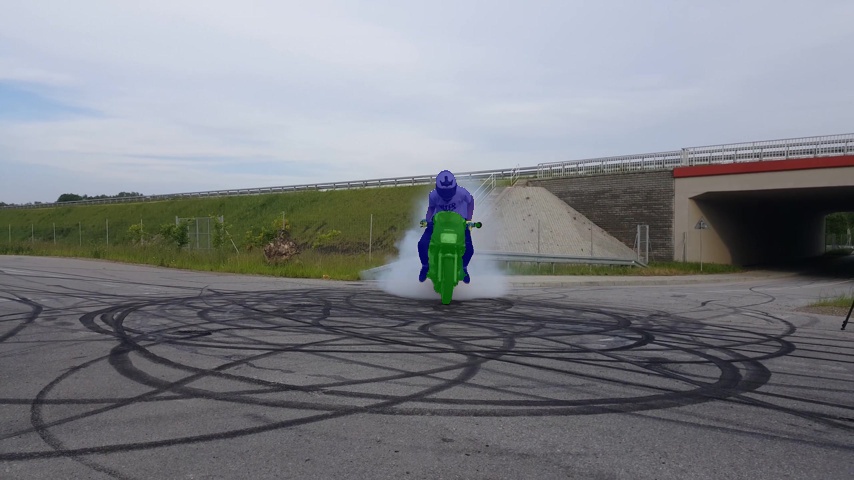}\hfill%
    \includegraphics[width=\cellWidth]{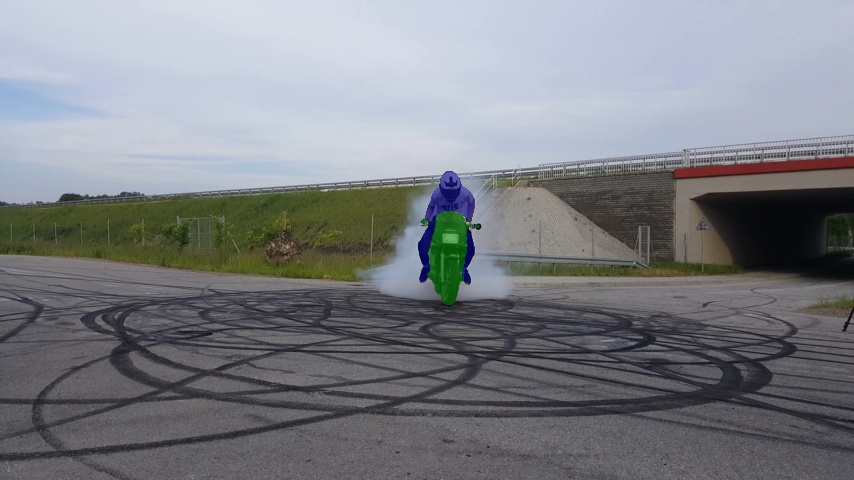}\hfill%
    \includegraphics[width=\cellWidth]{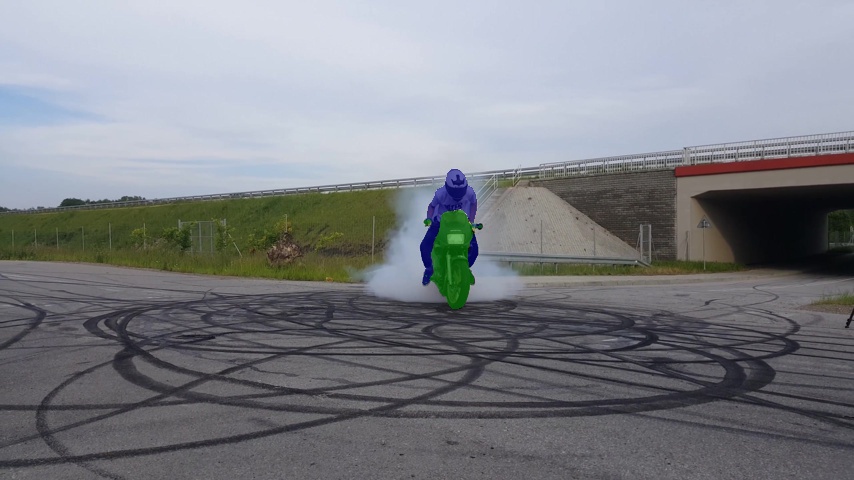} \\
    \includegraphics[width=\cellWidth]{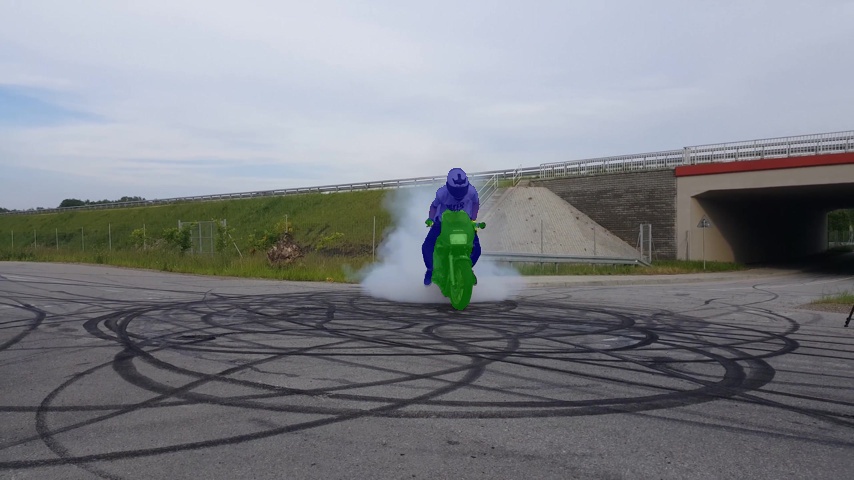}\hfill%
    \includegraphics[width=\cellWidth]{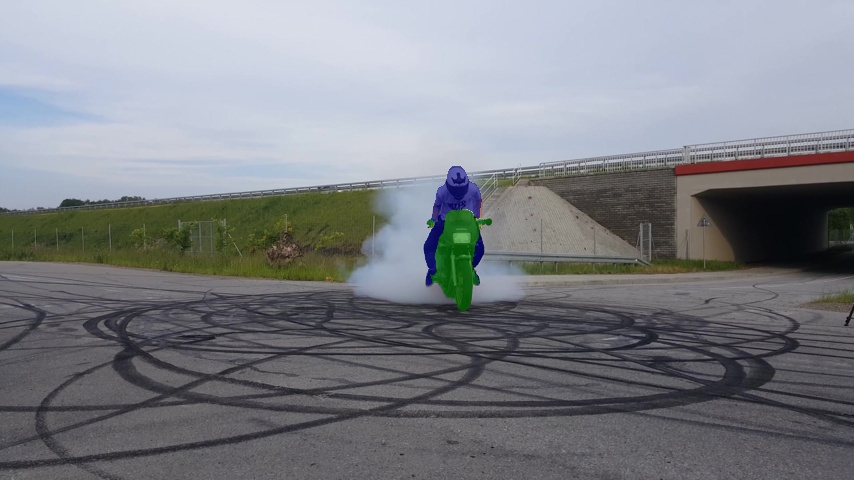}\hfill%
    \includegraphics[width=\cellWidth]{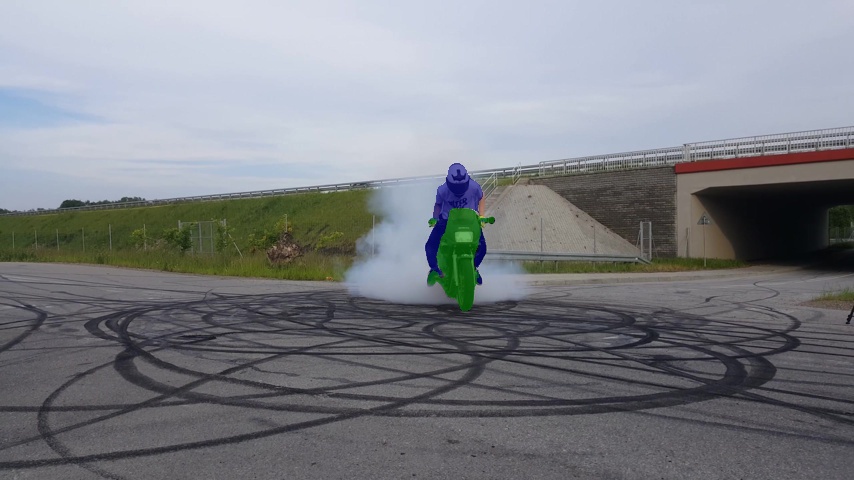} \\
    \includegraphics[width=\cellWidth]{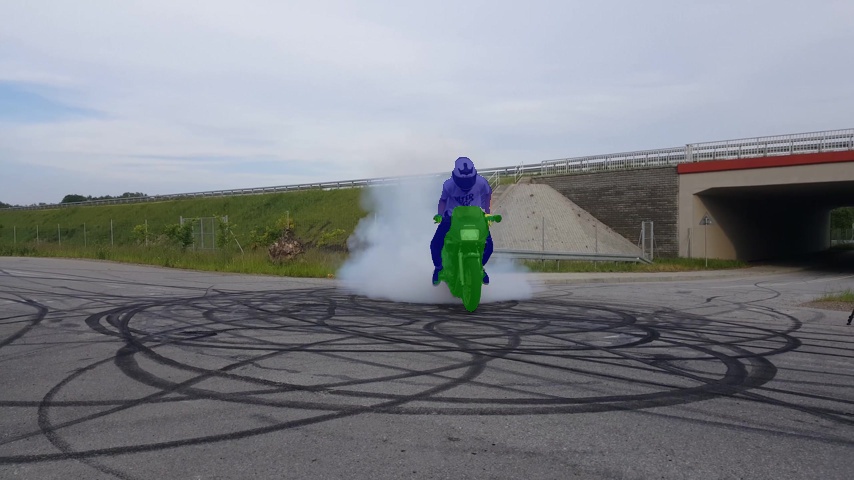}\hfill%
    \includegraphics[width=\cellWidth]{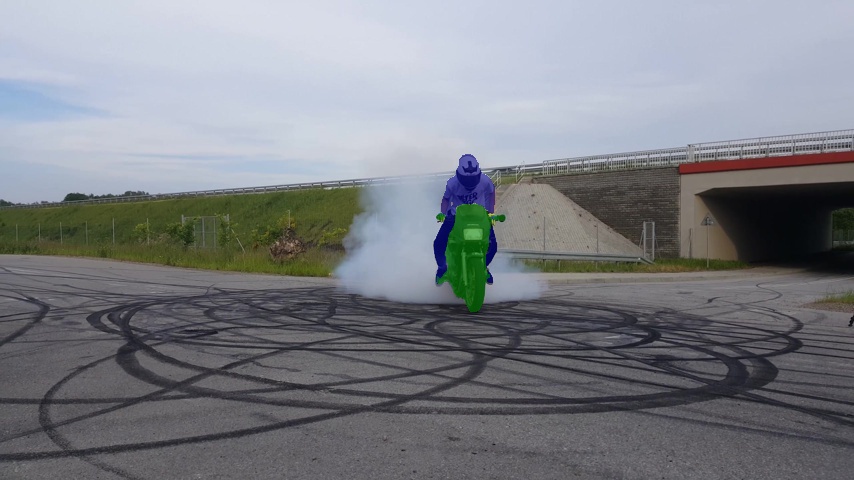}\hfill%
    \includegraphics[width=\cellWidth]{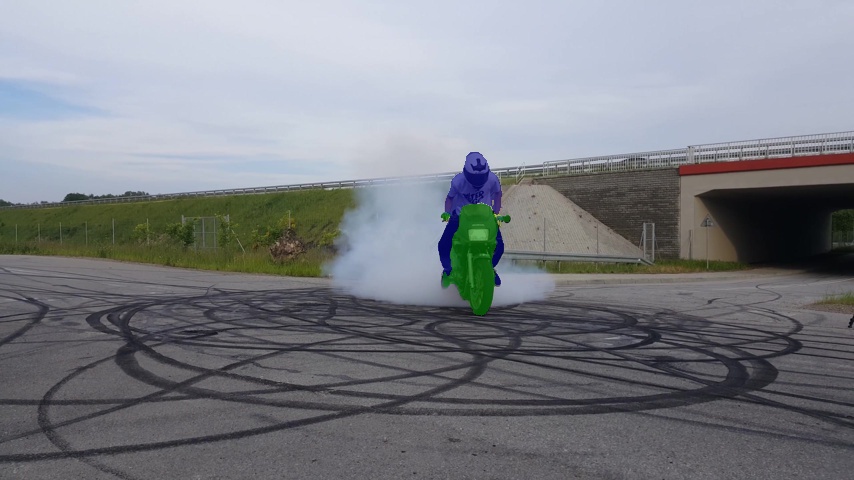} \\
\end{minipage}%
\begin{minipage}[c]{0.42\textwidth}%
\includegraphics[width=\linewidth,trim={1.22cm 0.7cm 1.6cm 1.3cm},clip]{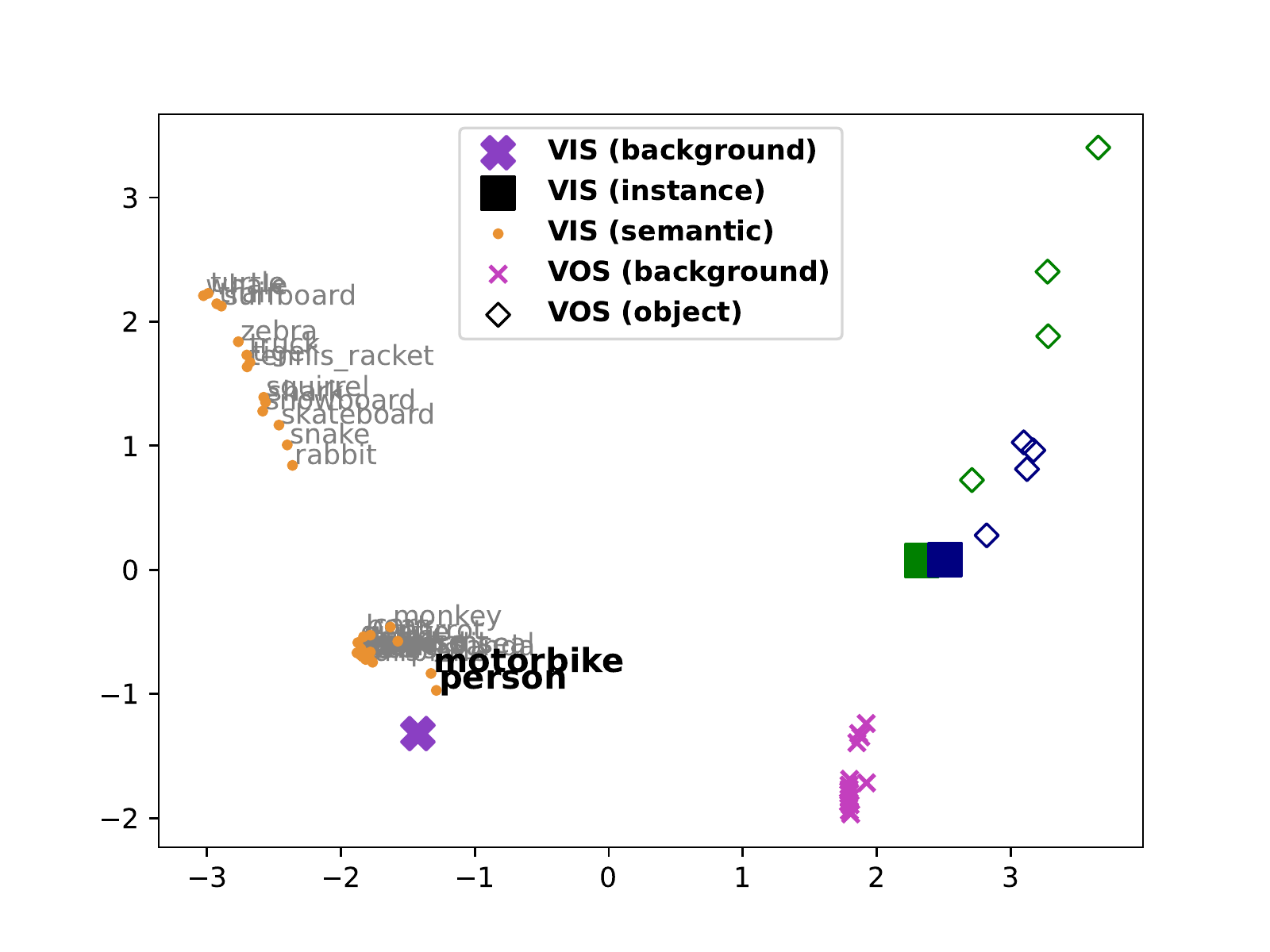}%
\end{minipage}%
\caption{Target query visualization for the `mbike-trick' sequence in DAVIS.}
\label{fig:supp_query_viz_mbiketrick}
\end{figure*}

\begin{figure*}
\begin{minipage}[c]{0.58\textwidth}%
    \centering
    \def\cellWidth{0.32\linewidth}
    \includegraphics[width=\cellWidth]{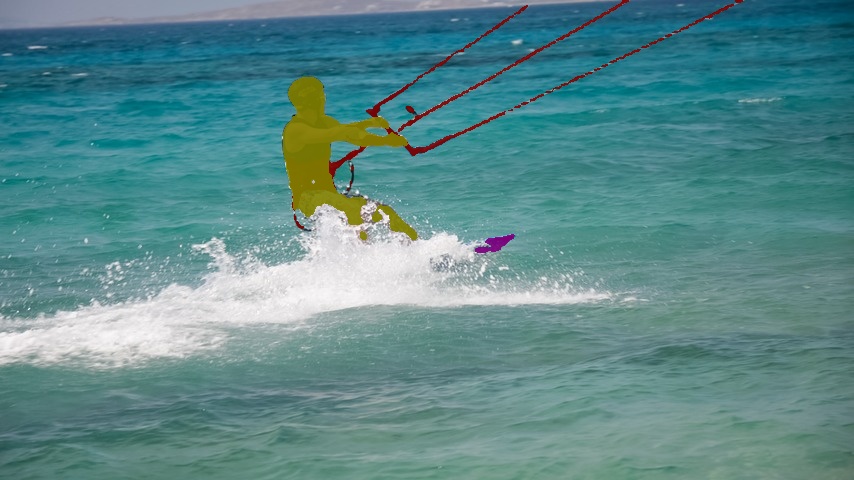}\hfill%
    \includegraphics[width=\cellWidth]{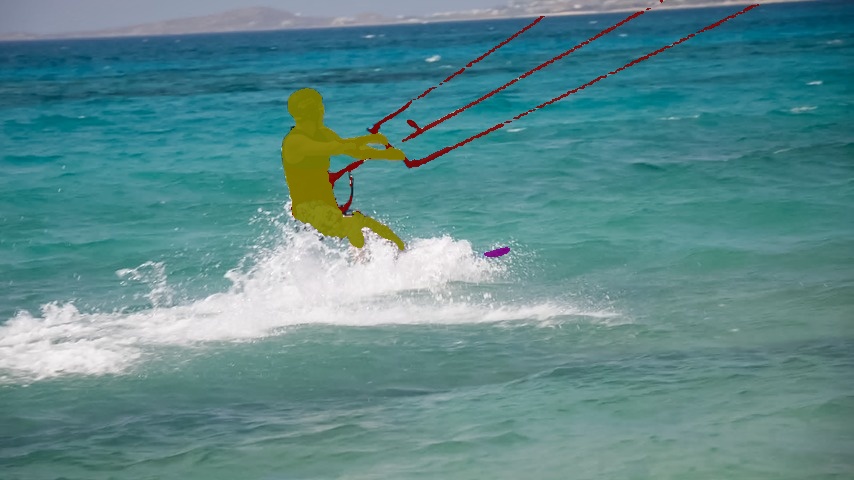}\hfill%
    \includegraphics[width=\cellWidth]{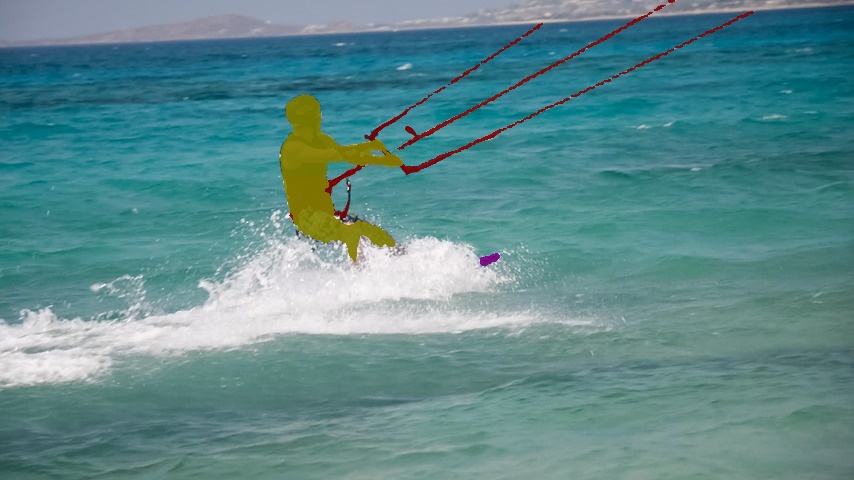} \\
    \includegraphics[width=\cellWidth]{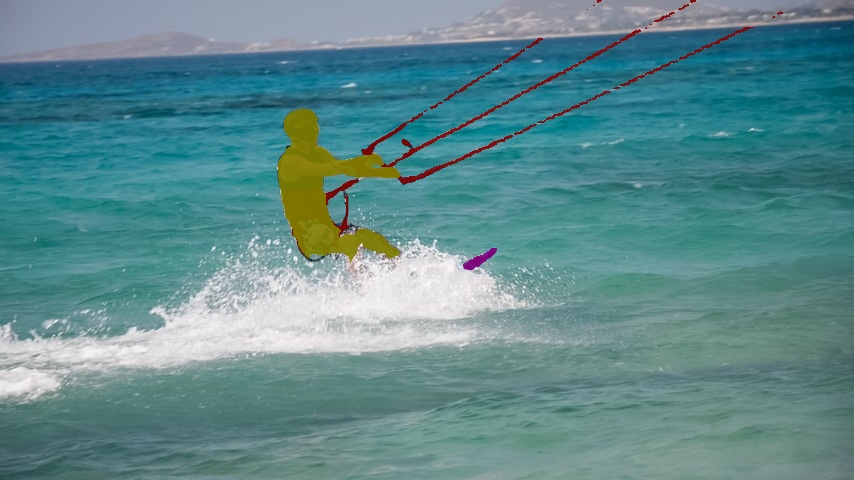}\hfill%
    \includegraphics[width=\cellWidth]{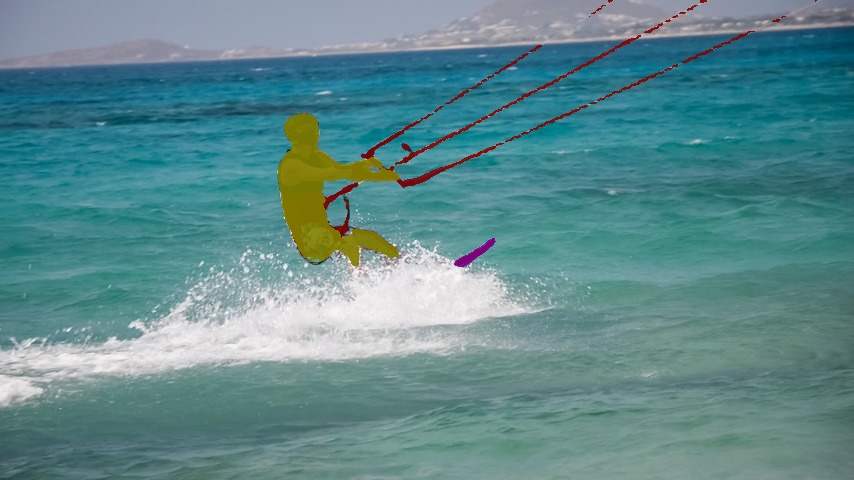}\hfill%
    \includegraphics[width=\cellWidth]{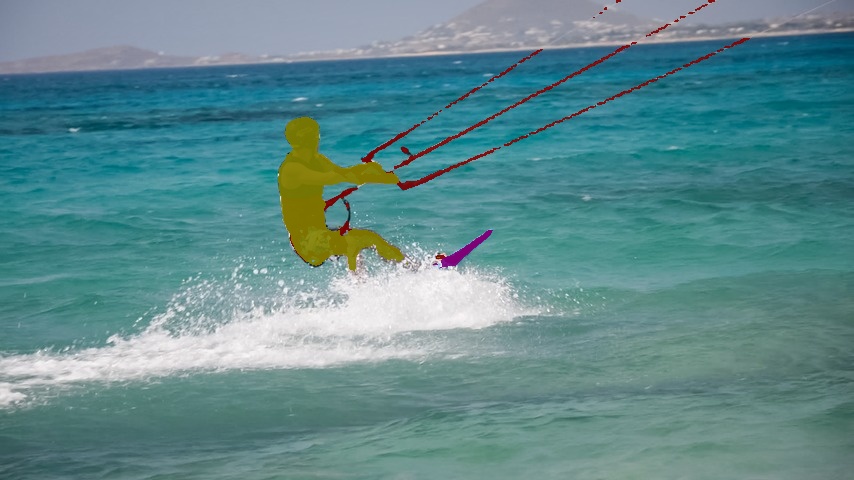} \\
    \includegraphics[width=\cellWidth]{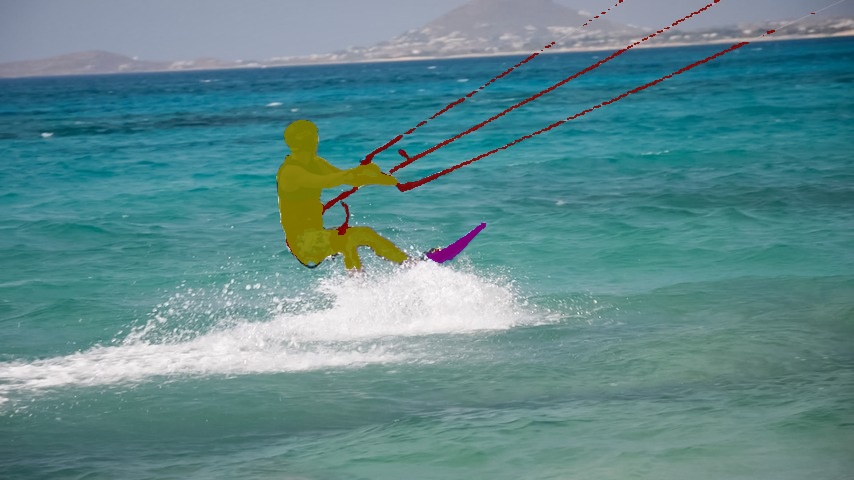}\hfill%
    \includegraphics[width=\cellWidth]{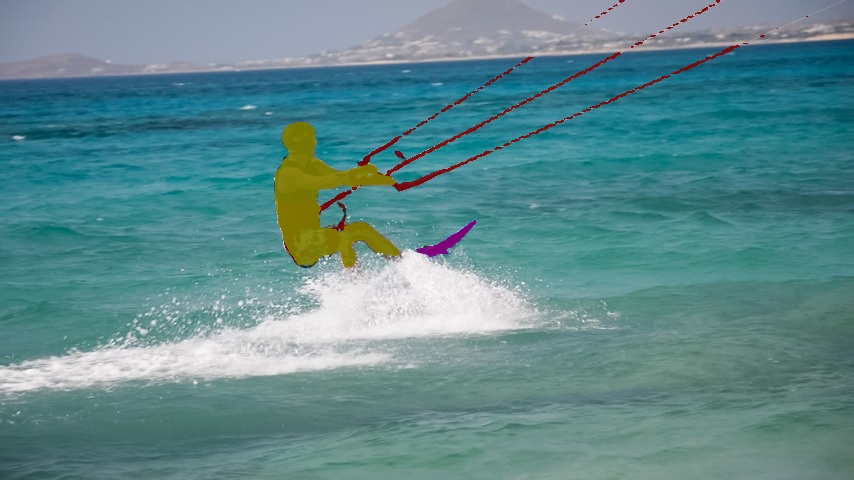}\hfill%
    \includegraphics[width=\cellWidth]{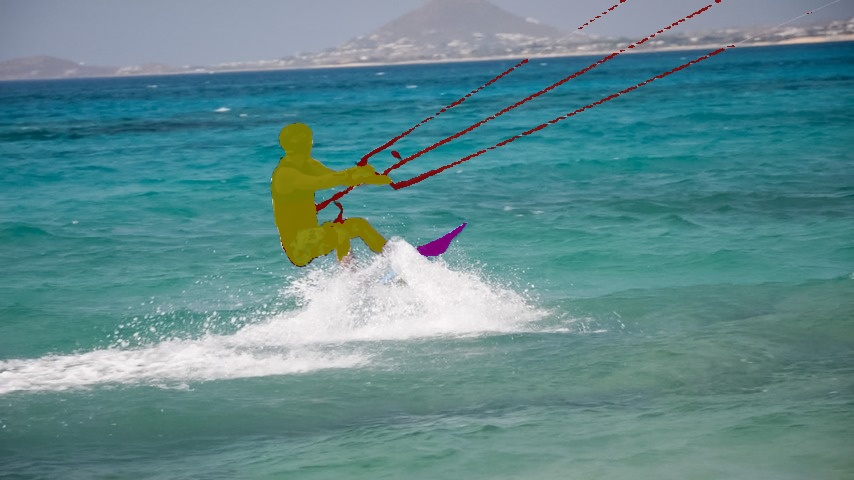} \\
\end{minipage}%
\begin{minipage}[c]{0.42\textwidth}%
\includegraphics[width=\linewidth,trim={1.22cm 0.7cm 1.6cm 1.3cm},clip]{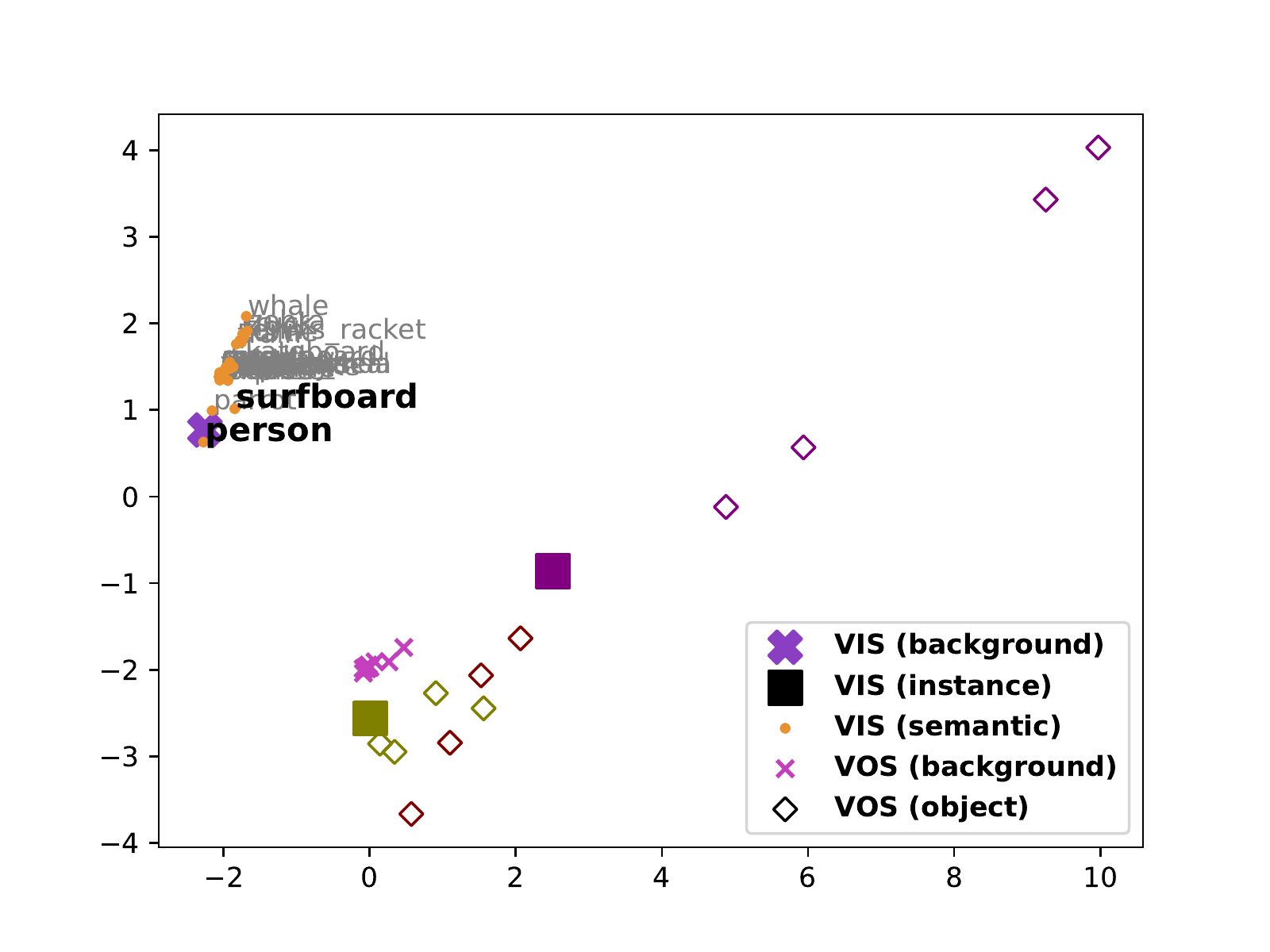}%
\end{minipage}%
\caption{Target query visualization for the `kitesurf' sequence in DAVIS.}
\label{fig:supp_query_viz_kitesurf}
\end{figure*}

To gain some insight into the feature representation learned by \ourMethod{} for different targets, we provide visualizations of the target queries for various tasks and input video clips in Fig.~\ref{fig:supp_query_viz_horsejump},\ref{fig:supp_query_viz_mbiketrick},\ref{fig:supp_query_viz_kitesurf}. The setup is as follows: for each video clip, we run inference twice: (1) as VIS where the targets are all instances belonging to the 40 object classes from YouTube-VIS~\cite{yang2019youtubevis}, and (2) as VOS by providing the first-frame mask for the objects. We deliberately used videos where the set of set of ground-truth objects would be the same for both tasks. The plot on the right visualizes the union of the target query set for both runs by projecting them from 256 dimensions down to 2 using PCA. The image tile on the left shows our model's predicted masks for the target objects (the prediction quality for these video is very good for both VIS and VOS, so we choose one set of results arbitrarily).

For ease of understanding, we use fixed colors for semantic and background queries (as indicated in the plot legend). For the object queries (VOS) and instance queries (VIS), the color of the query point is consistent with the color of the object mask in the image tile. Note that for VOS we used $q_o=4$ object queries per target, hence there are $4$ hollow diamond shaped points per object.

We stress that not all aspects of these plots are intuitively explainable. The main limitation here is the harsh dimensionality reduction from 256 dimensions to 2. Some speculative intuition based on the plots is as follows:

\begin{itemize}
    \item The internal representation for a given object is generally consistent across tasks. As an example, consider the horse and person targets in Fig.~\ref{fig:supp_query_viz_horsejump}: we note that the green query points (person) are close to each other for both VIS and VOS. Likewise the blue query points (horse) follow the same behavior.

    \item The network devotes a large portion of the feature space for instances/objects, and relatively less for the various semantic classes. As seen in all three plots, the semantic queries are tightly clustered together, whereas the instance/object queries are spread out over a larger span of the feature space.
\end{itemize}

\begin{figure*}
    \centering
    \begin{subfigure}[b]{0.495\textwidth}%
         \centering%
         \includegraphics[width=\textwidth,trim={1.22cm 0.7cm 1.5cm 1.3cm},clip]{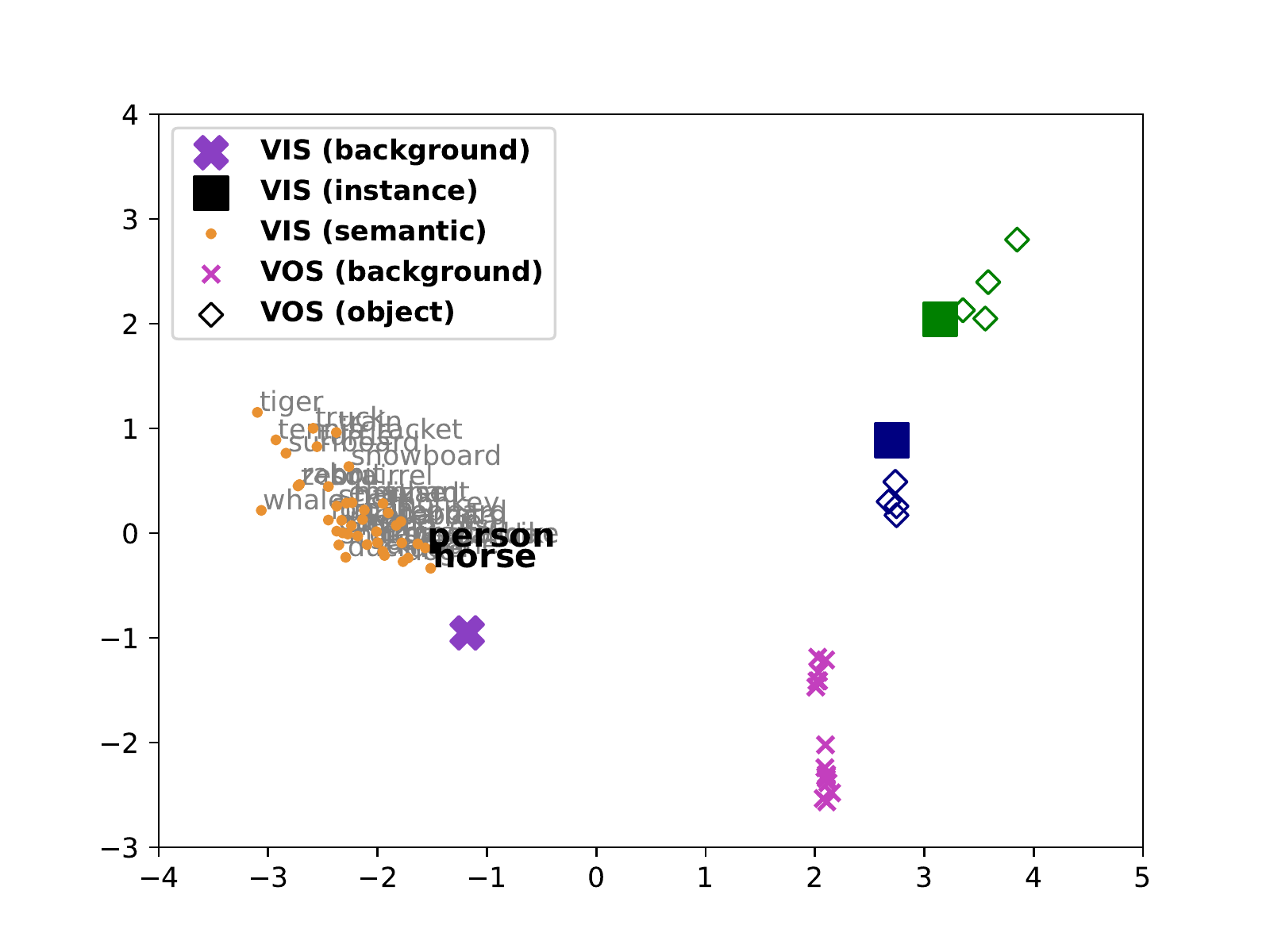}%
         \caption{First layer queries.}%
         \label{fig:supp_query_viz_first_layer}%
    \end{subfigure}%
    \hfill%
    \begin{subfigure}[b]{0.495\textwidth}%
         \centering%
         \includegraphics[width=\textwidth,trim={1.22cm 0.7cm 1.5cm 1.3cm},clip]{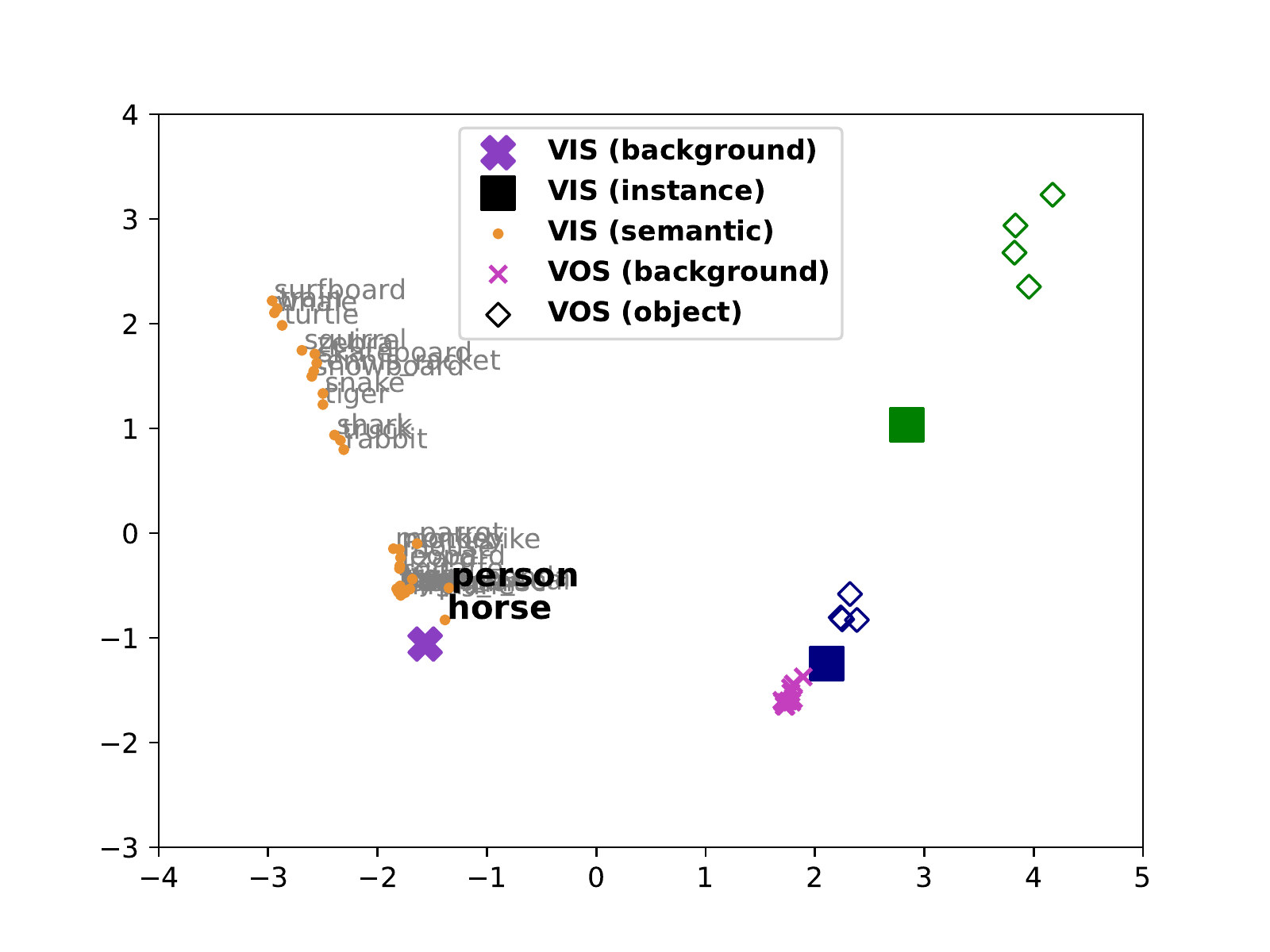}%
         \caption{Last layer queries.}%
         \label{fig:supp_query_viz_first_layer2}%
    \end{subfigure}%
    \caption{Evolution of the different queries from the first layer to the last layer of the transformer decoder. Queries correspond to the `horsejump-high' video from DAVIS as shown in Figure~\ref{fig:supp_query_viz_horsejump}}
    \label{fig:supp_query_viz_temporal_evolution_horsejump}
\end{figure*}

\PAR{Iterative Evolution of Feature Representation.}
Fig.~\ref{fig:supp_query_viz_temporal_evolution_horsejump} shows a side-by-side visualization of how the query feature representation evolves inside the transformer decoder as it iteratively refined the queries using multiple attention layers. The plot on the left shows the queries at the `zeroth' layer (\ie prior to any interaction with the video features), and the plot on the right shows the final output queries from the last layer (these are identical to the plot in Fig.~\ref{fig:supp_query_viz_horsejump} except for the axes range). We note that the distance between the queries for the two objects increases after refinement, and that the semantic queries are also slightly more spaced out after refinement.

\section{Qualitative Results}
\label{sec:supp_qualitative}

The following figures show qualitative results for the different tasks.
\ifArxivMode
\else
Additional results are also present in the video attached to this supplementary archive.
\fi
VIS on YouTube-VIS (Fig.~\ref{fig:suppl_qualitative_cat_dog},\ref{fig:suppl_qualitative_turtle},\ref{fig:suppl_qualitative_lizard_dude}) and OVIS (Fig.~\ref{fig:suppl_qualitative_fish},\ref{fig:suppl_qualitative_sheep},\ref{fig:suppl_qualitative_cats}), VPS on KITTI-STEP (Fig.~\ref{fig:suppl_qualitative_0006},\ref{fig:suppl_qualitative_0010},\ref{fig:suppl_qualitative_0016}), VOS on DAVIS (Fig.~\ref{fig:suppl_qualitative_dance_twirl},\ref{fig:suppl_qualitative_gold-fish},\ref{fig:suppl_qualitative_shooting}), and PET on BURST (Fig.~\ref{fig:suppl_qualitative_3dudes_and_gun},\ref{fig:suppl_qualitative_bear_fight},\ref{fig:suppl_qualitative_cars}).
One can see that TarViS is able to segment a broad range of objects depending on the target queries and overall is good at assigning consistent IDs.
Fig.~\ref{fig:suppl_qualitative_fail} shows an example of a failure case with several ID switches.
Given that we run inference on short overlapping clips, once an ID switch has been made, we cannot recover the original ID.
In the example, it seems that TarViS is not able to track the elephant while they are turning around, even though before and after the turn they are assigned consistent IDs.
Given that we also train on similar short clips, it is not surprising that TarViS struggles here and we could potentially improve this by looking into other training schemes that span longer clips.

\newcommand{\qualitativeRes}[3]{
\begin{figure*}
    \def\cellWidth{0.33\textwidth}%
    \centering%
    \includegraphics[width=\cellWidth]{figures/supplementary/qualitative/#1/#2/0}\hfill%
    \includegraphics[width=\cellWidth]{figures/supplementary/qualitative/#1/#2/1}\hfill%
    \includegraphics[width=\cellWidth]{figures/supplementary/qualitative/#1/#2/2}\\%
    \includegraphics[width=\cellWidth]{figures/supplementary/qualitative/#1/#2/3}\hfill%
    \includegraphics[width=\cellWidth]{figures/supplementary/qualitative/#1/#2/4}\hfill%
    \includegraphics[width=\cellWidth]{figures/supplementary/qualitative/#1/#2/5}\\%
    \caption{#3}%
    \label{fig:suppl_qualitative_#2}%
\end{figure*}
}

\qualitativeRes{ytvis}{cat_dog}{VIS on a YTVIS sequence showing a cat and a dog.}
\qualitativeRes{ytvis}{turtle}{VIS on a YTVIS sequence showing a turtle.}
\qualitativeRes{ytvis}{lizard_dude}{VIS on a YTVIS sequence showing a man and a lizard.}

\qualitativeRes{ovis}{fish}{VIS on an OVIS sequence showing an aquarium with fish.}
\qualitativeRes{ovis}{sheep}{VIS on an OVIS sequence showing several sheep.}
\qualitativeRes{ovis}{cats}{VIS on an OVIS sequence showing three cats.}

\qualitativeRes{kitti}{0006}{VPS on a KITTI STEP sequence showing a busy intersection.}
\qualitativeRes{kitti}{0010}{VPS on a KITTI STEP sequence showing how a car is followed for a while.}
\qualitativeRes{kitti}{0016}{VPS on a KITTI STEP sequence showing a busy pedestrian crossing.}

\qualitativeRes{davis}{dance_twirl}{VOS on a DAVIS sequence of a dancer.}
\qualitativeRes{davis}{gold-fish}{VOS on a DAVIS sequence showing several goldfish.}
\qualitativeRes{davis}{shooting}{VOS on DAVIS sequence an action movie scene.}

\qualitativeRes{burst}{3dudes_and_gun}{PET on a  BURST sequence showing three men and a gun.}
\qualitativeRes{burst}{bear_fight}{PET on a  BURST sequence showing two bears fighting, note there is no ID switch.}
\qualitativeRes{burst}{cars}{PET on a  BURST sequence showing several cars on a street.}

\qualitativeRes{ovis}{fail}{VIS on an OVIS sequence of several elephants and their trainers. This sequence shows that TarVis sometimes has issues with ID switches, especially when the appearance of objects changes, e.g. here the elephants are not tracked consistently while turning around..}

\end{document}